\def\year{2022}\relax
\documentclass[letterpaper]{article} % DO NOT CHANGE THIS
\usepackage{aaai22}  % DO NOT CHANGE THIS
\usepackage{times}  % DO NOT CHANGE THIS
\usepackage{helvet} % DO NOT CHANGE THIS
\usepackage{courier}  % DO NOT CHANGE THIS
\usepackage[hyphens]{url}  % DO NOT CHANGE THIS
\usepackage{graphicx} % DO NOT CHANGE THIS
\usepackage{natbib}  % DO NOT CHANGE THIS AND DO NOT ADD ANY OPTIONS TO IT
\usepackage{caption} % DO NOT CHANGE THIS AND DO NOT ADD ANY OPTIONS TO IT
\usepackage{subcaption}
\usepackage{multirow}
\usepackage{array}
\usepackage{booktabs}
\usepackage{appendix}
\usepackage{enumitem}

\usepackage{algorithm}
\usepackage{algorithmic}
\usepackage{soul}
\newcommand{\INPUT}{\textbf{Inputs: }}
\newcommand{\OUTPUT}{\textbf{Outputs: }}
\algsetup{
  linenodelimiter = {}
}
\usepackage{amsmath,amsfonts,bm}

\def\eqref#1{equation~\ref{#1}}
\def\1{\bm{1}}

\def\rvm{{\mathbf{m}}}

\def\rvx{{\mathbf{x}}}
\def\rvy{{\mathbf{y}}}
\def\rvz{{\mathbf{z}}}

\DeclareMathAlphabet{\mathsfit}{\encodingdefault}{\sfdefault}{m}{sl}
\SetMathAlphabet{\mathsfit}{bold}{\encodingdefault}{\sfdefault}{bx}{n}

\newcommand{\E}{\mathbb{E}}

\newcommand{\R}{\mathbb{R}}

\DeclareMathOperator*{\argmax}{arg\,max}
\DeclareMathOperator*{\argmin}{arg\,min}

\usepackage{amsmath}
\usepackage{amssymb}

\title{Diverse, Global and Amortised\\Counterfactual Explanations for Uncertainty Estimates}
\author{
    Dan Ley$^{1}$, Umang Bhatt$^{1,2}$, Adrian Weller$^{1,2}$\\
}
\affiliations{
    \textsuperscript{\rm 1}University of Cambridge, UK, \textsuperscript{\rm 2}The Alan Turing Institute, UK\\
    {\rm dwl36@cantab.ac.uk, \{usb20, aw665\}@cam.ac.uk}
}
\begin{document}

\maketitle

\begin{abstract}
To interpret uncertainty estimates from differentiable probabilistic models, recent work has proposed generating a single Counterfactual Latent Uncertainty Explanation (CLUE) for a given data point where the model is uncertain, identifying a single, on-manifold change to the input such that the model becomes more certain in its prediction. We broaden the exploration to examine $\delta$-CLUE, the set of potential CLUEs within a $\delta$ ball of the original input in latent space. We study the diversity of such sets and find that many CLUEs are redundant; as such, we propose DIVerse CLUE ($\nabla$-CLUE), a set of CLUEs which each propose a distinct explanation as to how one can decrease the uncertainty associated with an input. We then further propose GLobal AMortised CLUE (GLAM-CLUE), a distinct and novel method which learns amortised mappings on specific groups of uncertain inputs, taking them and efficiently transforming them in a single function call into inputs for which a model will be certain.
Our experiments show that $\delta$-CLUE, $\nabla$-CLUE, and GLAM-CLUE all address shortcomings of CLUE and provide beneficial explanations of uncertainty estimates to practitioners.
\end{abstract}

\section{Introduction}
For models that provide uncertainty estimates alongside their predictions, explaining the source of this uncertainty reveals important information. For instance, determining the features responsible for predictive uncertainty can help to identify in which regions the training data is sparse, which may in turn implicate under-represented sub-groups (by age, gender, race etc). In sensitive settings, domain experts can use uncertainty explanations to appropriately direct their attention to the specific features the model finds anomalous.

In prior work, \citet{adebayo2020debugging} touch on the unreliability of saliency maps for uncertain inputs, and \citet{tsirtsis2021counterfactual} observe that high uncertainty can result in vast possibilities for counterfactuals. Additionally, when models are uncertain, their predictions may be incorrect. We thus consider uncertainty explanations an important precedent for model explanations; only once uncertainty has been explained can state-of-the-art methods be deployed to explain the model's prediction. However, there has been little work in explaining predictive uncertainty.

\noindent\citet{depeweg2017uncertainty} introduce decomposition of uncertainty estimates, though recent work \citep{antoran2021getting} has demonstrated further leaps, proposing to find an explanation of a model's predictive uncertainty for a given input by searching in the latent space of an auxiliary deep generative model (DGM): they identify a single possible change to the input such that the model becomes more certain in its prediction. 
Termed CLUE (Counterfactual Latent Uncertainty Explanations), this method aims to generate counterfactual explanations (CEs) on-manifold that reduce the uncertainty of an uncertain input $\rvx_0$. These changes are distinct from adversarial examples, which find nearby points that change the label~\citep{goodfellow2014explaining}.

\begin{figure*}[ht]
\centering
\includegraphics[width=\textwidth]{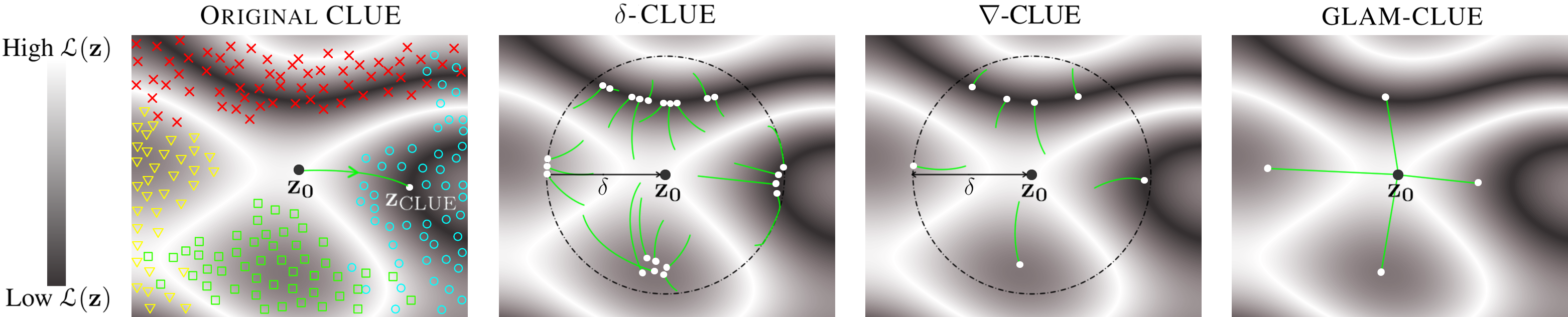}
\caption{\small Conceptual colour map of objective function $\mathcal{L}(z)$ with $\mathbf{z}_0$ located in high cost region. White circles indicate
explanations found. Left: Gradient descent to region of low cost (original CLUE). Training data shown in colour. Left Centre: Gradient descent constrained to $\delta$-ball. Diverse starting points yield diverse local minima, albeit with many redundant solutions ($\delta$-CLUE). Right Centre: Direct optimisation for diversity ($\nabla$-CLUE). Right: Efficient, unconstrained mappings without gradient descent (GLAM-CLUE), allowing computational speedups.}
\label{fig:DeltaFinal}
\end{figure*}

CLUE introduces a latent variable DGM with decoder $\mu_\theta(\mathbf{x}|\mathbf{z})$ and encoder $\mu_\phi(\mathbf{z}|\mathbf{x})$. $\mathcal{H}$ refers to any differentiable uncertainty estimate of a prediction $\mathbf{y}$. The pairwise distance metric takes the form $d(\mathbf{x}, \mathbf{x}_0) = \lambda_xd_x(\mathbf{x}, \mathbf{x}_0) + \lambda_yd_y(f(\mathbf{x}), f(\mathbf{x}_0))$, where $f(\mathbf{x})=\mathbf{y}$ is the model's mapping from an input $\rvx$ to a label, thus encouraging similarity in input space and/or prediction space. CLUE minimises:
\begin{equation}\label{eq:1}\mathcal{L}(\mathbf{z})=\mathcal{H}\left(\mathbf{y}|\mu_{\theta}(\mathbf{x}|\mathbf{z})\right)+d\left(\mu_{\theta}(\mathbf{x}|\mathbf{z}), \mathbf{x}_{0}\right) \end{equation}

% \begin{figure}[H]
%     \centering
%     \includegraphics[scale=0.1]{CLUEworkflow.png}
%     \caption{Workflow for automated decision making with transparency, taken from the original paper: ``Our probabilistic classifier produces a distribution over outputs. In cases of high uncertainty, CLUE allows us to identify features which are responsible for class ambiguity in the input".}
%     \label{fig:CLUEworkflow}
% \end{figure}
\noindent to yield $\mathbf{x}_{\mathrm{CLUE}}=\mu_{\theta}(\mathbf{x}|\mathbf{z}_{\mathrm{CLUE}})$ where $\mathbf{z}_{\mathrm{CLUE}}=\argmin_{\mathbf{z}} \mathcal{L}(\mathbf{z})$.
%In this paper, we tackle the problem of finding multiple, diverse CLUEs.
There are however limitations to CLUE, including the lack of a framework to deal with a diverse set of possible explanations and the lack of computational efficiency. Although finding multiple explanations was suggested, we find the proposed technique to be incomplete.

We start by discussing the multiplicity of CLUEs. Providing practitioners with many explanations for why their input was uncertain can be helpful if, for instance, they are not in control of the recourse suggestions proposed by the algorithm; advising someone to change their age is less actionable than advising them to change a mutable characteristic~\citep{poyiadzi2020face}. Specifically, we develop a method to generate a set of possible CLUEs within a $\delta$ ball of the original point in the latent space of the DGM used: we term this $\delta$-CLUE. We then introduce metrics to measure the diversity in sets of generated CLUEs such that we can optimise directly for it: we term this $\nabla$-CLUE. After dealing with CLUE's multiplicity issue, we consider how to make computational improvements. As such, we propose a distinct method, GLAM-CLUE (GLobal AMortised CLUE), which serves as a summary of CLUE for practitioners to audit their model's behavior on uncertain inputs. It does so by finding translations between certain and uncertain groups in a computationally efficient manner. Such efficiency is, amongst other factors, a function of the dataset, the model, and the number of CEs required; there thus exist applications where either $\nabla$-CLUE or GLAM-CLUE is most appropriate.

\section{Multiplicity in Counterfactuals}
% We start by discussing the multiplicity of CLUEs. Specifically, we first develop a method to generate a set of plausible CLUEs within a $\delta$ of the original point: we term this $\delta$-CLUE. We then introduce metrics to measure the diversity in sets of CLUEs. We then optimize directly for diversity in the set of generated CLUEs: we term this $\nabla$-CLUE.

\subsection{Constraining CLUEs: $\delta$-CLUE}
We propose $\delta$-CLUE \citep{ley2021deltaclue}, which generates a set of solutions that are all within a specified distance $\delta$ of $\mathbf{z}_0=\mu_\phi(\mathbf{z}|\mathbf{x_0})$ in latent space: $\mathbf{z}_0$ is the latent representation of the uncertain input $\mathbf{x}_0$ being explained. We achieve multiplicity by initialising the search randomly in different areas of latent space. While CLUE suggests this, its random generation method and lack of constraint are prone to a) finding minima in a limited region of the space or b) straying far from this region without control over the proximity of CEs (Appendix B).
% Experiments are performed on the MNIST dataset~\citep{lecun1998mnist}, where finding diverse CLUEs amounts to maximising the number of class labels we converge to in the search.
Figure~\ref{fig:DeltaFinal} contrasts the original and proposed objectives (left and left centre respectively).

%\subsection{Original Optimisation Method (CLUE)}

The original CLUE objective uses VAEs~\citep{kingma2014autoencoding} and BNNs~\citep{mackay1992practical} as the DGMs and classifiers respectively. The predictive uncertainty of the BNN is given by the entropy of the posterior over the class labels; we use the same measure. %take uncertainty to be the same. 
The hyperparameters ($\lambda_x, \lambda_y$) control the trade-off between producing low uncertainty CLUEs and CLUEs which are close to the original inputs.
To encourage sparse explanations, we take $d_x(\mathbf{x}, \mathbf{x}_0) = \|\mathbf{x}-\mathbf{x}_0\|_1$. We find this to suffice for our datasets, though other metrics such as FID scores \citep{heusel2018gans} could be used in more complex vision tasks for both evaluation (as in \citet{singla2020explanation}) and optimisation of CEs (see Appendix B). %\ref{appendix:deltaCLUE} Figure~\ref{fig:DeltaFinal} (left centre) shows a conceptual path taken by our optimisation.
% \subsection{Proposed Optimisation Method ($\delta$-CLUE)}
In our proposed $\delta$-CLUE method, the loss function %landscape 
%$\mathcal{L}(z)$
matches Eq~\ref{eq:1}, with the additional $\delta$ requirement as $\mathbf{x}_{\delta-\mathrm{CLUE}}=\mu_{\theta}\left(\mathbf{x}|\mathbf{z}_{\delta-\mathrm{CLUE}}\right)$ where $\mathbf{z}_{\delta-\mathrm{CLUE}}=\argmin_{\mathbf{z}:\ \rho(\mathbf{z}, \mathbf{z}_0)\leq\delta} \mathcal{L}(\mathbf{z})$ and $\mathbf{z_0}=\mu_\phi(\mathbf{z}|\mathbf{x_0})$.
We choose $\rho(\mathbf{z}, \mathbf{z}_0)=\|\mathbf{z}-\mathbf{z}_0\|_2$ (the $\ell_2$ norm) in this paper, as shown in Figure~\ref{fig:DeltaFinal}. We first set $\lambda_x=\lambda_y=0$ to explore solely the uncertainty landscape, given that the size of the $\delta$-ball removes the strict need for the distance component in $\mathcal{L}(z)$ and grants control over the locality of solutions, before trialling $\lambda_x=0.03$.
We apply the  $\delta$ constraint at each stage of the optimisation  (Figure~\ref{fig:DeltaFinal}, left centre), as in Projected Gradient Descent~\citep{boyd2004convex}.

%\subsection{Proposed Initialisation Method}

\begin{figure}[ht]
\begin{subfigure}{0.24\textwidth}
    \centering
    \includegraphics[width=0.95\textwidth]{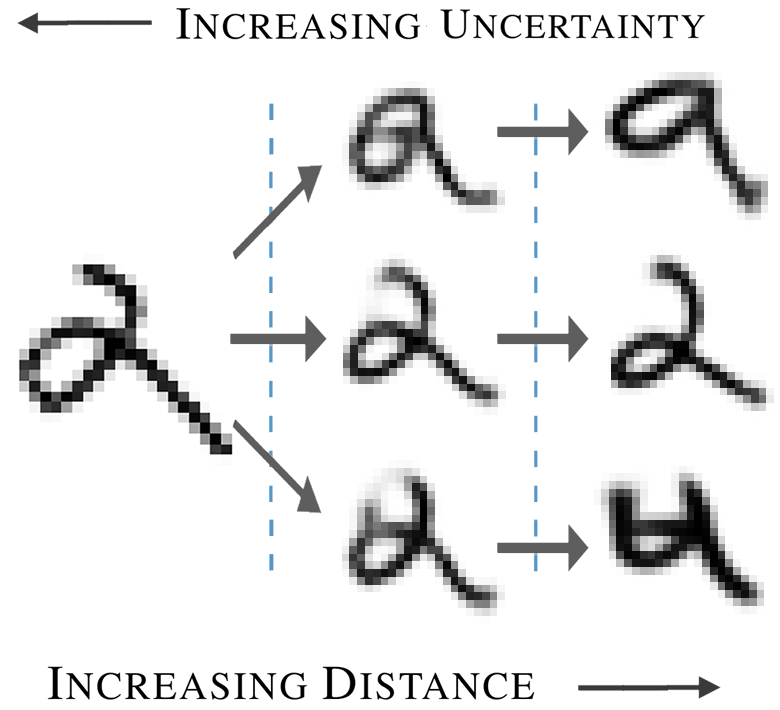}
    \caption{\small Uncertain $2\rightarrow9,2,4$.}
\end{subfigure}\begin{subfigure}{0.24\textwidth}
    \centering
    \includegraphics[width=0.95\textwidth]{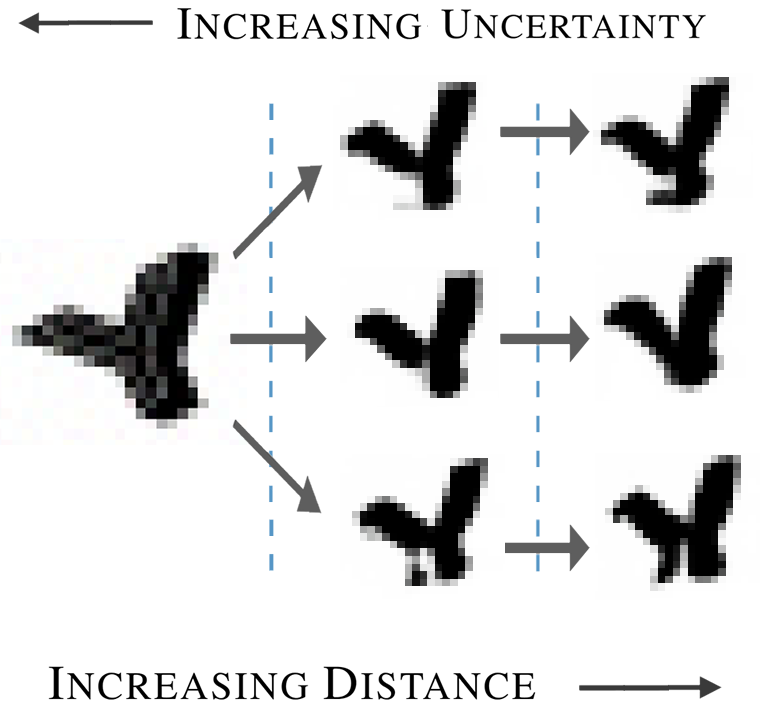}
    \caption{\small Uncertain Y $\rightarrow$ Y, V, X.}
\end{subfigure}
\caption{\small Visualisation of the trade-off between uncertainty $\mathcal{H}$ and distance $d$. Left: MNIST. Right: Synbols.}
\label{fig:synbolsmnist}
\end{figure}

For each uncertain input, we exploit the non-convexity of CLUE's objective to generate diverse $\delta$-CLUEs by initialising in different regions of latent space %to converge to different local minima
(Figure \ref{fig:DeltaFinal}). While previous work has considered sampling the latent space around an input \citep{pawelczyk2020learning}, we find that subsequent gradient descent yields improvements. 
Example results are in Figure \ref{fig:synbolsmnist}. $\delta$-CLUE is a special case of Algorithm~\ref{algorithm:divCLUEsimultaneous}, or explicitly Algorithm 3 %~\ref{algorithm:deltaCLUE}
(Appendix B).%\ref{appendix:deltaCLUE}
%We propose multiple initialisation schemes, $\mathcal{S}_i$; some may randomly initialise within the $\delta$-ball, while others could use training data or class boundaries to determine starting points.
\subsection{Diversity Metrics for Counterfactual Explanations}\label{measurediversity}
Once we have generated a set of viable CLUEs, we desire to measure the diversity within the set; as such, we require candidate convex similarity functions between points, which could be applied either pairwise or over all counterfactuals. We consider these between counterfactual labels (prediction space) or between counterfactuals themselves (input or latent space). A given diversity function $D$ can be applied to a set of $k>0$ counterfactuals in an appropriate space i.e. $D(\rvx_1,...,\rvx_k)$, $D(\rvz_1,...,\rvz_k)$ or $D(\rvy_1,...,\rvy_k)$ where $\rvx_i\in\R^{d'}$, $\rvz_i\in\R^{m'}$ and $\rvy_i\in\R^{c'}$ (we define the hard prediction $y_i=\max\limits_j(\rvy_i)_j$). Table~\ref{tab:diversity} summarises the metrics.

\begin{table}[ht]
\scriptsize
\begin{center}
\begin{tabular}{| >{\centering\arraybackslash}m{2.2cm} | >{\centering\arraybackslash}m{5.2cm}|}
\hline
{\sc\small Diversity Metric} & {\sc\small Function ($D$)}\\
\hline
{\sc\small Determinantal Point Processes} & $\det(\mathbf{K})$ where $\mathbf{K}_{i,j}=\cfrac{1}{1+d(\rvx_i, \rvx_j)}$\\
{\sc\small Average Pairwise Distance} & $\cfrac{1}{\binom{k}{2}}\sum\limits_{i=1}^{k-1}\sum\limits_{j=i+1}^{k}d(\rvx_i, \rvx_j)$\\
{\sc\small Coverage} & $\cfrac{1}{d'}\sum\limits_{i=1}^{d'}\left(\max_j(\rvx_j-\rvx_0)_i+\max_j(\rvx_0-\rvx_j)_i\right)$\\
{\sc\small Prediction Coverage} & $\cfrac{1}{c'}\sum\limits_{i=1}^{c'}\max_j[(\rvy_j)_i]$\\
{\sc\small Distinct Labels} & $\cfrac{1}{c'}\sum\limits_{j=1}^{c'}\mathbf{1}_{[\exists i\ :\ y_i=j]}$ \\
{\sc\small Entropy of Labels} & $-\cfrac{1}{\log c'}\sum\limits_{j=1}^{c'}p_j(k)\log p_j(k)$ \\
\hline
\end{tabular}
\end{center}
\caption{\small Diversity metrics, $D$. Where necessary, we define $D=0$ for $k=1$ and take $d$ to be some arbitrary distance metric.}
\label{tab:diversity}
\end{table}

%\begin{itemize}
%\item If, pre-optimisation, we can globally define modes within class labels, these could effectively replace class labels. If we define modes post-optimisation, as we have been doing, then they cannot be used within the algorithm from which they came i.e. we can't optimise for modes if we don't know what the modes are. Finding modes on the fly seems unstable as clustering algorithms will start off very refined when number of ... ?
%\item Some metrics may be much better suited to have large weights and low number of counterfactuals (I predict Entropy of Label Distribution and Coverage). Metrics that are perhaps inferior in optimisation could still be useful in evaluation.
%\item Clarify notation of div/CLUEs. Entropy for distance?
%\end{itemize}

\textbf{Leveraging Determinantal Point Processes}: We build on \citet{Mothilal_2020} to leverage determinantal point processes, referred to as DPPs \citep{Kulesza_2012}, as $\det(\mathbf{K})$ in Table \ref{tab:diversity}. DPPs implicitly normalise to $0\leq D\leq 1$. This metric is effective overall and achieves diversity by diverting attention away from the most popular (or salient) points to a diverse group of points instead. However, matrix determinants are computationally expensive for large $k$.

%[How sensitive is this to changes in gen model and in dist?]

%[Still unsure how we would apply this to modes we find \textit{whilst running the algorithm (sequential)}: clustering modes on-the-fly and trying to maximise the number of modes is probably just going to be the same as maximising distance between counterfactuals?]

\textbf{Diversity as Average Pairwise Distance}: We can calculate diversity as the average distance between all distinct pairs of counterfactuals (as in \citet{bhatt2021divine}). While we can adjust for the number of pairs (accomplishing invariance to $k$), this metric does not satisfy $0\leq D\leq1$, scaling instead with the pairwise distances characterised by the dataset.

\textbf{Coverage as a Diversity Metric}: Previous work in interpretability has leveraged the notion of coverage as a measure of the quality of sets of CEs. \citet{ribeiro2016why} define coverage to be the sum of distinct features contained in a set, weighted by feature importance: this could be applied to CEs to suggest a way of optimally choosing a subset from a full set of CEs. \citet{plumb2020explaining} introduce coverage as a measure of the quality of global CEs. Herein, we interpret coverage as a measure of diversity, using it directly for optimisation and evaluation of CEs. The metric, as given in Table \ref{tab:diversity}, rewards changes in both positive and negative directions separately (though penalises a lack of changes in positive/negative directions). See Appendix C. %For each feature, we find the 2 CEs that produce the largest positive and largest negative change for that feature and sum their magnitudes. We repeat this over all features, summing the results. $D$ is bounded by the scale of the features (as detailed in Appendix C). %\ref{appendix:diversitymetrics}).

%the sum, over all features, of the magnitudes of the largest positive and largest negative changes proposed by any one counterfactual for that feature.

%Unlike the metrics presented up to this point, we need not normalise the sum since the max function does not explicitly scale with $k$. [Check this.] [Works with latent space.]

\textbf{Prediction Coverage}: Since %$\rvy_0$ as an estimate of the true label is inaccurate for uncertain predictions, and
rewarding negative changes in $\rvy$-space is redundant (maximising the prediction of one label implicitly minimises the others), we adjust the coverage metric in $\rvy$-space to be the maximum prediction for a particular label found in a set of CEs, averaged over all predictions. This satisfies $\frac{1}{c'}\leq D\leq1$, where we require at least $k=c'$ CEs to achieve $D=1$, equivalent to finding at least one fully confident prediction for each label.

%Since we use entropy for uncertainty, and reject values with high entropy, this is essentially just maximising number of distinct labels. Think: a prediction of 0.5 will add to the coverage but will be discarded because it has high uncertainty, hence only high predictions (near 1) will be added, which is the same as counting distinct labels. Maybe.

\textbf{Targeting Diversity of Class Labels}: While recent work focuses on producing diverse explanations for binary classification problems \citep{russell2019efficient} and others summarise current methods therein \citep{pawelczyk2020counterfactual}, these metrics perform well in applications rich in class labels, and conversely are likely ineffective in binary tasks. Posterior probabilities are defined as $\rvy\in\R^{c'}$ and $y_i=\argmax_i\rvy_i$. We define the probability of class $j$ as $p_j(k)=\frac{\sum_{i=1}^k\mathbf{1}_{[y_i=j]}}{k}=\frac{\text{number of counterfactuals in class }j}{\text{number of counterfactuals}}$. Using this, we suggest diversity through the \textbf{Number of Distinct Labels} found, as well as the \textbf{Entropy of the Label Distribution}. The former metric loses its effect once all labels are found, whereas the latter does not. The former satisfies $0\leq D\leq1$, and given that the maximum entropy of a $c'$ dimensional distribution is $\log(c')$, so too does the latter.

%\textbf{Distinct labels}: very simple, ranges between 1 and $c'$, maximises total number of distinct class labels in the set of counterfactuals, however loses its effect once all classes are found (at worse, once $k>c'$), e.g. for 100 CLUEs and 10 classes, it sees nothing wrong with finding one instance of each class, and then 90 instances of a particular class it prefers. A problem rooting from the fact that this diversity metric can only increase as you add more counterfactuals to a set. However, if it's unlikely that all classes are found in a particular case this metric may have utility.

%\textbf{Entropy of labels:} unlike number of distinct labels, diversity can decrease as you add counterfactuals to the set (a positive). Normalise with $\log(c')$, since this is the maximum entropy of a $c'$ dimensional distribution.

\subsection{Optimizing for Diversity: $\nabla$-CLUE}
\label{optimisediversity}

The diversity metrics defined in Table~\ref{tab:diversity} find utility in the optimisation of a set of $k$ counterfactuals. We optimise for diversity in the CLUEs we generate through an explicit diversity term in our objective for the CLUEs found. We call this DIVerse CLUE or $\nabla$-CLUE. We posit that whilst some aforementioned metrics may perform poorly during optimisation, we retain them for evaluation.

Once the diversity metric is selected, the optimisation of $k$ counterfactuals can be performed \textbf{simultaneously} (Algorithm~\ref{algorithm:divCLUEsimultaneous}) in latent space \citep{Mothilal_2020}, or \textbf{sequentially} (Appendix D), %\ref{appendix:}
where the approach is analogous to a greedy algorithm of the former approach. The notation $X_\mathrm{CLUE}=\{\rvx_1,...,\rvx_k\}$ is adopted to represent a set of $k$ counterfactuals (similarly $Z_\mathrm{CLUE}$ and $Y_\mathrm{CLUE}$).

\begin{algorithm}[t]
%\small
\newdimen\origiwspc
\origiwspc=\fontdimen1\font
\fontdimen1\font=12em
 \INPUT $\delta$, $k$, $\mathcal{S}$, $r$, $\mathbf{x}_0$, $d$, $\rho$, $\mathcal{H}$, $\mu_\theta$, $\mu_\phi$, $D$, $\lambda_D$
 
 \fontdimen1\font=\origiwspc
 \begin{algorithmic}[1]
 \STATE Initialise $\varnothing$ of CLUEs: $X_{\mathrm{CLUE}}=\{\}$;
 \STATE Set $\delta$-ball centre of $\mathbf{z}_0=\mu_\phi(\mathbf{z}|\mathbf{x}_0)$;
 \FOR{$1\leq i\leq k$}
 \STATE Set initial value of $\mathbf{z}_i = \mathcal{S}(\mathbf{z}_0, r, i, k)$;
 \ENDFOR
 \WHILE{\textit{loss $\mathcal{L}$ has not converged}}
 %\STATE Decode: $\rvx_{1:k} = \mu_\theta(\rvx_{1:k}|\rvz_{1:k})$;
  \FOR{{$1\leq i\leq k$}}
   \STATE Decode: $\rvx_i = \mu_\theta(\rvx|\rvz_i)$;
   \STATE Use predictor to obtain $\mathcal{H}(\rvy|\rvx_i)$;
   \STATE $\mathcal{L}(\rvz_i) = \mathcal{H}(\rvy|\rvx_i)+d(\rvx_i,\rvx_0)$;
  \ENDFOR
  \STATE {$\mathcal{L}(\rvz_1,...,\rvz_k)=-\lambda_DD(\rvz_1,...,\rvz_k)+\frac{1}{k}\sum\limits_{i=1}^k\mathcal{L}(\rvz_i)$;}
  \STATE {Update $\rvz_1,...,\rvz_k$ with $\nabla_{\rvz_1,...,\rvz_k}\mathcal{L}(\rvz_1,...,\rvz_k)$;}
  \FOR{{$1\leq i\leq k$}}
      \STATE Constrain $\rvz_i$ to $\delta$ ball using $\rho(\rvz_i, \rvz_0)$;
  \ENDFOR
 \ENDWHILE
 \FOR{{$1\leq i\leq k$}}
    \STATE Decode explanation: $\rvx_i=\mu_\theta(\rvx|\mathbf{z}_i)$;
    \IF{$\mathcal{H}(\mathbf{y}|\mathbf{x}_i) < \mathcal{H}_{\mathrm{threshold}}$}
        \STATE $X_{\mathrm{CLUE}}\leftarrow X_{\mathrm{CLUE}}\cup\mathbf{x}_i$;
    \ENDIF
 \ENDFOR
\end{algorithmic}
\OUTPUT $X_{\mathrm{CLUE}}$, a set of $n\leq k$ diverse CLUEs
\caption{$\nabla$-CLUE (\textbf{simultaneous})}
\label{algorithm:divCLUEsimultaneous}
\end{algorithm}

\begin{figure}[ht]
    \centering
    \includegraphics[width=0.45\textwidth]{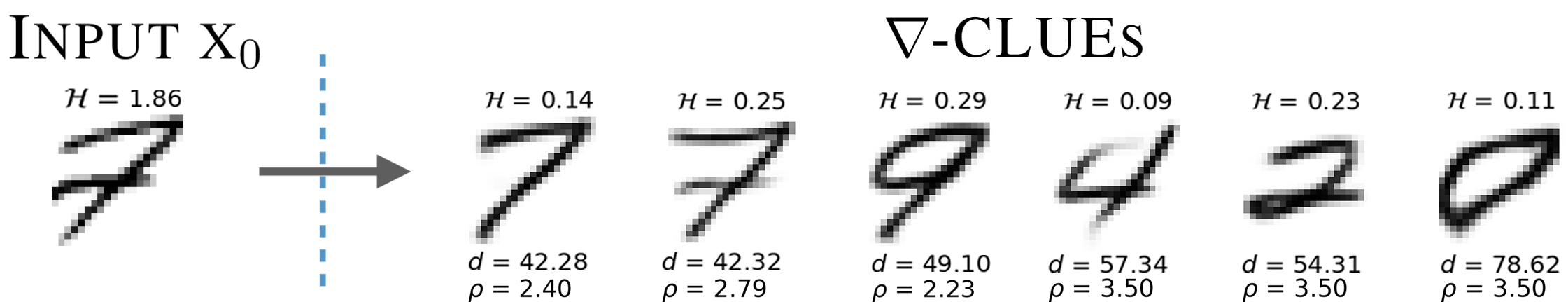}
    \caption{\small We produce a \textbf{diverse set} of candidate explanations that show how to reduce predictive uncertainty while remaining close to $x_0$ in both input and latent space ($\mathcal{H}$ is uncertainty, $d$ is input distance, $\rho$ is latent distance). We see that the left image might most easily be resolved into a confident 7 or 9. Results are taken from a larger set of $\nabla$-CLUEs and are not exemplary of setting $k=5$.}
    \label{fig:digits}
\end{figure}

We denote an initialisation scheme $\mathcal{S}$ of radius $r$ to generate starting points for the gradient descent. Note that the removal of the $\delta$ constraint or the initialisation may be achieved at $\delta=\infty$ and $r=0$ respectively (although the latter yields the same counterfactual $k$ times as a result of symmetry). Thus, \textbf{the} $\pmb{\nabla}$\textbf{-CLUE algorithm is equivalent to }$\pmb{\delta}$\textbf{-CLUE when $\pmb{\lambda_D=0}$}, which is itself equivalent to the original CLUE algorithm when $\delta=\infty$, $r=0$ and $k=1$. Example results are in Figure~\ref{fig:digits}.

\textbf{Simultaneous Diversity Optimisation (Algorithm~\ref{algorithm:divCLUEsimultaneous})}: By optimising simultaneously over $k$ counterfactuals in latent space, issues with how the diversity metric $D$ might scale with $k$ can be avoided. We have the simultaneous optimisation problem of minimising $\mathcal{L}(\rvz_1,...,\rvz_k)=-\lambda_DD(\rvz_1,...,\rvz_k)+\frac{1}{k}\sum_{i=1}^k\mathcal{L}(\rvz_i)$ where $\mathcal{L}(\mathbf{z}_i)=\mathcal{H}\left(\mathbf{y}|\mu_{\theta}(\mathbf{x}|\mathbf{z}_i)\right)+d\left(\mu_{\theta}(\mathbf{x}|\mathbf{z}_i), \mathbf{x}_{0}\right)$, to yield $X_{\mathrm{CLUE}}=\mu_{\theta}\left(X|Z_{\mathrm{CLUE}}\right)$ where $Z_\mathrm{CLUE} = \argmin_{\rvz_1,\ldots,\rvz_k} = \mathcal{L}(\rvz_1,...,\rvz_k)$.
Note that we apply the diversity function in latent space; it could equally be applied in input space.

\textbf{Sequential Diversity Optimisation (Appendix D)}:
Given a set of counterfactuals $Z_\mathrm{CLUE}$ (initially the empty set $\varnothing$), we can apply $\nabla$-CLUE sequentially, appending each new counterfactual to the set. At each iteration, we minimise $\mathcal{L}(\mathbf{z})=\lambda_DD(Z_\mathrm{CLUE}\cup\rvz)+\mathcal{H}\left(\mathbf{y}|\mu_{\theta}(\mathbf{x}|\mathbf{z})\right)+d\left(\mu_{\theta}(\mathbf{x}|\mathbf{z}), \mathbf{x}_{0}\right)$ to yield $\rvz_\mathrm{CLUE}$ which we append to the set. %\ref{algorithm:divCLUEsequential}.

\section{Global and Amortised Counterfactuals}
CLUE primarily focuses on local explanations of uncertainty estimates, as \citet{antoran2021getting} propose a method for finding a single, small change to an uncertain input that takes it from uncertain to certain with respect to a classifier. Such local explanations can be computationally expensive to apply to large sets of inputs. Large sets of counterfactuals are also difficult to interpret. We thus face challenges when using them to summarise global uncertainty behaviour, which is important in identifying areas in which the model does not perform as expected or the training data is sparse.

We desire a computationally efficient method that requires a finite portion of the dataset (or a finite set of CEs) from which global properties of uncertainty can be learnt and applied to unseen test data with high reliability. We propose GLAM-CLUE (GLobal AMortised CLUE), which achieves such reliability with considerable speedups.
\begin{figure*}[ht]
    \centering
    \includegraphics[width=0.9\textwidth]{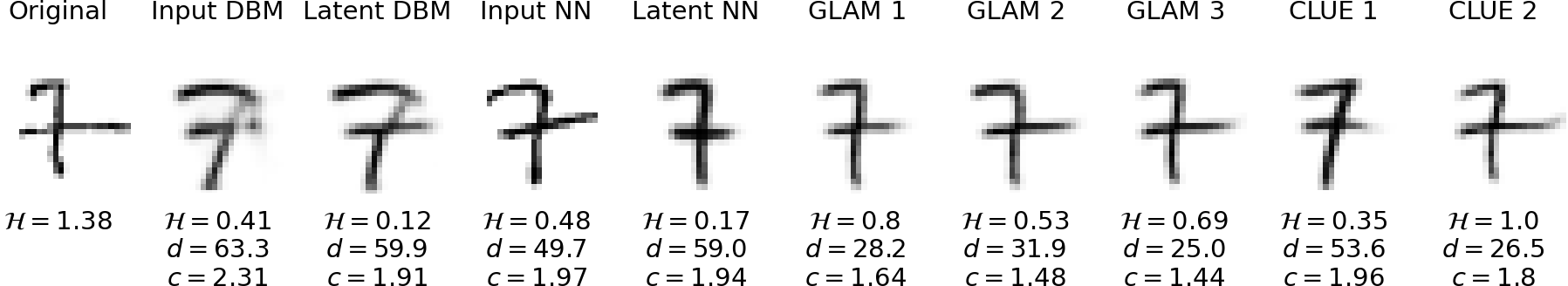}
    \caption{\small Comparison of the explanations generated for an uncertain input (far left) by the baselines, GLAM-CLUE, and CLUE. $\mathcal{H}$ is uncertainty, $d$ is input distance, $c=\mathcal{H}+\lambda_xd$ is cost. Low uncertainties in some baseline schemes are invalidated by unrealistic distances. GLAM 1/2/3 are described in the Experiments/GLAM-CLUE section. CLUE 1/2 are generated from $\lambda_x=0$ and $\lambda_x=0.03$ respectively.}
    \label{fig:GLAMvsBaselines}
\end{figure*}

\subsection{Proposed Method: GLAM-CLUE}
\label{GLAM-CLUE}

GLAM-CLUE takes groups of high/low certainty points and learns mappings of arbitrary complexity between them in latent space (\textbf{training step}). Mappers are then applied to generate CEs from uncertain inputs (\textbf{inference step}). It can be seen as a global equivalent to CLUE. Initially, inputs are taken from the training data to learn such mappings, but we demonstrate that we can make improvements by instead using CLUEs generated from uncertain points in the training data.
%Additionally, these $\nabla$-CLUEs could be used to improve performance relating to diversity within GLAM-CLUE.
%In applications where uncertain training data is lacking, we can instead partition the test set (which contains the uncertain points requiring explanations) into two groups: a validation set from which we learn the GLAM-CLUE mappings and a held out test set to measure their performance. 
Algorithm \ref{algorithm:GLAM-CLUE} defines a mapper of arbitrary complexity from uncertain groups to certain groups in latent space: $\rvz_\text{certain}=G(\rvz_\text{uncertain})$. These mappers have parameters $\pmb{\theta}$.

\begin{figure}[ht]
\centering
\includegraphics[width=0.45\textwidth]{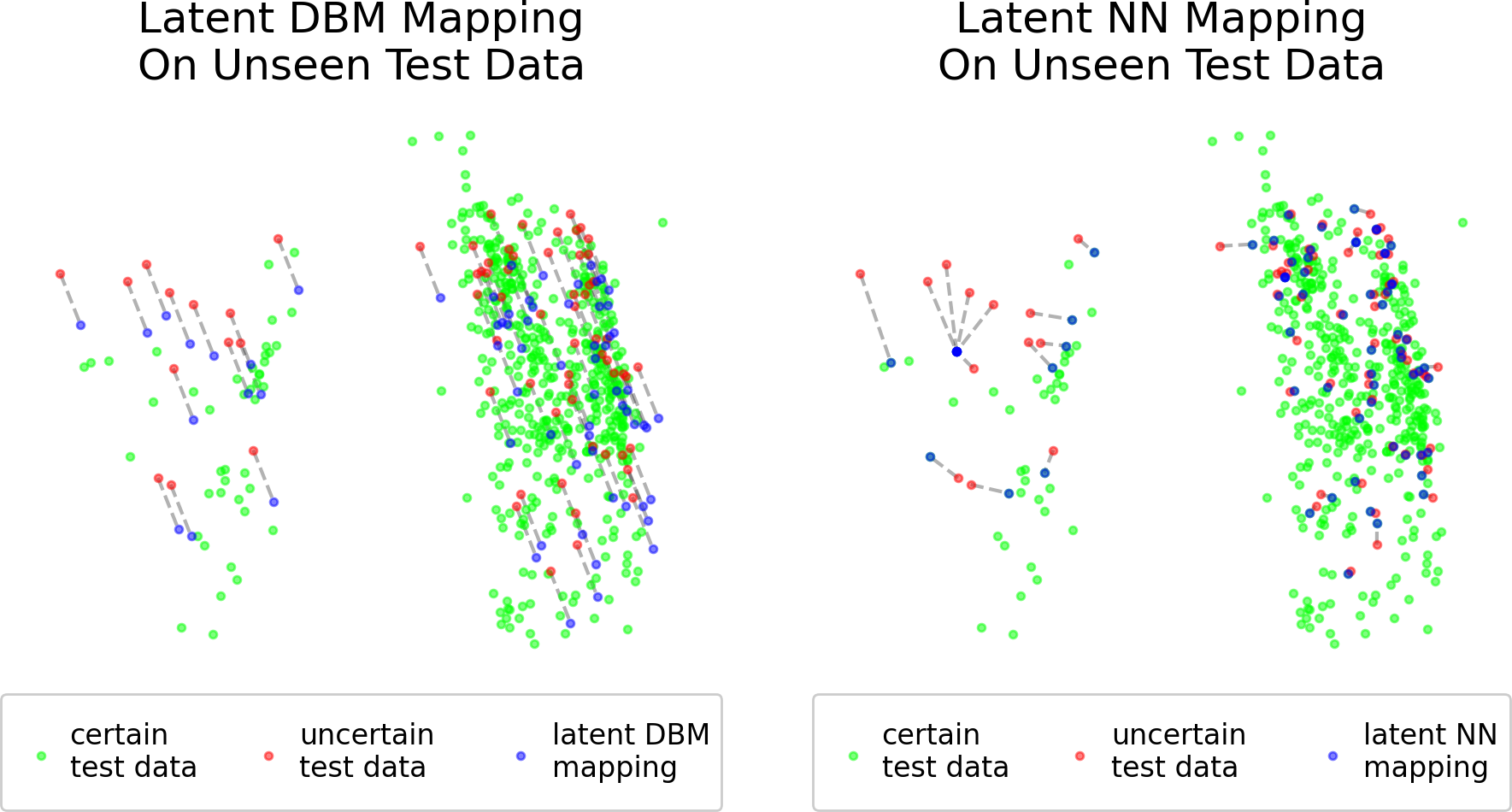}
\caption{\small Visualisation of the DBM and NN baselines for MNIST's digit 4 in a 2D latent space. Left: Uncertain points in the test data with their respective latent DBM mappings. Right: Uncertain points in the test data with their respective NN mappings. High certainty training data is shown in green throughout.}
\label{fig:GLAM-CLUEbaselines}
\end{figure}

\begin{algorithm}[t]
\small
 \INPUT {\footnotesize Inputs $X_{\text{uncertain}}, X_{\text{certain}}$, groups $Y_{\text{uncertain}}, Y_{\text{certain}}$, DGM encoder $\mu_\phi$, loss $\mathcal{L}$, trainable parameters $\pmb{\theta}$}
 \begin{algorithmic}[1]
 \FORALL{groups $(i\rightarrow j$) in $(Y_{\text{uncertain}}$, $Y_{\text{certain}}$)}
     \STATE Select $X_{i}$ from $X_\text{uncertain},Y_\text{uncertain}$;
     \STATE Select $X_{j}$ from $X_\text{certain},Y_\text{certain}$;
     \STATE Encode: $Z_{i}=\mu_\phi(Z|X_{i})$;
     \WHILE{\textit{loss $\mathcal{L}$ has not converged}}
        \STATE Update { $\pmb{\theta}_{i\rightarrow j}$} with {$\nabla_{\pmb{\theta}_{i\rightarrow j}}\mathcal{L}(\pmb{\theta}_{i\rightarrow j}|Z_{i}, X_{j})$;}
     \ENDWHILE
 \ENDFOR
\end{algorithmic}
\OUTPUT A collection of mapping parameters $\pmb{\theta}_{i\rightarrow j}$ for given mappers $G_{i\rightarrow j}$ that take uncertain inputs from group $i$ and produce nearby certain outputs in group $j$
\caption{GLAM-CLUE (Training Step)}
\label{algorithm:GLAM-CLUE}
\end{algorithm}
To strive for global explanations, we restrict each mapper in our experiments to be a single latent translation from an uncertain class $i$ to a certain class $j$: $\rvz_j=G_{i\rightarrow j}(\rvz_i)=\rvz_i+\pmb{\theta}_{i\rightarrow j}$. When run on test data, mappers should reduce the uncertainty of points while keeping them close to the original. To train the parameters of the translation $\pmb{\theta}$, we use the loss function detailed in Equation~2, similar to \citet{vanlooveren2020interpretable}, who inspect the $k$ nearest data points (our min operation implies $k=1$). We infer from Figure \ref{fig:diversityMNISTdeltaCLUE}, right, that regularisation in latent space implies regularisation in input space. We learn separate mappers for each pair of groups defined by the practitioner (Figure \ref{fig:GLAM-CLUEmappings}); Algorithm~\ref{algorithm:GLAM-CLUE} loops over these groups, partitioning the data accordingly, and returning distinct parameters $\pmb{\theta}_{i\rightarrow j}$ for each case.

\[\mathcal{L}(\pmb{\theta}|Z_{\text{uncertain}}, X_{\text{certain}})=\]
\[\lambda_\theta\lVert\pmb{\theta}\rVert_1+\frac{1}{|Z_\text{uncertain}|}\sum_{\rvz\in Z_\text{uncertain}}\min_{\rvx\in X_\text{certain}}\lVert\mu_\theta(\rvz+\pmb{\theta})-\rvx\rVert_2^2\ (2)\]
Few works in the counterfactual literature address uncertainty explanations; we avoid the comparison with state-of-the-art counterfactual methods for the reasons discussed in the introduction. However, there exist multiple standard baselines against which we can test performance.
Firstly, we can perform Difference Between Means (DBM) of uncertain data to certain data in either input or latent space. This can be added to uncertain test data and reconstructed in the case of input space, or decoded in the case of latent space. Another baseline is the Nearest Neighbours (NN) in high certainty training data, in either input or latent space. Figure~\ref{fig:GLAM-CLUEbaselines} visualises these baselines in latent space. Our experiments demonstrate that GLAM-CLUE outperforms these baselines significantly, and performs on par with CLUE. \citet{pawelczyk2021carla} create a benchmarking tool which shows that CLUE performs on par with the current state-of-the-art. By extension, so too does our scheme, but 200 times faster.

\noindent When the class of uncertain test data is unknown, mappings could be applied over each combination of classes, picking the best performing CEs. When the number of classes is large, a scheme to select a limited number of these (e.g. the top $n$ predictions from the classifier) could be used.%, or the $n$ most common classes returned by $\nabla$-CLUE if it was used in training
%the latter is likely both more reliable and more expensive.
Generic mappings from uncertainty to certainty would not require this selection but on the whole would be harder to train (simple translations are likely invalid for the far right case of Figure \ref{fig:GLAM-CLUEmappings}). We posit that more complex models such as neural networks could improve the performance of mappings at the risk of losing the global sense of the explanation.
\subsection{Grouping Uncertainty}
Most counterfactual explanation techniques center around determining ways to change the class label of a prediction; for example, Transitive Global Translations (TGTs) consider each possible combination of classes and the mappings between them~\citep{plumb2020explaining}. We choose here to partition the data into classes, but also into certain and uncertain groups according to the classifier used. By using these partitions, we learn mappings from uncertain points to certain points, either within specific classes or in the general case.
While TGTs constrain a mapping $G$ from group $i$ to $j$ to be symmetric ($G_{i\rightarrow j}=G_{j\rightarrow i}^{-1}$) and transitive ($G_{i\rightarrow k}=G_{j\rightarrow k}\circ G_{i\rightarrow j}$), we see no direct need for the symmetry constraint. There exists an infinitely large domain of uncertain points, unlike the bounded domain for certain points, implying a many-to-one mapping. We also forgo the transitivity constraint: defining direct mappings from uncertain points to specific certain points is sufficient.%, although the effectiveness of the latter is measured (check this last part, might not happen).
\begin{figure}[ht]
    \centering
    \includegraphics[width=0.43\textwidth]{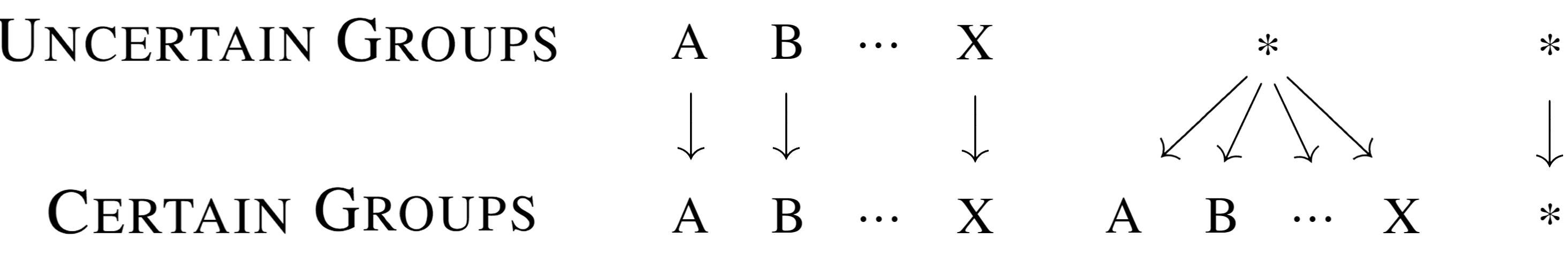}
    \caption{\small Example mappings from uncertainty to certainty in groups A to X, without necessarily satisfying symmetry or transitivity. Asterisks represent members belonging to any group.}
    \label{fig:GLAM-CLUEmappings}
\end{figure}

\noindent Our method is general to all schemes (and more) in Figure~\ref{fig:GLAM-CLUEmappings}. Our experiments consider these groups to be class labels, testing against the far left scheme which considers mapping from uncertain points to certain points within a given class. Future work may consider modes within classes, as well as the more general far right scheme of learning mappings from arbitrary uncertain inputs to their certain analogues. The original CLUE method is analogous to the far right scheme, which is agnostic to the particular classes it maps to and from (although struggles with diverse mappings).

\section{Experiments}
We perform experiments on 3 datasets to validate our methods: UCI Credit classification \citep{Dua:2019}, MNIST image classification \citep{lecun1998mnist} and Synbols image classification \citep{lacoste2020synbols}. On Credit and MNIST, we train VAEs as our DGMs \citep{kingma2014autoencoding} and BNNs for classification \citep{mackay1992practical}. For Synbols, we train Hierarchical VAEs \citep{pmlr-v70-zhao17c} and a resnet deep ensemble, owing to higher dataset complexity (rotations, sizes and obscurity of shapes).
We demonstrate that our constraints allow practitioners to better control the uncertainty-distance trade-off of CEs ($\delta$-CLUE) and the diversity of CEs ($\nabla$-CLUE). We then show that we can efficiently generate explanations that apply globally to groups of inputs with our amortised scheme (GLAM-CLUE).

\subsection{$\delta$-CLUE}
\begin{figure}
\centering
\includegraphics[width=0.45\textwidth]{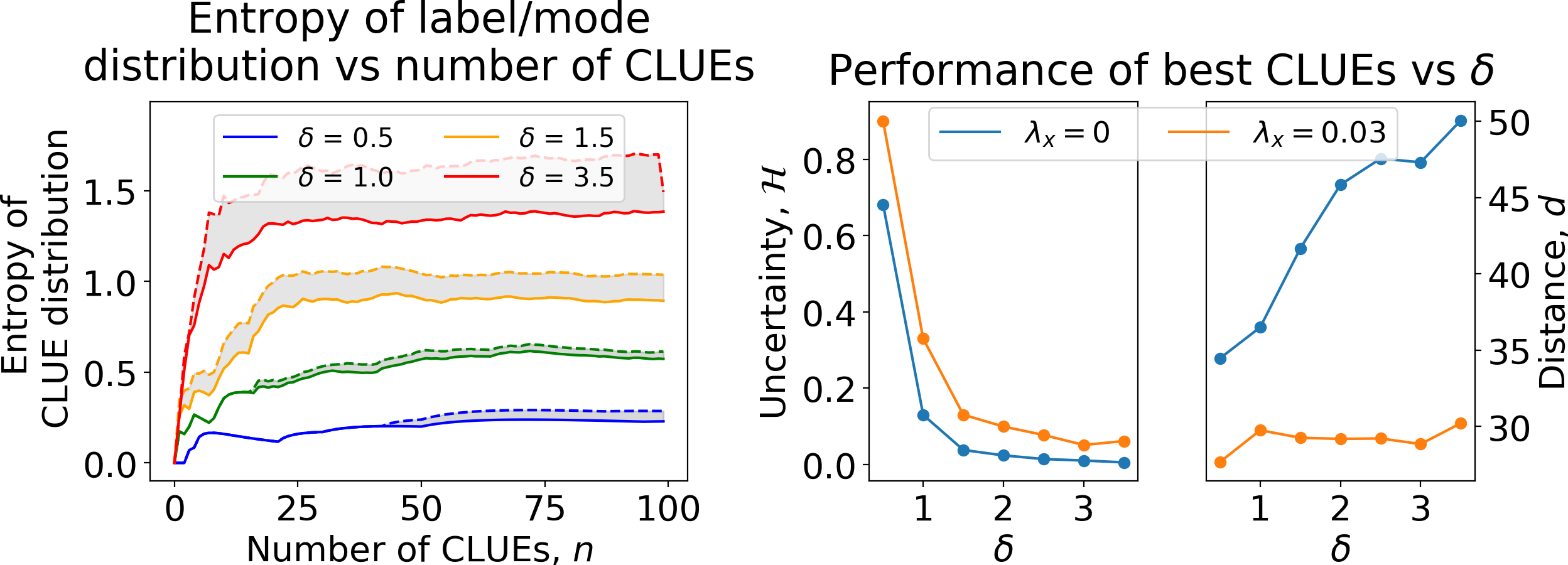}
\caption{\small Left: Diversity analysis in MNIST. Entropy of the distribution of class labels (solid) and modes (dashed) found as number of CLUEs increases. Labels vary from 0 to 9 whilst there exist multiple modes within each label. Observe the entropy saturating as we converge to all minima within the $\delta$ ball. Right: Performance of $\delta$-CLUEs (uncertainty, $\mathcal{H}$, input space distance, $d$). Batch size: 8 most uncertain MNIST digits. Learning rate: 0.1. Iterations: 30.}
\label{fig:diversityMNISTdeltaCLUE}
\end{figure}

We learn from the $\delta$-CLUE experiments that the $\delta$ value controls the trade-off between the uncertainty of the CLUEs generated and their distance from the original point (Figure~\ref{fig:synbolsmnist}).
Importantly, by tuning $\lambda_x$ in the distance term $d$ of Equation \ref{eq:1}, we achieve lower distances with only small uncertainty increases (Figure \ref{fig:diversityMNISTdeltaCLUE}, right). We observe further in Figure \ref{fig:diversityMNISTdeltaCLUE}, left that diversity increases with $\delta$, although a large number of CLUEs can be required before such levels become saturated (left). Modes are defined as groups of points within specific classes. Full analysis in Appendix B. %\ref{appendix:deltaCLUE}.

\textbf{Takeaway:} $\delta$-CLUE produces a high performing set of diverse explanations. However, we require many iterations to achieve such diversity ($\nabla$-CLUE addresses this).

\subsection{$\nabla$-CLUE}
We perform an ablative study, increasing the diversity weight $\lambda_D$ and optimising the DPP diversity metric in $\rvz$-space, measuring the effect that this has on each other metric. We use the simultaneous $\nabla$-CLUE scheme in Algorithm \ref{algorithm:divCLUEsimultaneous} for a fixed number of $k=10$ CLUEs and parameters: $\delta=r=4$ for MNIST; $\delta=r=1$ for UCI Credit. The optimal $\delta$ value(s) can be determined through experimentation (Figure \ref{fig:diversityMNISTdeltaCLUE}, right), although Appendix B %\ref{appendix:deltaCLUE}
discusses alternative methods such as inspecting nearest neighbours in the data. %(where $m$ is the dimensionality of latent space).
%Note that $\lambda_D=0$ is exactly equivalent to $\delta$-CLUE.
%Appendix \ref{appendix:divCLUE} details further experimental analysis.
\begin{figure}[hb]
    \centering
    \includegraphics[scale=0.24]{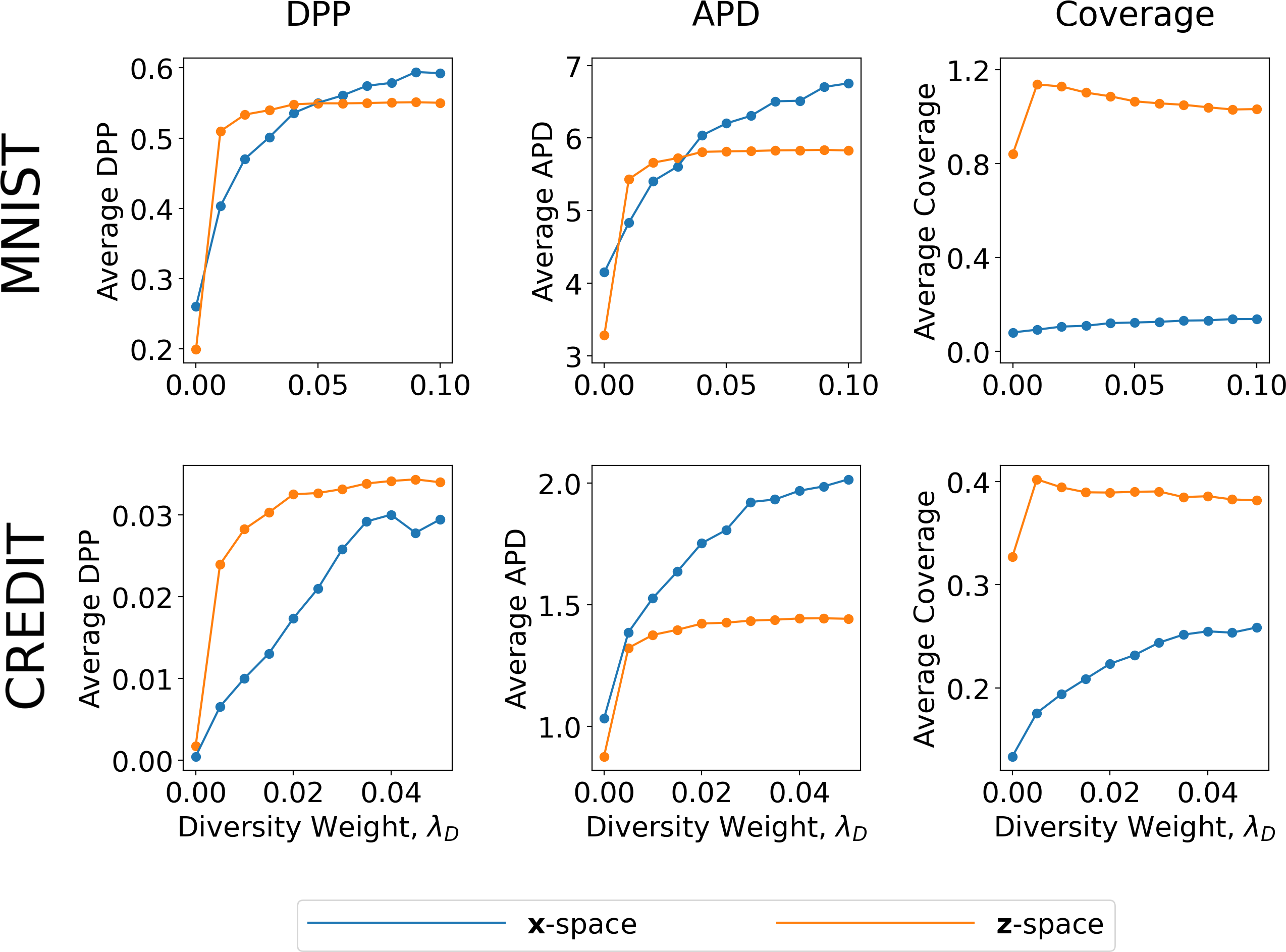}
    \caption{\small Effect of $\lambda_D$ on diversity. Row 1: MNIST. Row 2: UCI Credit. Columns 1 to 3: DPP, APD and Coverage diversity metrics applied to the set of $k=10$ $\nabla$-CLUEs. $\lambda_D=0$ is $\delta$-CLUE. Batch size: 8 most uncertain inputs. Learning rate: 0.1. Iterations: 30.}
    
    % Effect of $\lambda_D$ on diversity and performance. Row 1: MNIST. Row 2: UCI Credit. Columns 1 to 3: DPP, APD and Coverage diversity metrics applied to the set of $k=10$ $\nabla$-CLUEs in either input space or latent space. Column 4: Diversities in prediction space (distinct labels and entropy of labels). Column 5: Performance of $\nabla$-CLUEs (average uncertainty and average distance compared to original uncertainty).}
    \label{fig:divCLUEsim}
\end{figure}

\noindent\textbf{Takeaway:} When optimising for one diversity metric, increasing $\lambda_D$ monotonically improves diversity by almost every other metric. Average uncertainty suffers only a small amount relative to the gains we achieve in diversity and $\nabla$-CLUE requires fewer counterfactuals to achieve the same level of diversity as $\delta$-CLUE.

\subsection{GLAM-CLUE}

Gradient descent at the inference step (generation of CEs) is computationally expensive. Uncertainty estimates, distance metrics, and diversity metrics (notably DPPs, which operate on $k\times k$ matrices) all require evaluation over many iterations, to yield only a single counterfactual to a local uncertain input. While local explanations have utility in certain settings, GLAM-CLUE computes CEs for all uncertain test points in a single, amortised function call, permitting considerable speedups. We demonstrate that the performance of these counterfactual beats mean performance of all the baselines discussed, achieving lower variance also.

\begin{figure*}[ht]%{L}{0.6\textwidth}
    \centering
    \includegraphics[width=0.85\textwidth]{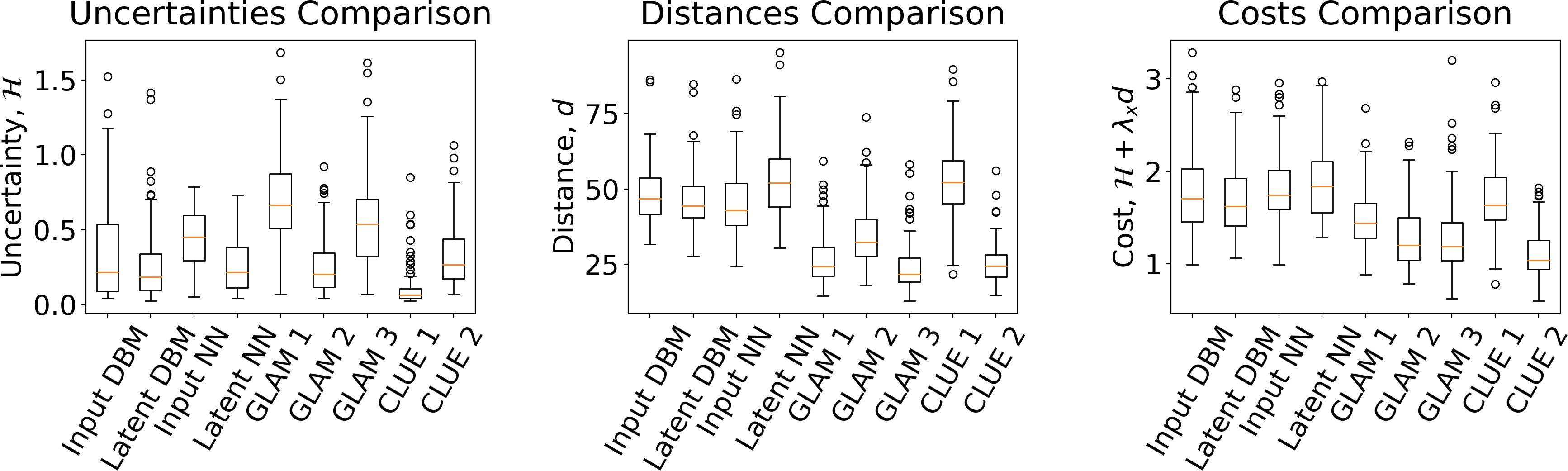}
    \caption{\small GLAM-CLUE schemes vs baselines when mapping uncertain 4s to certain 4s in MNIST. Left: Distributions of uncertainties, $\mathcal{H}$ (original uncertainties exceed 1.5). Centre: Distributions of input distances, $d$. Right: Distributions of total costs, $\mathcal{H}+\lambda_xd$, with $\lambda_x=0.03$ as used by \citet{antoran2021getting}. Similar results for all classes (Appendix E). CLUE 1/CLUE 2 are generated from $\lambda_x=0$ and $\lambda_x=0.03$ respectively. Batch size: 6000 (all 4s in training set. Learning rate: 0.1. Iterations: 30. Multiple random seed runs yield negligible differences.}
    \label{fig:GLAMCLUE}
\end{figure*}

We train 3 mappers: GLAM 1 learns from all certain and uncertain 4s in the MNIST training data; GLAM 2/3 learn from all uncertain 4s in the training data and their corresponding certain CLUEs, for $\lambda_x=0$ and $\lambda_x=0.03$ respectively. Figure \ref{fig:GLAMCLUE} shows improvements when using GLAM 2 and 3, demonstrating that CLUEs capture properties of uncertainty more reliably than the training data, at the expense of extra computation time to generate the CLUEs used.

%Note that GLAM-CLUE mappings can be applied from any uncertain class to any certain class. In order to compare them alongside the VAE reconstruction, we simply test mappings that move from the uncertain domain of the true class to the certain domain of the same true class for each unseen test point. This concept is illustrated in Figure \ref{fig:GLAM-CLUEbaselines}, which maps from uncertain 4s to certain 4s in MNIST.

We observe that while the baseline schemes achieve low uncertainties, they do so at the expense of moving much further away from the input (Figure~\ref{fig:GLAMvsBaselines}), implying infeasible actionability. An advantage to GLAM-CLUE is that the uncertainty-distance trade-off can be tuned with $\lambda_\theta$ in Equation 2: larger $\lambda_\theta$ restricts translations in latent space, thus lowering distances in input space but raising uncertainties. For a given $\lambda_x$, GLAM-CLUE's fast learning rate allows for the optimal $\lambda_\theta$ to be determined quickly. Furthermore, 98\% of uncertain 4 to certain 4 GLAM-CLUE mappings resulted in a classification of 4 (87\% for CLUE which simply minimises uncertainty and is not class specific).

\textbf{Takeaway:} Amortisation of counterfactuals works. A simple global translation for class specific points is shown to produce counterfactuals of comparable quality to CLUE. Notably, performance of GLAM-CLUE is improved when training on CLUEs rather than training data, optimal when we generate CLUEs using $\lambda_x=0.03$, as used in evaluation.

\subsection{Computational Speedup}
\begin{table}[b]
\centering
\begin{tabular}{|c|c|c|}
\bottomrule
Input DBM & Latent DBM & Input NN\\
\hline
0.0306 & 0.0262 & 0.0236\\
\toprule
\bottomrule
Latent NN & GLAM-CLUE & CLUE\\
\hline
0.0245 & 0.0238 & 4.68\\
\toprule
\end{tabular}
\caption{\small Avg. time in seconds for 1 MNIST CE (inference step). We achieve similar speedups (186 times faster) for UCI Credit.}
\label{tab:GLAMspeedups}
\end{table}

At the inference step, GLAM-CLUE performs significantly faster than CLUE in terms of \textbf{average CPU time}, detailed in Table \ref{tab:GLAMspeedups}. For uncertain 4s in the MNIST test set, CLUE required on average 220 seconds to converge; GLAM-CLUE took around 1 second to compute. The bottleneck in these processes is the uncertainty evaluation of the BNN, and as such these timings are not necessarily representative of all models. A drawback to GLAM-CLUE is that the optimisation required on average 17.6 seconds to train. Should CLUEs be included during training (i.e. GLAM 2 and 3), extra time is required to obtain these. Moving beyond basic mappers to more advanced models, we expect performance to improve at the cost of an increased training step time.

\textbf{Takeaway:} GLAM-CLUE produces explanations around 200 times faster than CLUE. This speedup, alongside the baselines, means that we have the option to take the best performing counterfactual out of GLAM-CLUE and the baselines, without requiring significant computation.

\section{Related and Future Work}
% \paragraph{Explanations of Uncertainty}:
The majority of this paper is dedicated to increasing the practical utility of the uncertainty explanations proposed as CLUE in \citet{antoran2021getting}, and we mitigate CLUE's multiplicity and efficiency issues. Very few works address explaining the uncertainty of probabilistic models. \citet{booth2020bayes} take a user-specified level of uncertainty for a sample in an auxiliary discriminative model and generate the corresponding sampling using deep generative models (DGM). \citet{joshi2018xgems} propose xGEMs that use a DGM to find CEs (as we do) though not for uncertainty.
\citet{Mothilal_2020} and \citet{russell2019efficient} use linear programs to find a diverse set of CEs, though also not for uncertainty. Neither paper considers computational advances nor ventures to consider global CEs, as we do.
\citet{plumb2020explaining} define a mapper that transforms points from one low-dimensional group to another. \citet{mahajan2020preserving} and \citet{yang_2021} redesign DGMs to generate CEs quickly, similar to GLAM-CLUE. In spirit of such works, we propose amortising CLUE to find a transformation that leads the model to treat the transformed uncertain points from Group A as certain points from Group B. This method could extend beyond CLUE to other classes of CEs.

Future explorations include higher dimensional datasets such as CIFAR10 \citep{krizhevsky2012} and CelebA \citep{liu2015faceattributes} that would fully test CLUE and the extensions proposed in this paper, potentially requiring the use of FID scores \citep{heusel2018gans} to replace the simple distance metric in both evaluation \citep{singla2020explanation} and optimisation. DGM alternatives such as GANs \citep{goodfellow2014generative} could be explored therein. Further, since \citet{antoran2021getting} demonstrate success on human subjects in the use of DGMs for counterfactuals, our reasoning is that we can hope to retain this efficacy with our extensions of CLUE, though ideally additional human experiments would further validate our methods. Multiple runs at various random seeds would also shed light on the sensitivity of the $\nabla$-CLUE algorithm.

\section{Conclusion}
Explanations from machine learning systems are receiving increasing attention from practitioners and industry~\citep{bhatt2020explainable}. As these systems are deployed in high stakes settings, well-calibrated uncertainty estimates are in high demand~\citep{spiegelhalter2017risk}. For a method to interpret uncertainty estimates from differentiable probabilistic models, \citet{antoran2021getting} propose generating a Counterfactual Latent Uncertainty Explanation (CLUE) for a given data point on which the model is uncertain. In this work, we examine how to make CLUEs more useful in practice. We devise $\delta$-CLUE, a method to generate a set of potential CLUEs within a $\delta$ ball of the original input in latent space, before proposing DIVerse CLUE ($\nabla$-CLUE), a method to find a set of CLUEs in which each proposes a distinct explanation for how to decrease the uncertainty associated with an input (to tackle the redundancy within $\delta$-CLUE). However, these methods prove to be potentially computationally inefficient for large amounts of data. To that end, we propose GLobal AMortised CLUE (GLAM-CLUE), which learns an amortised mapping that applies to specific groups of uncertain inputs. GLAM-CLUE efficiently transforms an uncertain input in a single function call into an input that a model will be certain about.
We validate our methods with experiments, which show that $\delta$-CLUE, $\nabla$-CLUE, and GLAM-CLUE address shortcomings of CLUE. We hope our proposed methods prove beneficial to practitioners who seek to provide explanations of uncertainty estimates to stakeholders.

%\section*{Ethical Considerations}
%Placeholder text.

\section*{Acknowledgments}
UB acknowledges support from DeepMind and the Leverhulme Trust via the Leverhulme Centre for the Future of Intelligence (CFI) and from the Mozilla Foundation.
AW acknowledges support from a Turing AI Fellowship under grant EP/V025379/1, The Alan Turing Institute, and the Leverhulme Trust via CFI. The authors thank Javier Antor\'{a}n for his helpful comments and pointers.

\bibliography{ref}

\begin{thebibliography}{39}
\providecommand{\natexlab}[1]{#1}

\bibitem[{Adebayo et~al.(2020)Adebayo, Muelly, Liccardi, and
  Kim}]{adebayo2020debugging}
Adebayo, J.; Muelly, M.; Liccardi, I.; and Kim, B. 2020.
\newblock Debugging Tests for Model Explanations.
\newblock In \emph{Advances in Neural Information Processing Systems}.

\bibitem[{Antor\'{a}n et~al.(2021)Antor\'{a}n, Bhatt, Adel, Weller, and
  Hern{\'a}ndez-Lobato}]{antoran2021getting}
Antor\'{a}n, J.; Bhatt, U.; Adel, T.; Weller, A.; and Hern{\'a}ndez-Lobato,
  J.~M. 2021.
\newblock Getting a {CLUE}: A Method for Explaining Uncertainty Estimates.
\newblock In \emph{International Conference on Learning Representations}.

\bibitem[{Bhatt et~al.(2021)Bhatt, Chien, Zafar, and Weller}]{bhatt2021divine}
Bhatt, U.; Chien, I.; Zafar, M.~B.; and Weller, A. 2021.
\newblock DIVINE: Diverse Influential Training Points for Data Visualization
  and Model Refinement.
\newblock arXiv:2107.05978.

\bibitem[{Bhatt et~al.(2020)Bhatt, Xiang, Sharma, Weller, Taly, Jia, Ghosh,
  Puri, Moura, and Eckersley}]{bhatt2020explainable}
Bhatt, U.; Xiang, A.; Sharma, S.; Weller, A.; Taly, A.; Jia, Y.; Ghosh, J.;
  Puri, R.; Moura, J.~M.; and Eckersley, P. 2020.
\newblock Explainable Machine Learning in Deployment.
\newblock In \emph{Proceedings of the 2020 Conference on Fairness,
  Accountability, and Transparency}, 648--657.

\bibitem[{Booth et~al.(2020)Booth, Zhou, Shah, and Shah}]{booth2020bayes}
Booth, S.; Zhou, Y.; Shah, A.; and Shah, J. 2020.
\newblock Bayes-TrEx: Model Transparency by Example.
\newblock In \emph{Thirty-Fifth AAAI Conference on Artificial Intelligence}.

\bibitem[{Boyd, Boyd, and Vandenberghe(2004)}]{boyd2004convex}
Boyd, S.; Boyd, S.~P.; and Vandenberghe, L. 2004.
\newblock \emph{Convex Optimization}.
\newblock Cambridge university press.

\bibitem[{Depeweg et~al.(2017)Depeweg, Hernández-Lobato, Doshi-Velez, and
  Udluft}]{depeweg2017uncertainty}
Depeweg, S.; Hernández-Lobato, J.~M.; Doshi-Velez, F.; and Udluft, S. 2017.
\newblock Uncertainty Decomposition in Bayesian Neural Networks with Latent
  Variables.
\newblock arXiv:1706.08495.

\bibitem[{Dosovitskiy and Djolonga(2020)}]{Dosovitskiy2020You}
Dosovitskiy, A.; and Djolonga, J. 2020.
\newblock You Only Train Once: Loss-Conditional Training of Deep Networks.
\newblock In \emph{International Conference on Learning Representations}.

\bibitem[{Dua and Graff(2017)}]{Dua:2019}
Dua, D.; and Graff, C. 2017.
\newblock {UCI} Machine Learning Repository.

\bibitem[{Goodfellow et~al.(2014)Goodfellow, Pouget-Abadie, Mirza, Xu,
  Warde-Farley, Ozair, Courville, and Bengio}]{goodfellow2014generative}
Goodfellow, I.~J.; Pouget-Abadie, J.; Mirza, M.; Xu, B.; Warde-Farley, D.;
  Ozair, S.; Courville, A.; and Bengio, Y. 2014.
\newblock Generative Adversarial Networks.
\newblock In \emph{Advances in Neural Information Processing Systems}.

\bibitem[{Goodfellow, Shlens, and Szegedy(2015)}]{goodfellow2014explaining}
Goodfellow, I.~J.; Shlens, J.; and Szegedy, C. 2015.
\newblock Explaining and Harnessing Adversarial Examples.
\newblock In \emph{International Conference on Learning Representations}.

\bibitem[{Grover and Toghi(2019)}]{grover2019mnist}
Grover, D.; and Toghi, B. 2019.
\newblock {MNIST} dataset classification utilizing k-NN classifier with
  modified sliding-window metric.
\newblock In \emph{Science and Information Conference}, 583--591. Springer.

\bibitem[{Heusel et~al.(2018)Heusel, Ramsauer, Unterthiner, Nessler, and
  Hochreiter}]{heusel2018gans}
Heusel, M.; Ramsauer, H.; Unterthiner, T.; Nessler, B.; and Hochreiter, S.
  2018.
\newblock GANs Trained by a Two Time-Scale Update Rule Converge to a Local Nash
  Equilibrium.
\newblock In \emph{Advances in Neural Information Processing Systems}.

\bibitem[{Joshi et~al.(2018)Joshi, Koyejo, Kim, and Ghosh}]{joshi2018xgems}
Joshi, S.; Koyejo, O.; Kim, B.; and Ghosh, J. 2018.
\newblock {xGEMs}: Generating Examplars to Explain Black-Box Models.
\newblock \emph{arXiv preprint arXiv:1806.08867}.

\bibitem[{Kingma and Welling(2013)}]{kingma2014autoencoding}
Kingma, D.~P.; and Welling, M. 2013.
\newblock Auto-Encoding Variational Bayes.
\newblock In \emph{International Conference on Learning Representations}.

\bibitem[{Krizhevsky(2012)}]{krizhevsky2012}
Krizhevsky, A. 2012.
\newblock Learning Multiple Layers of Features from Tiny Images.
\newblock \emph{University of Toronto}.

\bibitem[{Kulesza(2012)}]{Kulesza_2012}
Kulesza, A. 2012.
\newblock Determinantal Point Processes for Machine Learning.
\newblock \emph{Foundations and Trends® in Machine Learning}, 5(2-3):
  123–286.

\bibitem[{Lacoste et~al.(2020)Lacoste, Rodr{\'{\i}}guez, Branchaud{-}Charron,
  Atighehchian, Caccia, H.~Laradji, Drouin, Craddock, Charlin, and
  V{\'{a}}zquez}]{lacoste2020synbols}
Lacoste, A.; Rodr{\'{\i}}guez, P.; Branchaud{-}Charron, F.; Atighehchian, P.;
  Caccia, M.; H.~Laradji, I.; Drouin, A.; Craddock, M.; Charlin, L.; and
  V{\'{a}}zquez, D. 2020.
\newblock Synbols: Probing Learning Algorithms with Synthetic Datasets.
\newblock In \emph{Advances in Neural Information Processing Systems}.

\bibitem[{LeCun(1998)}]{lecun1998mnist}
LeCun, Y. 1998.
\newblock The MNIST Database of Handwritten Digits.
\newblock \emph{http://yann. lecun. com/exdb/mnist/}.

\bibitem[{Ley, Bhatt, and Weller(2021)}]{ley2021deltaclue}
Ley, D.; Bhatt, U.; and Weller, A. 2021.
\newblock {$\delta$-CLUE: Diverse} Sets of Explanations for Uncertainty
  Estimates.
\newblock In \emph{ICLR Workshop on Security and Safety in Machine Learning
  Systems}.

\bibitem[{Liu et~al.(2015)Liu, Luo, Wang, and Tang}]{liu2015faceattributes}
Liu, Z.; Luo, P.; Wang, X.; and Tang, X. 2015.
\newblock Deep Learning Face Attributes in the Wild.
\newblock In \emph{Proceedings of International Conference on Computer Vision
  (ICCV)}.

\bibitem[{MacKay(1992)}]{mackay1992practical}
MacKay, D.~J. 1992.
\newblock A Practical Bayesian Framework for Backpropagation Networks.
\newblock \emph{Neural computation}, 4(3): 448--472.

\bibitem[{Mahajan, Tan, and Sharma(2020)}]{mahajan2020preserving}
Mahajan, D.; Tan, C.; and Sharma, A. 2020.
\newblock Preserving Causal Constraints in Counterfactual Explanations for
  Machine Learning Classifiers.
\newblock In \emph{NeurIPS Workshop on CausalML: Machine Learning and Causal
  Inference for Improved Decision Making}.

\bibitem[{Mothilal, Sharma, and Tan(2020)}]{Mothilal_2020}
Mothilal, R.~K.; Sharma, A.; and Tan, C. 2020.
\newblock Explaining machine learning classifiers through diverse
  counterfactual explanations.
\newblock \emph{Proceedings of the 2020 Conference on Fairness, Accountability,
  and Transparency}.

\bibitem[{Pawelczyk et~al.(2021)Pawelczyk, Bielawski, van~den Heuvel, Richter,
  and Kasneci}]{pawelczyk2021carla}
Pawelczyk, M.; Bielawski, S.; van~den Heuvel, J.; Richter, T.; and Kasneci, G.
  2021.
\newblock CARLA: A Python Library to Benchmark Algorithmic Recourse and
  Counterfactual Explanation Algorithms.
\newblock In \emph{Advances in Neural Information Processing Systems (Benchmark
  \& Data Set Track)}.

\bibitem[{Pawelczyk, Broelemann, and
  Kasneci(2020{\natexlab{a}})}]{pawelczyk2020learning}
Pawelczyk, M.; Broelemann, K.; and Kasneci, G. 2020{\natexlab{a}}.
\newblock Learning Model-Agnostic Counterfactual Explanations for Tabular Data.
\newblock In \emph{Proceedings of The Web Conference 2020}, 3126--3132.

\bibitem[{Pawelczyk, Broelemann, and
  Kasneci(2020{\natexlab{b}})}]{pawelczyk2020counterfactual}
Pawelczyk, M.; Broelemann, K.; and Kasneci, G. 2020{\natexlab{b}}.
\newblock On Counterfactual Explanations under Predictive Multiplicity.
\newblock In \emph{Conference on Uncertainty in Artificial Intelligence},
  809--818. PMLR.

\bibitem[{Plumb et~al.(2020)Plumb, Terhorst, Sankararaman, and
  Talwalkar}]{plumb2020explaining}
Plumb, G.; Terhorst, J.; Sankararaman, S.; and Talwalkar, A. 2020.
\newblock Explaining Groups of Points in Low-Dimensional Representations.
\newblock In \emph{International Conference on Machine Learning}, 7762--7771.
  PMLR.

\bibitem[{Poyiadzi et~al.(2020)Poyiadzi, Sokol, Santos-Rodriguez, De~Bie, and
  Flach}]{poyiadzi2020face}
Poyiadzi, R.; Sokol, K.; Santos-Rodriguez, R.; De~Bie, T.; and Flach, P. 2020.
\newblock FACE: Feasible and Actionable Counterfactual Explanations.
\newblock In \emph{Proceedings of the AAAI/ACM Conference on AI, Ethics, and
  Society}, 344--350.

\bibitem[{Ribeiro, Singh, and Guestrin(2016)}]{ribeiro2016why}
Ribeiro, M.~T.; Singh, S.; and Guestrin, C. 2016.
\newblock Why Should I Trust You?: Explaining the Predictions of any
  Classifier.
\newblock In \emph{Proceedings of the 22nd {ACM} {SIGKDD} {I}nternational
  {C}onference on {K}nowledge {D}iscovery and {D}ata {M}ining}, 1135--1144.
  ACM.

\bibitem[{Russell(2019)}]{russell2019efficient}
Russell, C. 2019.
\newblock Efficient Search for Diverse Coherent Explanations.
\newblock In \emph{Proceedings of the Conference on Fairness, Accountability,
  and Transparency}, 20--28.

\bibitem[{Singla et~al.(2020)Singla, Pollack, Chen, and
  Batmanghelich}]{singla2020explanation}
Singla, S.; Pollack, B.; Chen, J.; and Batmanghelich, K. 2020.
\newblock Explanation by Progressive Exaggeration.
\newblock In \emph{International Conference on Learning Representations}.

\bibitem[{Spiegelhalter(2017)}]{spiegelhalter2017risk}
Spiegelhalter, D. 2017.
\newblock Risk and Uncertainty Communication.
\newblock \emph{Annual Review of Statistics and Its Application}, 4: 31--60.

\bibitem[{Tsirtsis, De, and Gomez-Rodriguez(2021)}]{tsirtsis2021counterfactual}
Tsirtsis, S.; De, A.; and Gomez-Rodriguez, M. 2021.
\newblock Counterfactual Explanations in Sequential Decision Making Under
  Uncertainty.
\newblock In \emph{Advances in Neural Information Processing Systems}.

\bibitem[{Van~Looveren and Klaise(2021)}]{vanlooveren2020interpretable}
Van~Looveren, A.; and Klaise, J. 2021.
\newblock Interpretable Counterfactual Explanations Guided by Prototypes.
\newblock In \emph{Machine Learning and Knowledge Discovery in Databases.
  Research Track}, 650--665.

\bibitem[{Weinberger and Saul(2009)}]{weinberger2009distance}
Weinberger, K.~Q.; and Saul, L.~K. 2009.
\newblock Distance metric learning for large margin nearest neighbor
  classification.
\newblock \emph{Journal of machine learning research}, 10(2).

\bibitem[{Yang et~al.(2021)Yang, Alva, Chen, and Hu}]{yang_2021}
Yang, F.; Alva, S.~S.; Chen, J.; and Hu, X. 2021.
\newblock Model-Based Counterfactual Synthesizer for Interpretation.
\newblock \emph{Proceedings of the 27th ACM SIGKDD Conference on Knowledge
  Discovery \& Data Mining}.

\bibitem[{Zhang et~al.(2018)Zhang, Isola, Efros, Shechtman, and
  Wang}]{zhang2018unreasonable}
Zhang, R.; Isola, P.; Efros, A.~A.; Shechtman, E.; and Wang, O. 2018.
\newblock The unreasonable effectiveness of deep features as a perceptual
  metric.
\newblock In \emph{Proceedings of the IEEE conference on computer vision and
  pattern recognition}, 586--595.

\bibitem[{Zhao, Song, and Ermon(2017)}]{pmlr-v70-zhao17c}
Zhao, S.; Song, J.; and Ermon, S. 2017.
\newblock Learning Hierarchical Features from Deep Generative Models.
\newblock In Precup, D.; and Teh, Y.~W., eds., \emph{Proceedings of the 34th
  International Conference on Machine Learning}, volume~70 of \emph{Proceedings
  of Machine Learning Research}, 4091--4099. PMLR.

\end{thebibliography}

\appendix

\section*{Appendix}

This appendix is formatted as follows.

\begin{enumerate}[leftmargin=2.5\parindent]
    \item We discuss \textbf{datasets and models} in Appendix A.
    \item We provide a full analysis of the \textbf{$\pmb{\delta}$-CLUE} experiments and design choices in Appendix B.
    \item We discuss \textbf{diversity metrics for counterfactual explanations} further in Appendix C.
    \item We analyse \textbf{$\pmb{\nabla}$-CLUE} in Appendix D.
    \item We discuss \textbf{GLAM-CLUE} in Appendix E.
\end{enumerate}

Where necessary, we provide a discussion for potential limitations of our work and future improvements that could be studied.

\section{A\quad Datasets and Models}
\label{appendix:datasets}
One tabular dataset and two image datasets are employed in our experiments (all publicly available). Details are provided in Table~\ref{tab:datasets}.
\begin{table*}[ht]
\centering
\begin{tabular}{|c|c|c|c|c|c|}
\bottomrule
Name & Targets & Input Type & Input Dimension & No. Train & No. Test\\
\toprule
\bottomrule
Credit & Binary & Continuous \& Categorical & 24 & 27000 & 3000\\
\hline
MNIST & Categorical & Image (Greyscale) & $28\times28$ & 60000 & 10000\\
\hline
Synbols & Categorical & Image (RGB) & $3\times32\times32$ & 60000 & 20000\\
\toprule
\end{tabular}
\caption{\small Summary of the datasets used in our experiments.}
\label{tab:datasets}
\end{table*}

The default of credit card clients dataset, which we refer to as “Credit” in this paper, can be obtained from \url{https://archive.ics.uci.edu/ml/datasets/default+of+credit+card+clients/}. We augment input dimensions by performing a one-hot-encoding over necessary variables (i.e. gender, education). Note that this dataset is different from the also common German credit dataset.

The MNIST handwritten digit image dataset can be obtained from and is described in detail at \url{http://yann.lecun.com/exdb/mnist/}. For the aforementioned datasets, we thank \citet{antoran2021getting} for making their private BNN and VAE models available for use in our work.

For the experiments on Synbols, we use the black and white dataset, as provided by ElementAI at \url{https://github.com/ElementAI/synbols/}. Additional models (resnet classifiers, hierarchical VAEs) and loading scripts are taken from \url{https://github.com/ElementAI/synbols-benchmarks/}.

\section{B\quad $\delta$-CLUE}
\label{appendix:deltaCLUE}
We perform constrained optimisation during gradient descent (Figure \ref{fig:DeltaFinal}, centre). A later part of this Appendix (Constrained vs Unconstrained Search) provides justification for this decision. In our experiments, we search the latent space of a VAE to generate $\delta$-CLUEs for the $8$ most uncertain digits in the MNIST test set, according to our trained BNN.

\begin{figure}
\centering
\includegraphics[scale=0.48]{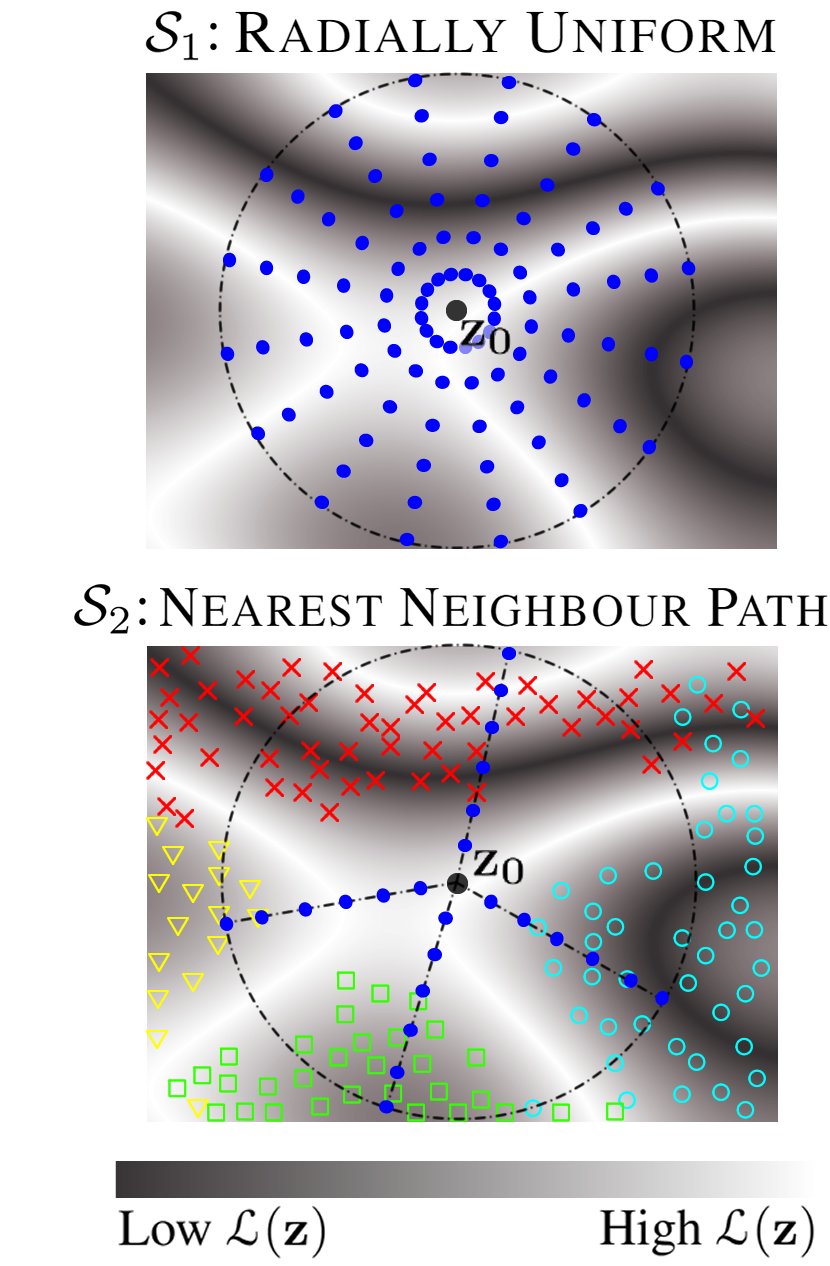}
\captionof{figure}{\small Two possible initialisation schemes $\mathcal{S}_i$ to yield diverse minima. One is random, the other deterministic. Details are provided in this Appendix.}
\label{fig:SearchStrats}
\end{figure}

We trial this over \textbf{a)} a range of several $\delta$ values from $0.5$ to $3.5$, \textbf{b)} two latent space loss functions: \textbf{Uncertainty} $\mathcal{L}_{\mathcal{H}}=\mathcal{H}$ and \textbf{Distance} $\mathcal{L}_{\mathcal{H}+d}=\mathcal{H}+d$ and \textbf{c)} two initialisation schemes as depicted in Figure~\ref{fig:SearchStrats}. Initialisation scheme $\mathcal{S}_1$ picks a random direction at a uniform random radius within the delta ball, while the other scheme $\mathcal{S}_2$ is along paths determined by the nearest neighbours (\textbf{NN}) for each class in the training data.  We label these experiment variants as: \textbf{Uncertainty Random}: [$\mathcal{L}_{\mathcal{H}}$, $\mathcal{S}_1$], \textbf{Uncertainty NN}: [$\mathcal{L}_{\mathcal{H}}$, $\mathcal{S}_2$], \textbf{Distance Random}: [$\mathcal{L}_{\mathcal{H}+d}$, $\mathcal{S}_1$] and \textbf{Distance NN}: [$\mathcal{L}_{\mathcal{H}+d}$, $\mathcal{S}_2$].

\begin{algorithm}[ht]
\newdimen\origiwspc
\origiwspc=\fontdimen2\font
\fontdimen2\font=100em
 \INPUT {\small $\delta$, $k$, $\mathcal{S}$, $r$, $\mathbf{x}_0$, $d$, $\rho$, $\mathcal{H}$, $\mu_\theta$, $\mu_\phi$}
 \fontdimen2\font=\origiwspc
 \begin{algorithmic}[1]
 \STATE Initialise $\varnothing$ of CLUEs: $X_{\mathrm{CLUE}}=\{\}$;
 \STATE Set $\delta$-ball centre of $\mathbf{z}_0=\mu_\phi(\mathbf{z}|\mathbf{x}_0)$;
 \FOR{$1\leq i\leq k$}
     \STATE Set initial value of $\mathbf{z}_i = \mathcal{S}(\mathbf{z}_0, r, i, k)$;
     %\\ (simply $\mathcal{S}(\mathbf{z}_0, \delta)$ for random initialisations)
     \WHILE{\textit{loss $\mathit{\mathcal{L}}$ has not converged}}
         \STATE Decode: $\mathbf{x} = \mu_\theta(\mathbf{x}|\mathbf{z}_i)$;
         \STATE Use predictor to obtain $\mathcal{H}(\mathbf{y}|\mathbf{x})$;
         \STATE $\mathcal{L}=\mathcal{H}(\mathbf{y}|\mathbf{x})+d(\mathbf{x}, \mathbf{x}_0)$;
         \STATE Update $\mathbf{z}_i$ with $\nabla_\mathbf{z}\mathcal{L}$;
         \IF{$\rho(\mathbf{z}_i, \mathbf{z}_0)>\delta$}
            \STATE Project $\mathbf{z}_i$ onto the surface of the $\delta$-ball as $\mathbf{z}_i = \delta\times\frac{\mathbf{z}_i-\mathbf{z}_0}{\rho(\mathbf{z}_i,\mathbf{z}_0)}$;
         \ENDIF
     \ENDWHILE
     \STATE Decode explanation: $\mathbf{x}_{\delta-\mathrm{CLUE}}=\mu_\theta(\mathbf{x}|\mathbf{z}_i)$;
    \IF{$\mathcal{H}(\mathbf{y}|\mathbf{x}_{\delta_\mathrm{CLUE}}) < \mathcal{H}_{\mathrm{threshold}}$}
        \STATE $X_{\mathrm{CLUE}}\leftarrow X_{\mathrm{CLUE}}\cup\mathbf{x}_{\delta_\mathrm{CLUE}}$;
    \ENDIF
 \ENDFOR
 \end{algorithmic}
 \OUTPUT $X_{\mathrm{CLUE}}$, a set of $n\leq k$ CLUEs
\caption{$\delta$-CLUE}
\label{algorithm:deltaCLUE}
\end{algorithm}

\begin{figure}[ht]
\centering
\begin{subfigure}{0.23\textwidth}
    \centering
    \includegraphics[width=0.95\textwidth]{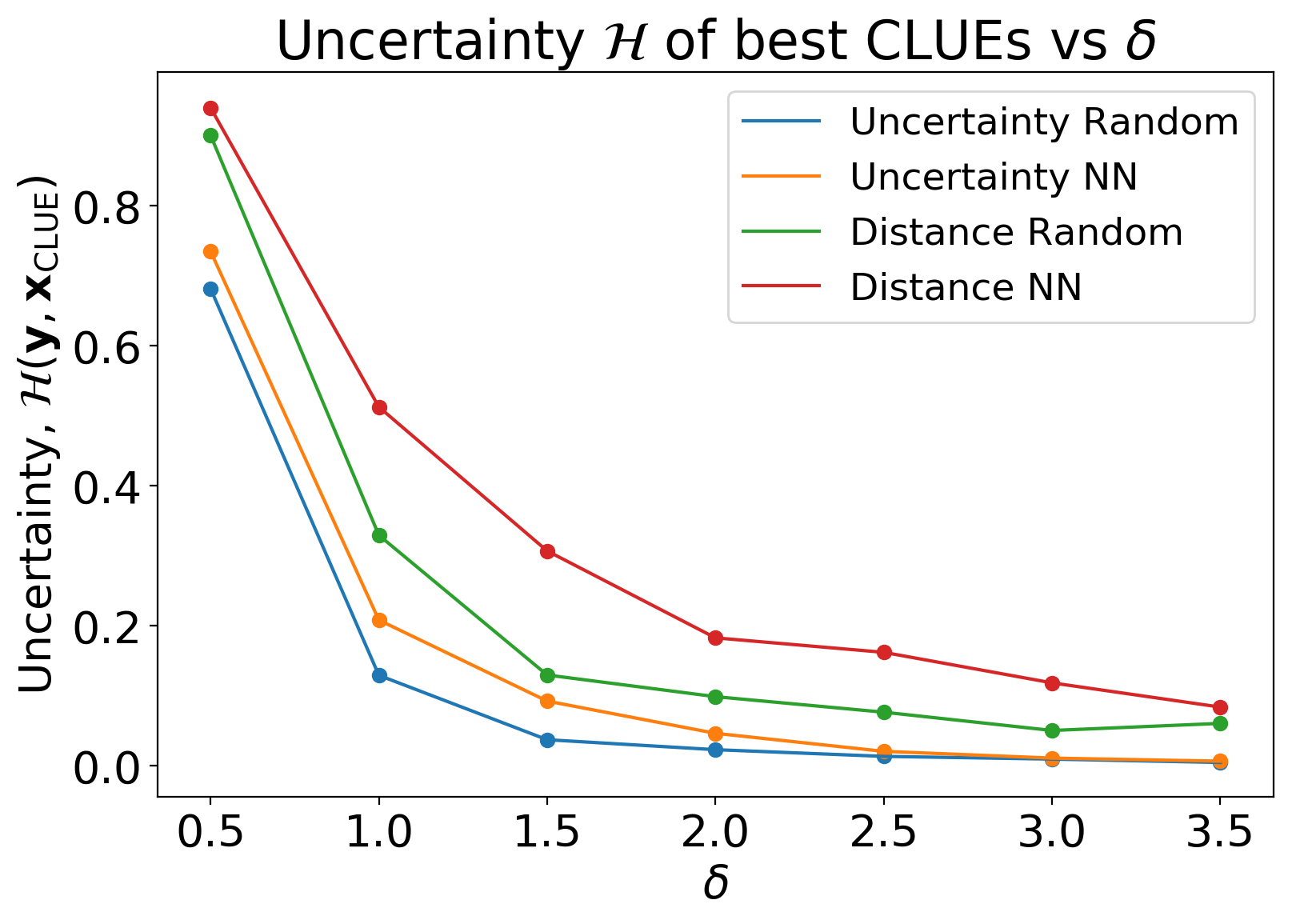}
\end{subfigure}
\centering
\begin{subfigure}{0.23\textwidth}
    \centering
    \includegraphics[width=0.95\textwidth]{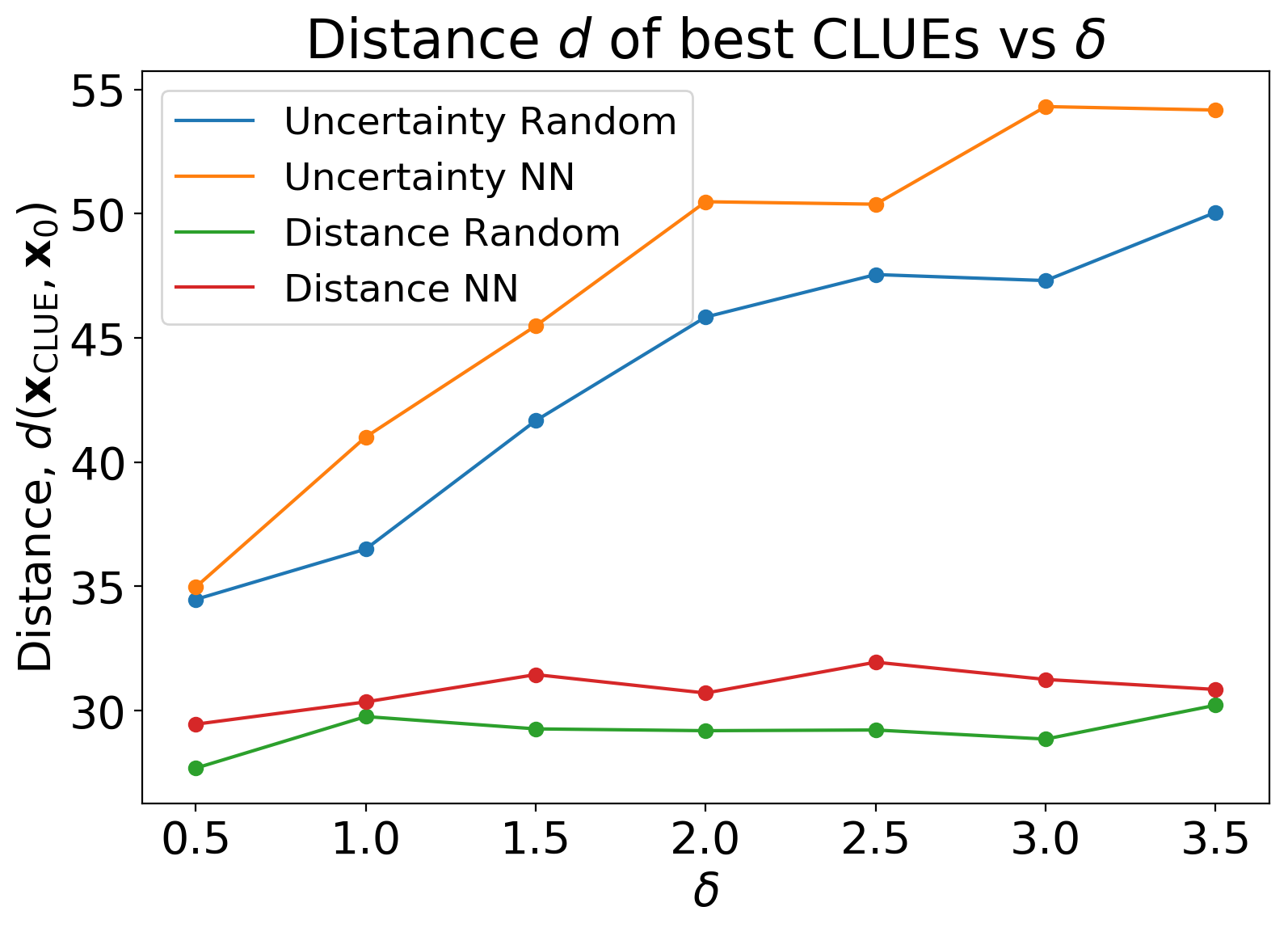}
\end{subfigure}
\caption{\small Left: Increasing the size of the $\delta$ ball yields lower uncertainty CLUEs. Right: The average distance of CLUEs from $\mathbf{x}_0$ increases with $\delta$. Note that scheme $\mathcal{S}_1$ (blue and green) outperforms scheme $\mathcal{S}_2$ (orange and red) for this dataset.}
\label{fig:uncertdistmin}
\end{figure}

In Figure~\ref{fig:uncertdistmin}, the $\mathcal{L}_{\mathcal{H}}$ %\textbf{Uncertainty Random/Uncertainty NN}
experiments (blue and orange) demonstrate how the best CLUEs found improve as the $\delta$ ball expands, at the cost of increased distance from the original input. The $\mathcal{L}_{\mathcal{H}+d}$ %\textbf{Distance Random/Distance NN}
experiments (green and red) suggest that the $\mathcal{L}_{\mathcal{H}+d}$ objective can vastly improve performance when it comes to distance (right), at the expense of higher (but acceptable) uncertainty.

\begin{figure}[ht]
    \centering
    \includegraphics[scale=0.18]{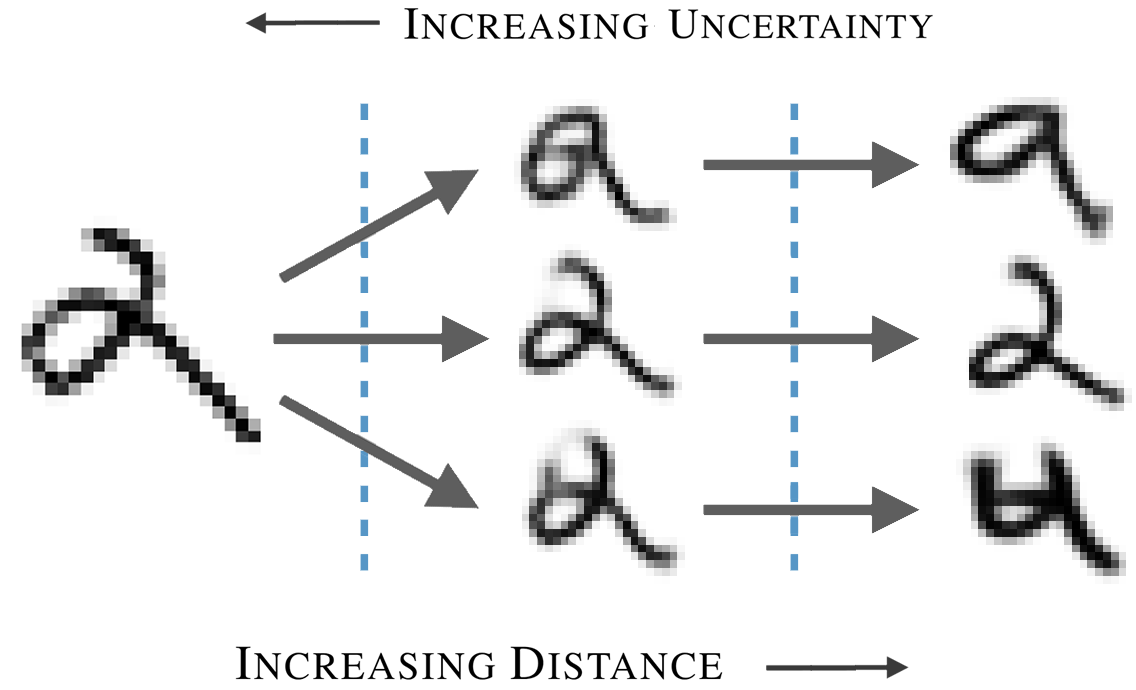}
    \caption{\small MNIST visualisation of the trade off between uncertainty $\mathcal{H}$ and distance $d$ (example of 3 diverse labels discovered by $\delta$-CLUE).}
    \label{fig:progress}
\end{figure}

\textbf{Takeaway 1:} as $\delta$ increases, using either loss $\mathcal{L}_{\mathcal{H}}$ or $\mathcal{L}_{\mathcal{H}+d}$, we reduce the uncertainty of our CLUEs at the expense of greater distance $d$. Loss $\mathcal{L}_{\mathcal{H}+d}$ experiences larger performance gains in the distance curves (green and red, Figure \ref{fig:uncertdistmin}, right).

We demonstrate that $\delta$-CLUEs are successful in converging sufficiently to all local minima within the ball, given large enough $n$ (Figure \ref{fig:entropydiversity}, left). Additionally, as the size of the $\delta$ ball increases, the random generation scheme $\mathcal{S}_1$ used in experiments \textbf{Uncertainty Random} and \textbf{Distance Random} converge to the highest numbers of diverse CLUEs (Figure \ref{fig:entropydiversity}, right, blue and green). In both loss function landscapes ($\mathcal{L}_{\mathcal{H}}$ and $\mathcal{L}_{\mathcal{H}+d}$), we obtain similarly high levels of diversity as $\delta$ increases.

\textbf{Takeaway 2:} we can achieve a diverse plethora of high quality CLUEs when it comes to both class labels and modes of change within classes, permitting a full summary of uncertainty.

Given a diverse set of proposed $\delta$-CLUEs (Figure \ref{fig:progress}), the performances of each class can be ranked by choosing an appropriate $\delta$ value and loss $\mathcal{L}$ for the mentioned trade-offs (see the last subsection of this Appendix). Here, the 2 achieves lower uncertainty for a given distance, whilst the 9 and 4 require higher distances to achieve the same uncertainty. Without a $\delta$ constraint, we can move far from the original input and obtain a CLUE from any class that is certain to the BNN.

\textbf{Takeaway 3:} we can produce a \textbf{label distribution} over the $\delta$-CLUEs to better summarise the diverse changes that could be made to reduce uncertainty.
%There is also potential for this label distribution to yield soft assignments for the data that could be used to re-train the BNN with and improve performance.
\subsection{Gradient Descent vs Sampling}

\begin{figure}[ht]
    \centering
    \includegraphics[width=0.45\textwidth]{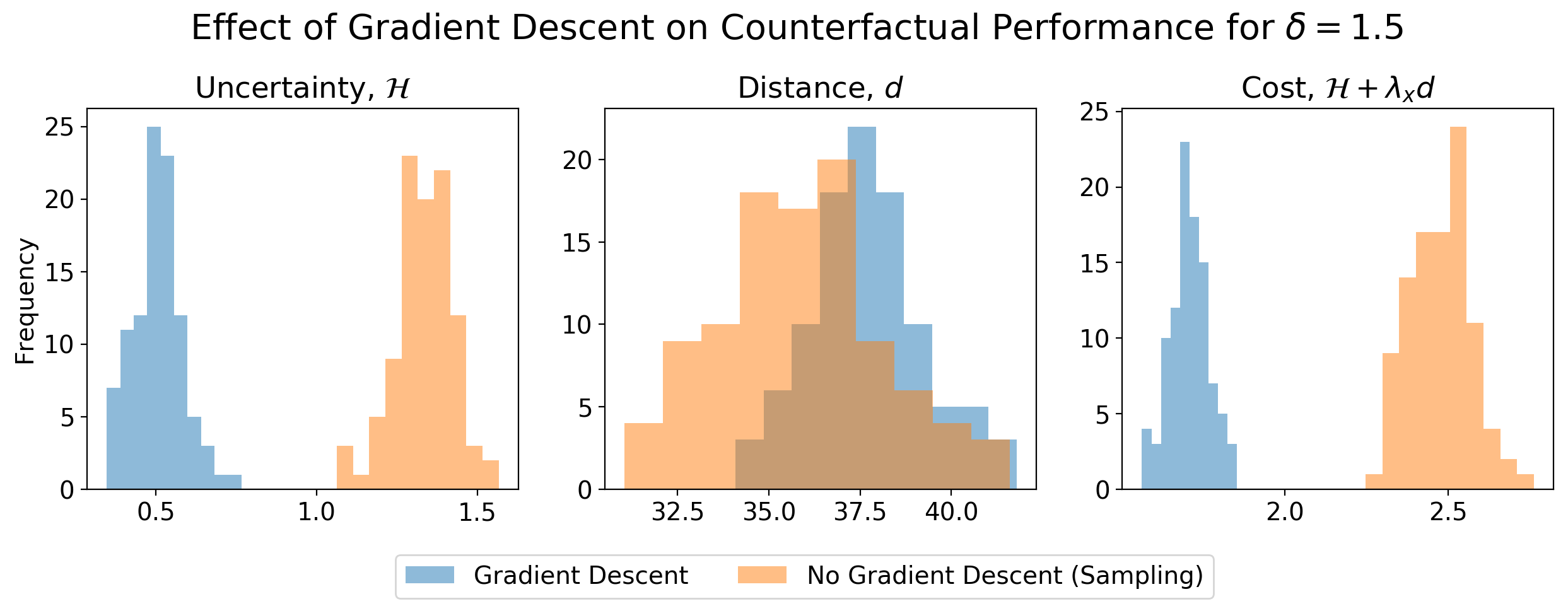}
    \caption{\small Effect of gradient descent on performance. Blue: gradient descent vs Orange: no gradient descent (sampling). For a particular $\delta$ value, we compute 100 random samples for each of the 8 most uncertain MNIST digits. We then perform constrained gradient descent over each of these values and compare the performance to the initial sample.}
    \label{fig:gradientdescent1.5}
\end{figure}

\begin{figure}[ht]
    \centering
    \includegraphics[width=0.45\textwidth]{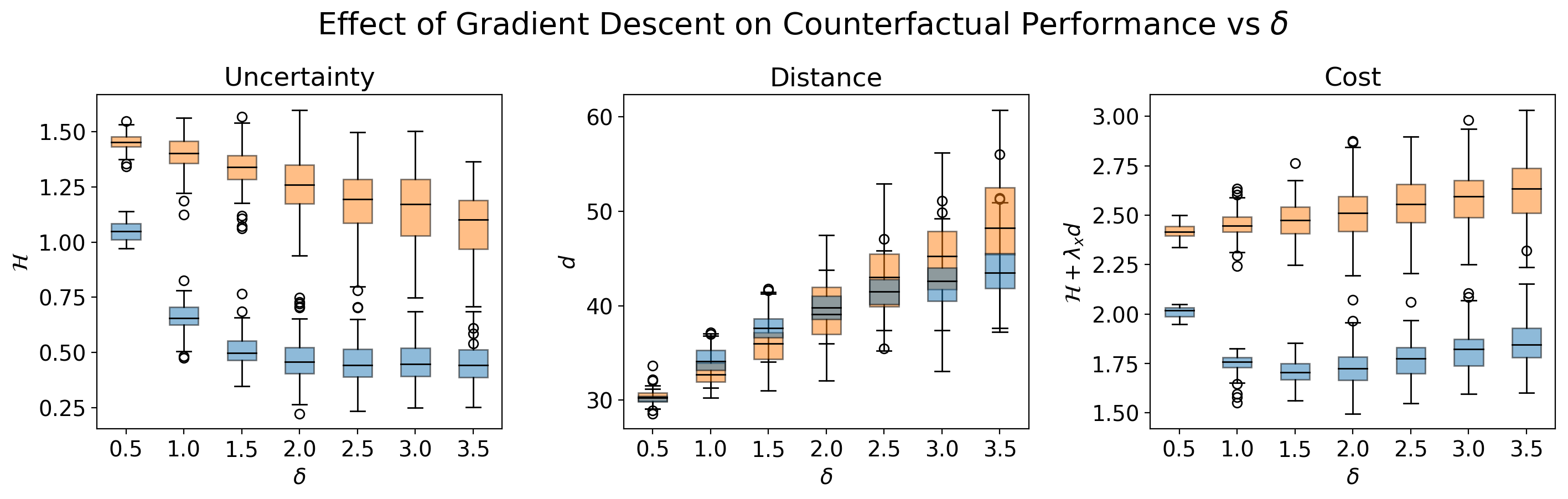}
    \caption{\small We repeat Figure~\ref{fig:gradientdescent}, over a range of $\delta$ values, showing that random sampling followed by gradient descent is far superior to only random sampling.}
    \label{fig:gradientdescent}
\end{figure}

Figures~\ref{fig:gradientdescent1.5} and~\ref{fig:gradientdescent} demonstrate the superiority of performing gradient descent over sampling in terms of the quality of counterfactuals found. Sampling does have the advantage of being computationally faster, and so more advanced search methods in future work could utilise sampling before performing gradient descent to achieve optimal performance with faster computation.


\subsection{Distance Metrics}
\label{appendix:distancemetrics}

In this section, we take $d_x(\mathbf{x}, \mathbf{x}_0) = \|\mathbf{x}-\mathbf{x}_0\|_1$ to encourage sparse explanations. The original CLUE paper found that for regression, $d_y(f(\mathbf{x}), f(\mathbf{x}_0))$ is mean squared error, and for classification, cross-entropy is used, noting that the best choice for $d(\cdot,\cdot)$ will be task-specific.

In some applications, these simple metrics may be insufficient, and recent work by \cite{zhang2018unreasonable} alludes to the shortcomings of even more complex distance metrics such as PSNR and SSIM. For MNIST digits (28x28 pixels), \textit{Mahanalobis distance} has been shown to be effective \citep{weinberger2009distance}, as well as other methods that achieve translation invariance \citep{grover2019mnist}.

\begin{figure}[ht]
\centering
\includegraphics[width=0.4\textwidth]{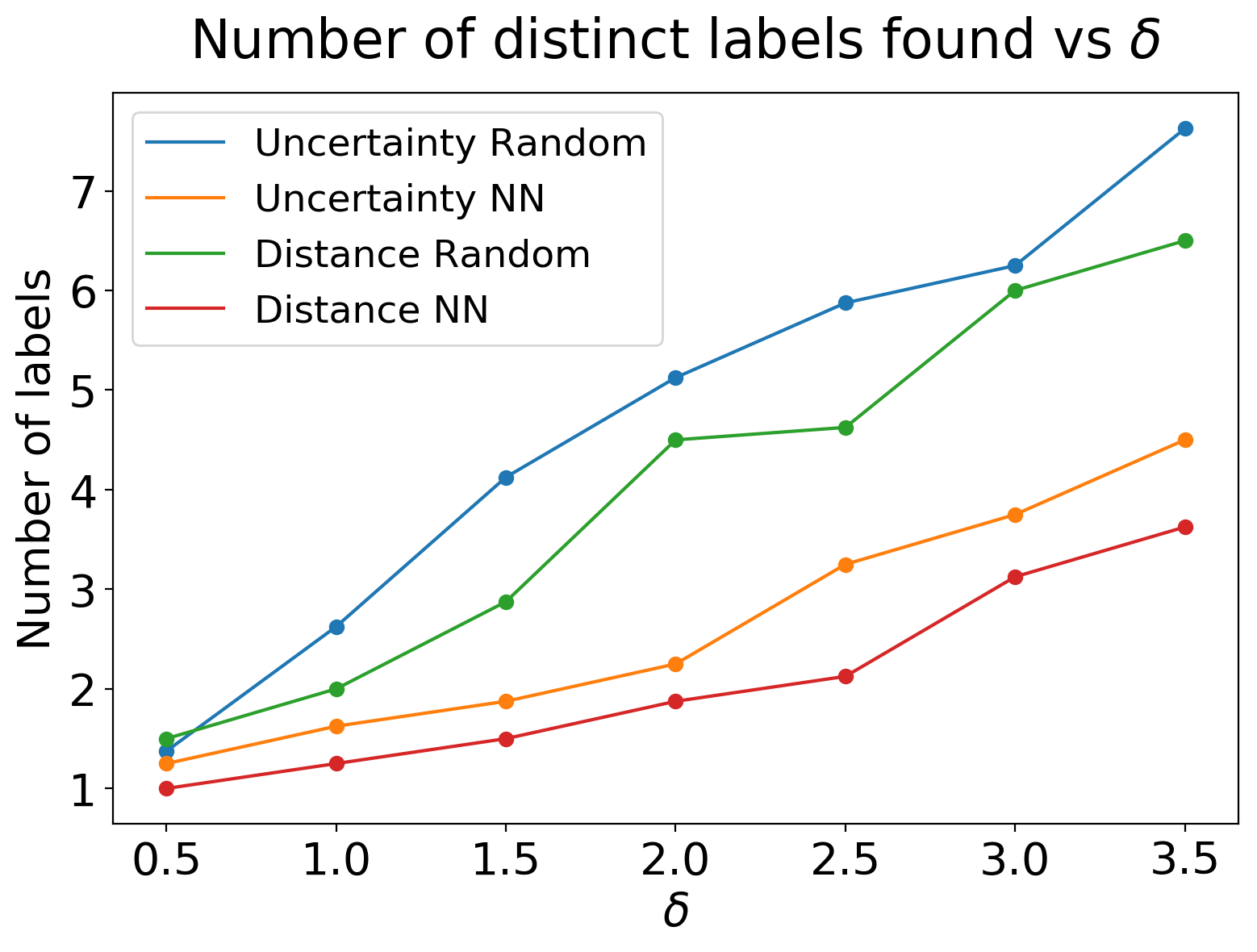}
\caption{\small Average number of distinct labels found by sets of 100 CLUEs as $\delta$ increases. For small $\delta$, typically only 1 class exists (low diversity). The random search $\mathcal{S}_1$ (blue and green) achieves the greatest diversity.}
\label{fig:entropydiversity}
\end{figure}

\begin{figure}[ht]
    \centering
    \includegraphics[width=0.45\textwidth]{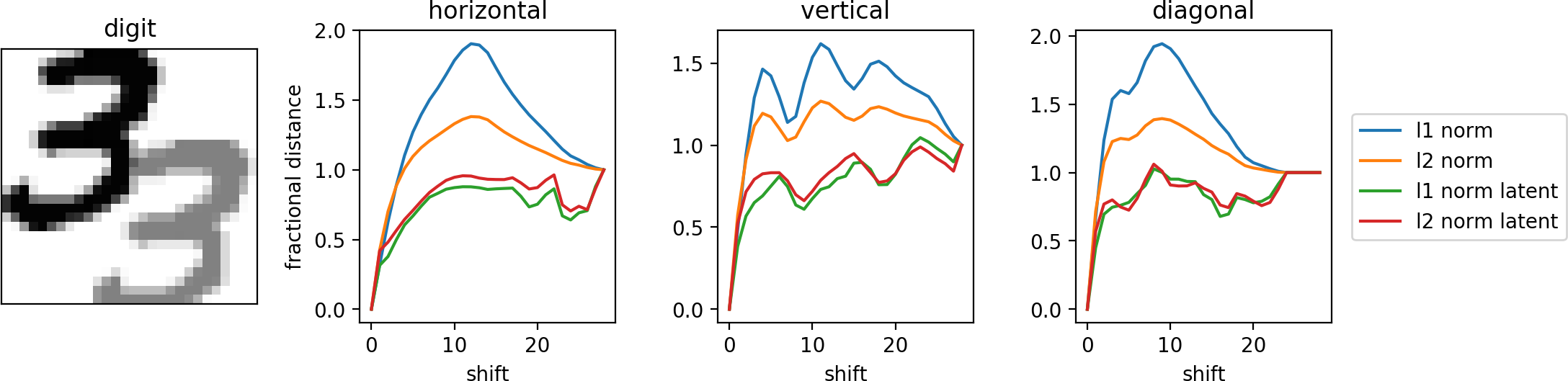}
    \caption{\small We apply horizontal, vertical and diagonal translations of an MNIST digit (in input space and latent space for $\ell_1$ and $\ell_2$ norms). As we increase the pixel shift, we compute the distance between the shifted and original digits, divided by the distance between an empty image and the original (to normalise over different metrics, resulting in convergence to 1.0). For reference, the shaded digit indicates the original digit shifted diagonally by 10 pixels.}
    \label{fig:translationrobustness}
\end{figure}

For instance, the experiment in Figure \ref{fig:translationrobustness} details how simple distance norms (either in input space or latent space) lack robustness to translations of even 5 pixels.

\subsection{Constrained vs Unconstrained Search}
\label{appendix:constrainedvsunconstrained}

\begin{figure}[ht]
    \centering
    \includegraphics[scale=0.35]{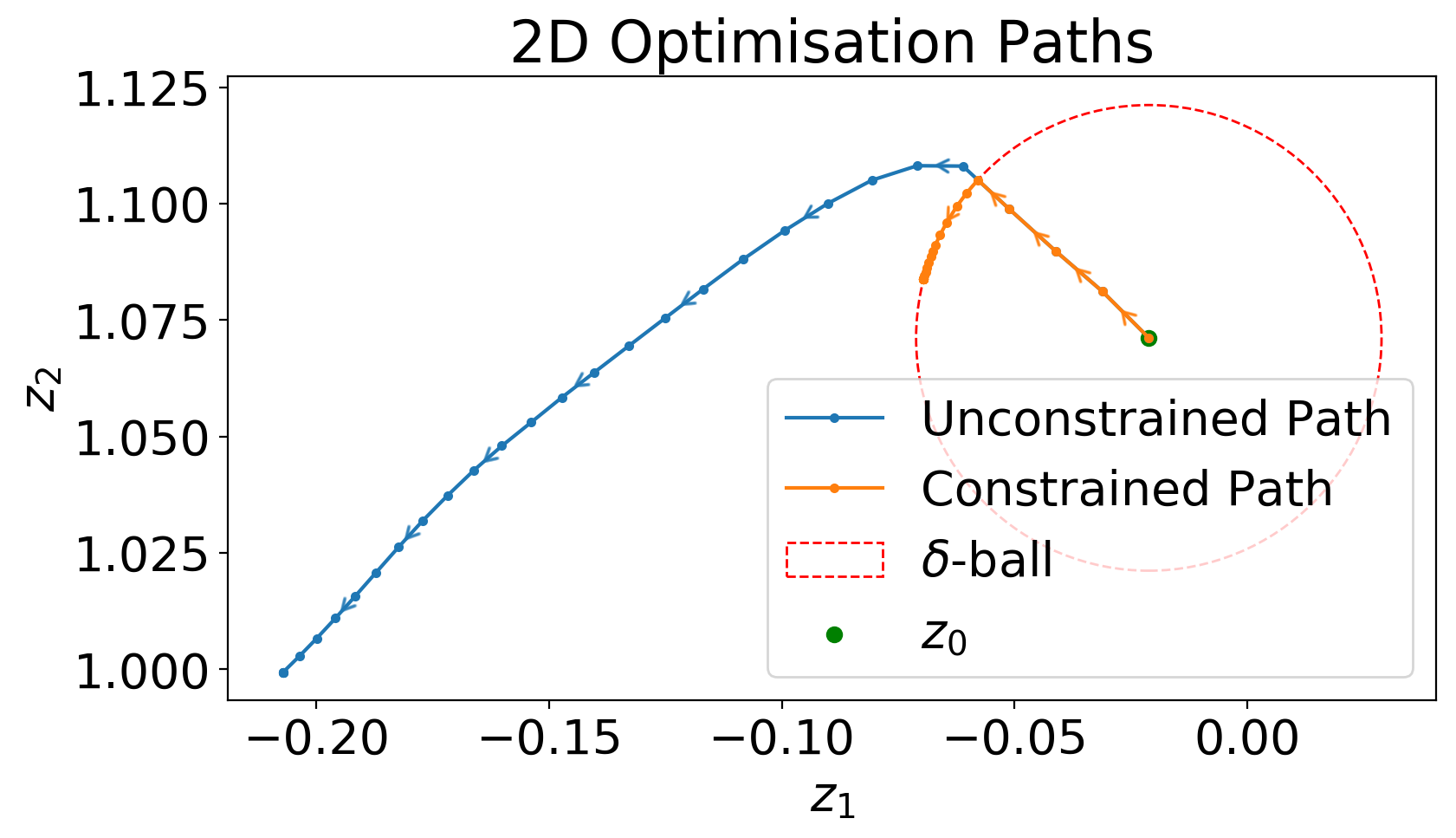}
    \caption{\small Constrained vs unconstrained gradient descents in a 2D VAE latent space $\mathcal{L}(\mathbf{z})=\mathcal{H}$. We project values outside of the $\delta$ ball onto its surface at each step of the gradient descent.}
    \label{fig:constrainedvsunconstrained}
\end{figure}

For small $\delta$, minima within the $\delta$ ball are rare, and so it is necessary to use a constrained optimisation method in our experiments (Figure~\ref{fig:constrainedvsunconstrained}), to avoid solutions being rejected.

\begin{figure}[ht]
\begin{subfigure}{0.23\textwidth}
    \centering
    \includegraphics[width=0.95\textwidth]{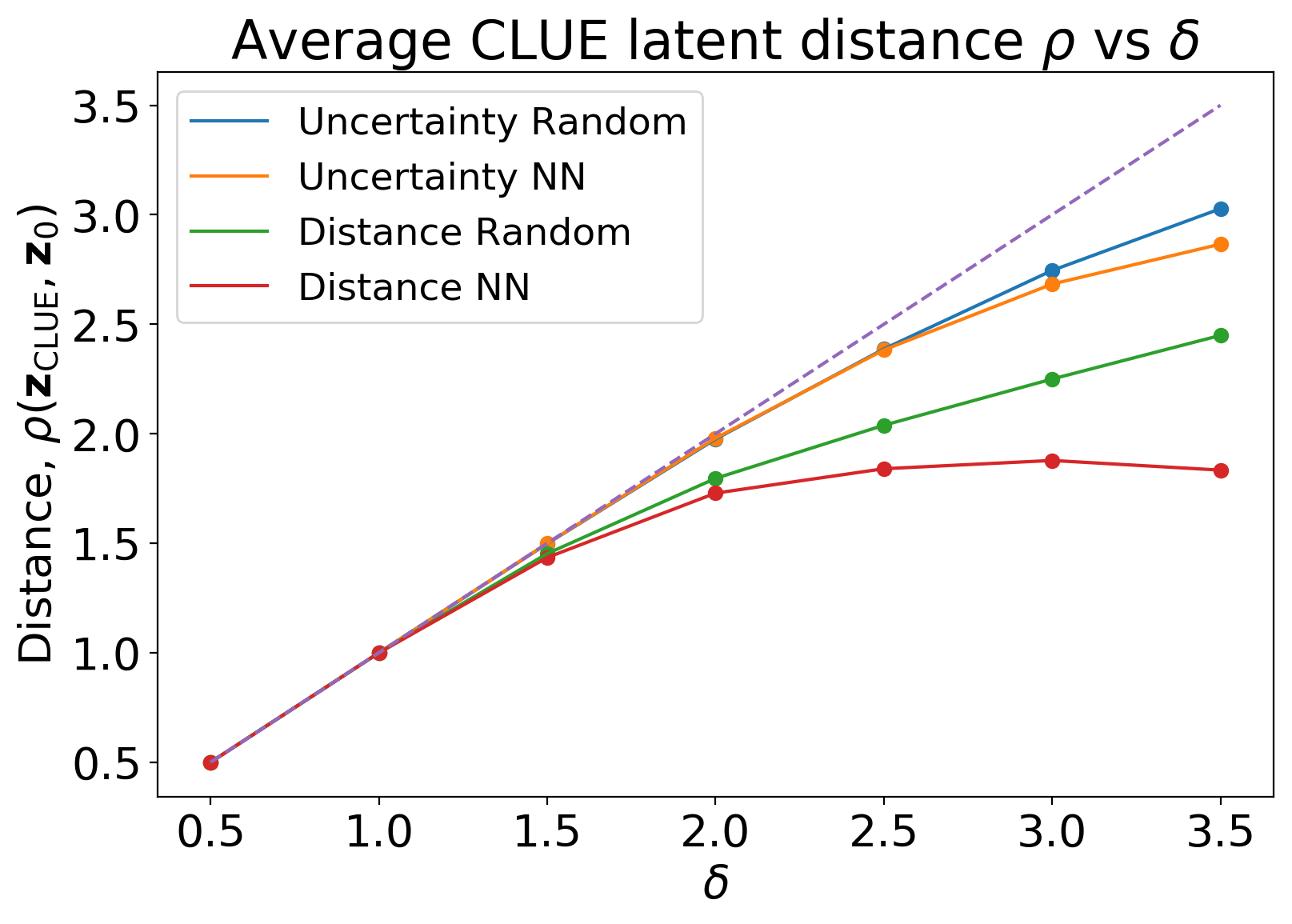}
\end{subfigure}
\begin{subfigure}{0.23\textwidth}
    \centering
    \includegraphics[width=0.95\textwidth]{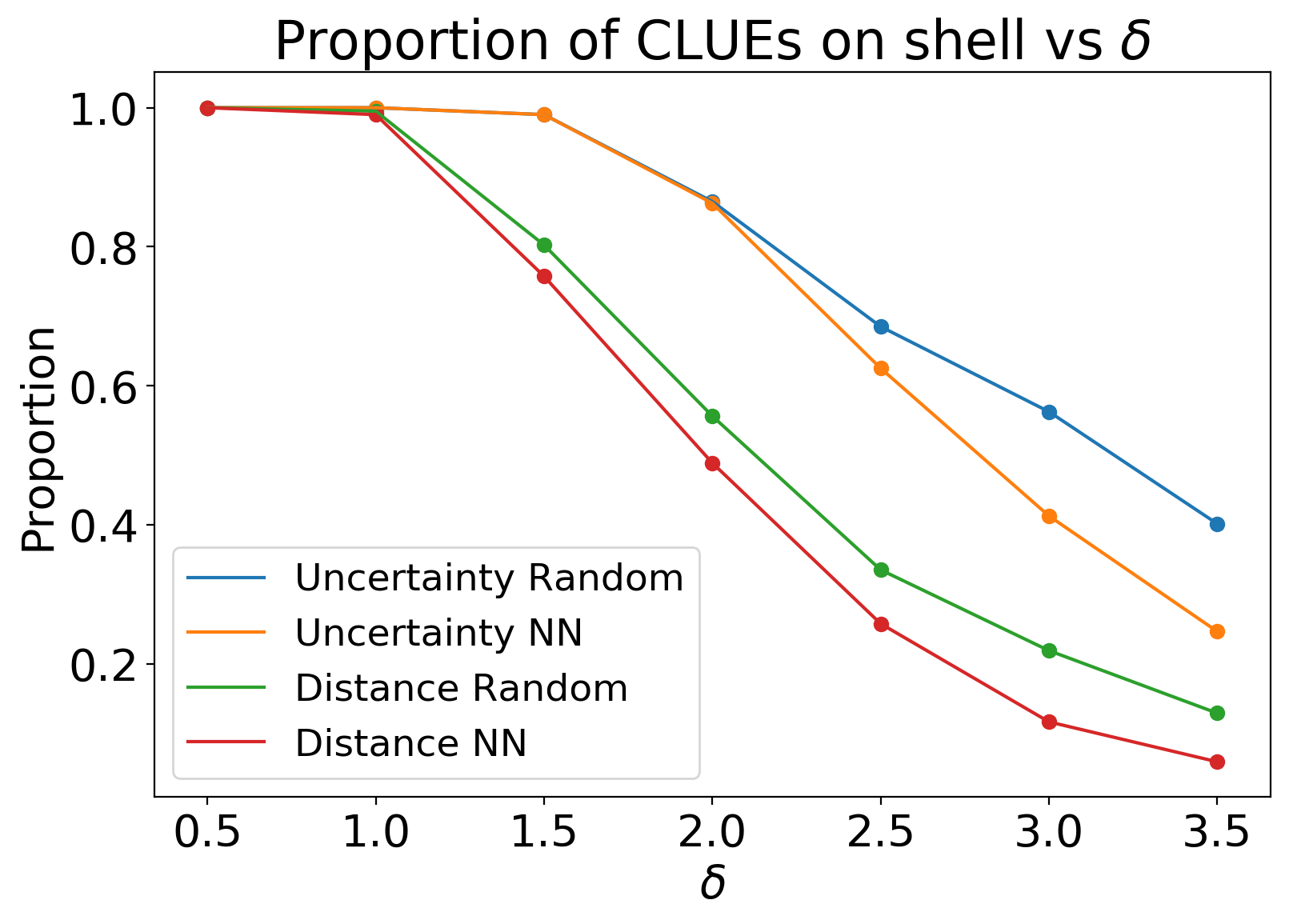}
\end{subfigure}
\caption{\small Justification for use of a constrained method. More solutions lie on the ball than inside it for a given $\delta$. Left: How the average final distance in latent space varies with $\delta$. Right: proportion of points that lie on the shell as $\delta$ increases. At small $\delta$, almost all minima lie on the shell, whereas at larger $\delta$ more lie inside.}
\label{fig:zdists}
\end{figure}

Thus, we observe in Figure \ref{fig:zdists}, right, that for small $\delta$, virtually all $\delta$-CLUEs lie on the surface of the ball. The left figure indicates that average latent space distances $\rho(\mathbf{z}_\mathrm{CLUE}, \mathbf{z}_0)$ lie close to the line $\delta=\delta$ (purple, dashed), with the distance weighted loss $\mathcal{L}_{\mathcal{H}+d}=\mathcal{H}+d$ producing more nearby $\delta$-CLUEs, as expected. In either case, the effect of the constraint weakens for larger $\delta$, as more minima exist within the ball instead of on it. Depending on user preference, the optimal $\delta$ value represents the trade off between uncertainty reduction and distance from the original input.

\textbf{As stated in the main text}, there may exist methods to determine $\delta$ pre-experimentation; the distribution of training data in latent space could potentially uncover relationships between uncertainty and distance, both for individual inputs and on average. For instance, we might search in latent space for the distance to nearest neighbours within each class to determine $\delta$. In many cases, it could be useful to provide a summary of counterfactuals at various distances and uncertainties, making a range of $\delta$ values more appropriate.

\subsection{Initialisation Schemes $\mathcal{S}_i$}
\label{appendix:schemes}

This appendix details the initialisation schemes $\mathcal{S}_i$ that are used to generate start points for the algorithm. While some schemes may appear preferential in 2 dimensions, the manner at which these scale up to higher dimensions means that we could require an infeasible number of initialisations to cover the appropriate landscape, and so deterministic schemes such as a path towards nearest neighbours within each class ($\mathcal{S}_2$), or a gradient descent into predictions within each class ($\mathcal{S}_5$) might be desirable. We should also note that although the radius, $r$, of a specific scheme could vary, it is assumed throughout to match the $\delta$ value. The following mathematical analysis applies to an $\ell_2$-norm $\rho(\mathbf{z}, \mathbf{z}_0)=\|\mathbf{z}-\mathbf{z}_0\|_2$:

\[\mathcal{S}_1:\rho(\mathbf{z}, \mathbf{z}_0)\sim \mathcal{U}(0, r)\implies\E[\mathbf{\rho(\mathbf{z}, \mathbf{z}_0)}]=\frac{r}{2}\]
\[\text{(pick a random radial direction)}\]
\[\mathcal{S}_3:\rho(\mathbf{z}, \mathbf{z}_0)\sim \mathcal{N}\left(0, \frac{r^2}{4}\right)\ \text{s.t. } 0\leq\rho(\mathbf{z}, \mathbf{z}_0)\leq r\]
\[\text{(pick a random radial direction)}\]
\[\mathcal{S}_4:[\mathbf{z}-\mathbf{z}_0]_i\sim \mathcal{U}\left(-r,r\right)\ \text{s.t. } \rho(\mathbf{z}, \mathbf{z}_0)\leq r\]
\[\text{(perturb each dimension)}\]%\implies\E[\mathbf{\rho(\mathbf{z}, \mathbf{z}_0)}]=\E\left[\left(\sum_{i=1}^N[\mathbf{z}-\mathbf{z}_0]_i^2\right)^\frac{1}{2}\right]\]%

\begin{figure}
\centering
\includegraphics[width=0.45\textwidth]{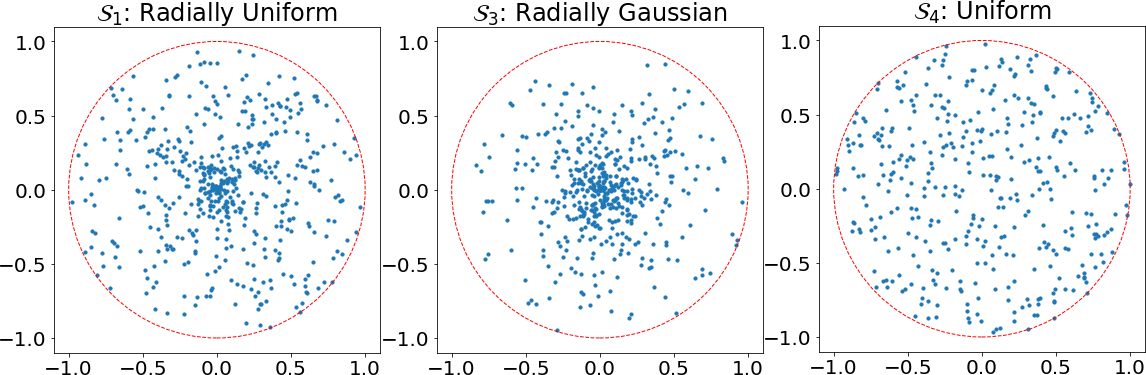}
\caption{\small Random generation schemes $\mathcal{S}_1$, $\mathcal{S}_3$ and $\mathcal{S}_4$ depicted in 2D space. In Schemes $\mathcal{S}_3$/$\mathcal{S}_4$ we reject samples outside of the search radius ($\rho(\mathbf{z}, \mathbf{z}_0)>r$). Future schemes may generate within a sub-ball that is smaller than the ball with which we constrain, though this may only be effective in specific latent landscapes.}
\label{fig:S}
\end{figure}

\begin{figure}
    \centering
    \includegraphics[width=0.45\textwidth]{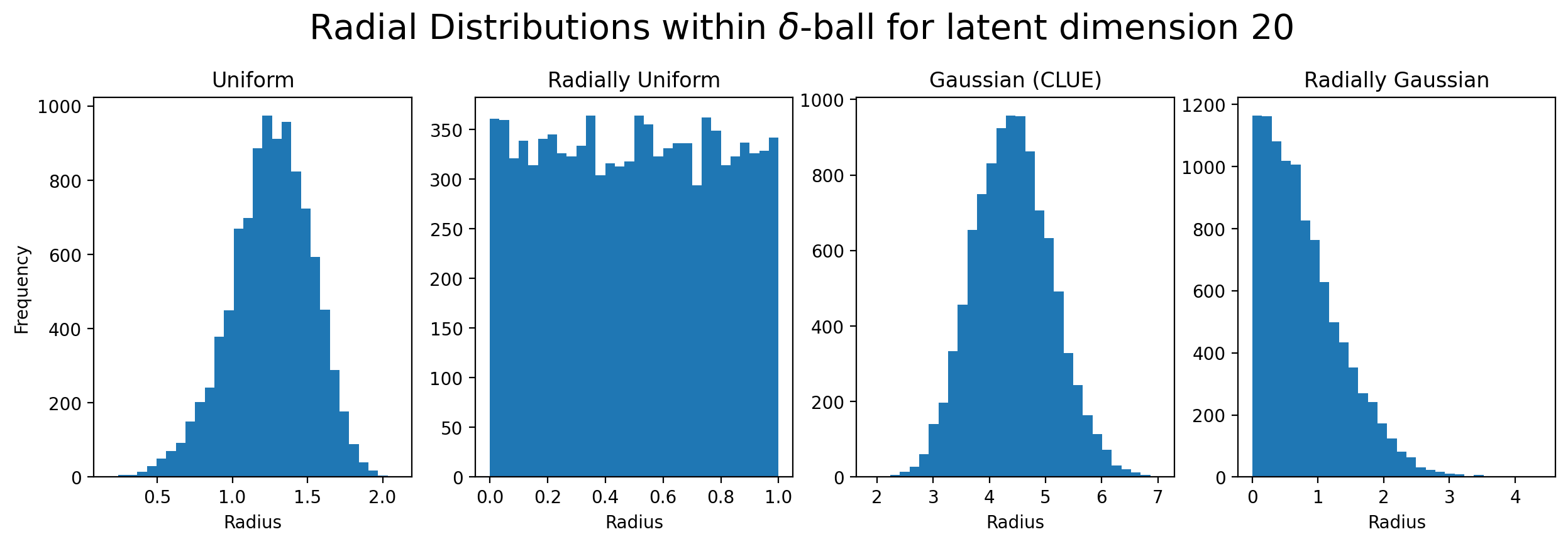}
    \caption{\small Comparison of initialisation methods. Left: perturbing each dimension uniformally ($\mathcal{S}_4$). Left centre: sampling the radius uniformally $\mathcal{U}\sim[0, 1]$ and picking a random direction uniformally ($\mathcal{S}_1$). Right centre: perturbing each dimension with Gaussian noise $\mathcal{N}(0,1)$ \citep{antoran2021getting}. Right: sampling the radius with a Gaussian $\mathcal{N}(0,1)$ and picking a random direction uniformally ($\mathcal{S}_3$). Overall, our radially uniform method allows for easiest control over the region of latent space explored (note the scales of the other methods, which are functions of either the probabilistic sampling or the latent dimension, both of which introduce added complexities when performing searches.}
    \label{fig:samplinginit}
\end{figure}

\noindent We propose two potential deterministic schemes, that may outperform a random scheme when a) the latent dimension is large, b) $\delta$ becomes very large, c) we impose a larger distance weight in the objective function or d) we change datasets. Here $\mathbf{z}_i$ represents the starting point for explanation $i$, $k$ is the total number of explanations (both used in Algorithm 1), $Y$ represents the total number of class labels $y$, and $j\in\mathbb{Z}^+$. This produces a total of $Y\times j_{\text{max}}=Y\times\left\lfloor{\frac{k}{Y}}\right\rfloor=k$ explanations if $Y|n$.

\[\mathcal{S}_2:\mathbf{z}_i = \mathbf{z}_0+\delta \times\frac{j}{m}\times \frac{\mathbf{z}_y-\mathbf{z}_0}{\rho(\mathbf{z}_y,\mathbf{z}_0)}\ \forall y\]
\[\mathcal{S}_5: \mathbf{z}_i = \mathbf{z}_0+\mathbf{s}_{yj}\ \forall y\]
\[\text{where }1\leq j\leq m\ \text{and }m=\left\lfloor{\frac{k}{Y}}\right\rfloor\]

\noindent where, for the $\mathcal{S}_5$ scheme, $\mathbf{s}_{yj}$ is defined along a path from $\mathbf{z}_0$ to a radius $\delta$, where at all points the direction of $\mathbf{s}$ is $\nabla_\mathbf{z}p(\text{class}(\mathbf{z})=y)$, and $\frac{j}{m}$ is defined as the fraction travelled along that path.

\begin{figure}[ht]
    \centering
    \includegraphics[width=0.45\textwidth]{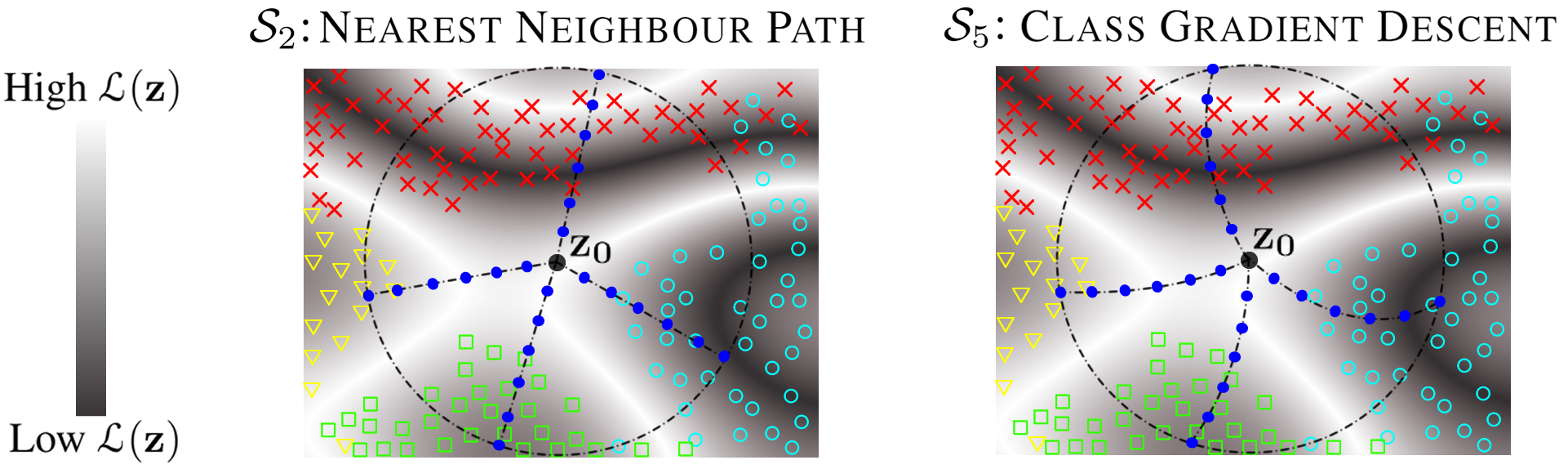}
    \caption{\small Left: Scheme $\mathcal{S}_2$, nearest neighbour path, searches for the nearest low uncertainty points in training data for each class, before initialising starting points fractionally on the path towards said neighbour. Right: Scheme $\mathcal{S}_5$ performs a gradient descent in the prediction space of the BNN, towards maximising the probability of each class. It too initialises starting points along said path.}
    \label{fig:SDeterministic}
\end{figure}

\noindent A series of modifications to these schemes may improve their performance:

\begin{itemize}
    \item Generating within small regions around each of the points along the path (in $\mathcal{S}_2$ and $\mathcal{S}_5$).
    \item Performing a series of further subsearches in latent space around each of the best $\delta$-CLUEs under a particular scheme.
    \item Combining $\delta$-CLUEs from multiple schemes to achieve greater diversity.
    \item Appendix D details maximising the \textit{diversity} of initialisations, before performing gradient descent on the loss function, which could find utility in cases of low $k$.
\end{itemize}

\subsection{Further MNIST $\delta$-CLUE Analysis}
\label{appendix:analysis}

\begin{figure}[ht]
\begin{subfigure}{0.23\textwidth}
    \centering
    \includegraphics[width=0.95\textwidth]{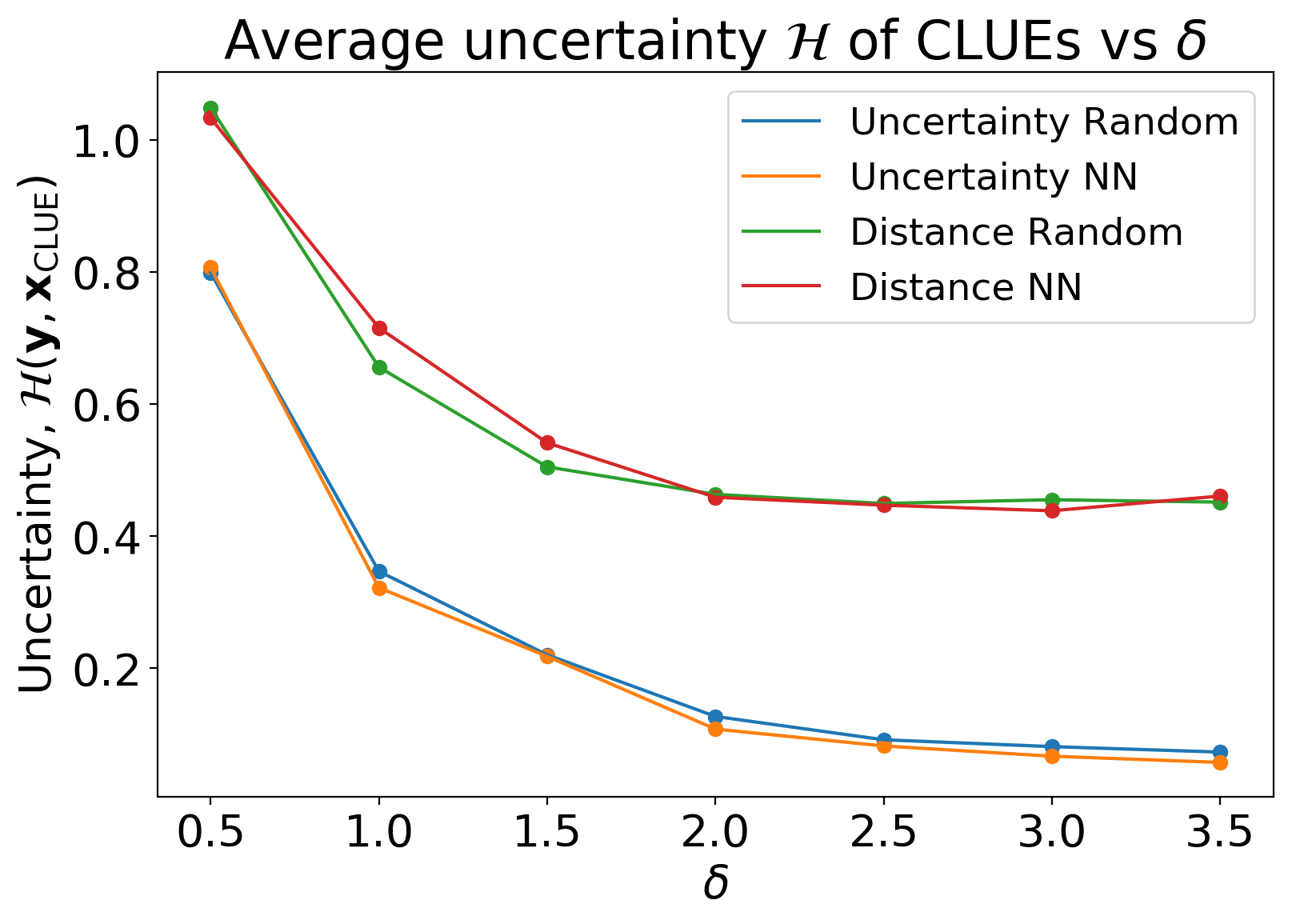}
\end{subfigure}
\begin{subfigure}{0.23\textwidth}
    \centering
    \includegraphics[width=0.95\textwidth]{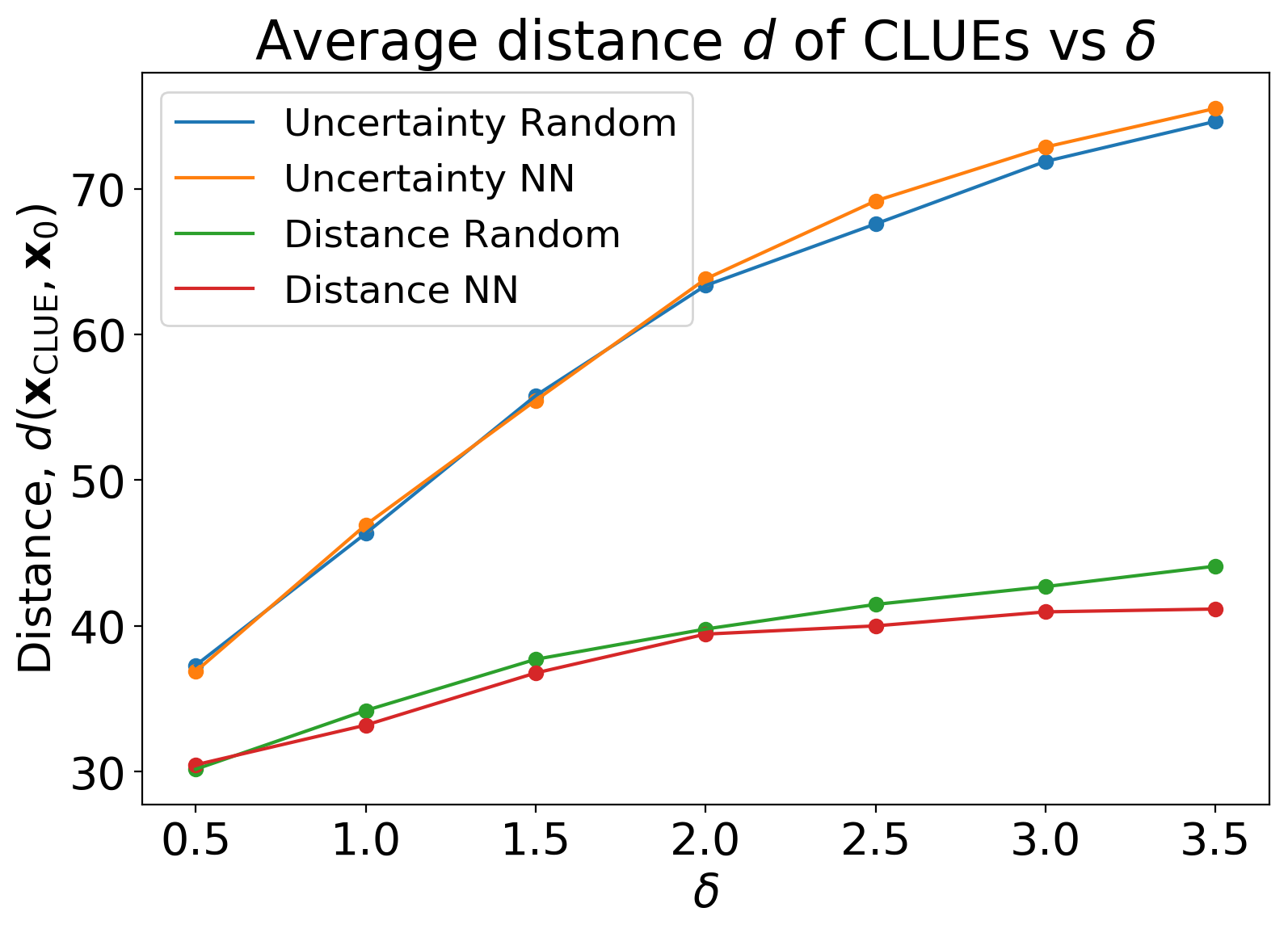}
\end{subfigure}
\caption{\small In Figure \ref{fig:uncertdistmin} of the main text, we plot the best (minimum) uncertainties/distances of the $\delta$-CLUEs. Here, we reproduce the plot for average uncertainties/distances and observe that it follows similar trends, shifted vertically, with higher disparity between the $\mathcal{L}_{\mathcal{H}}$ and $\mathcal{L}_{\mathcal{H}+d}$ loss functions.}
\label{fig:uncertdistavg}
\end{figure}

\begin{figure}[ht]
\begin{subfigure}{0.23\textwidth}
    \centering
    \includegraphics[width=0.95\textwidth]{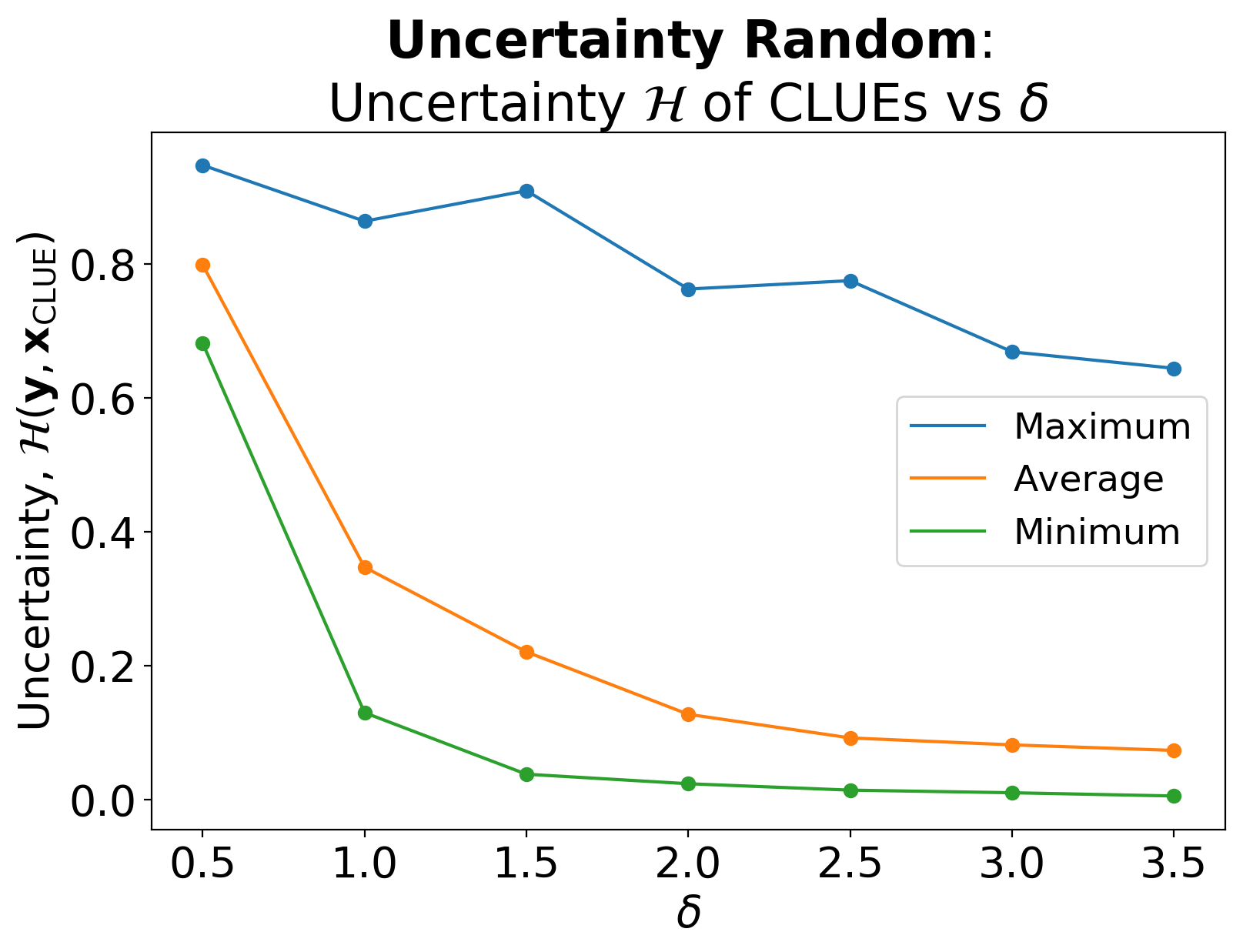}
\end{subfigure}
\begin{subfigure}{0.23\textwidth}
    \centering
    \includegraphics[width=0.95\textwidth]{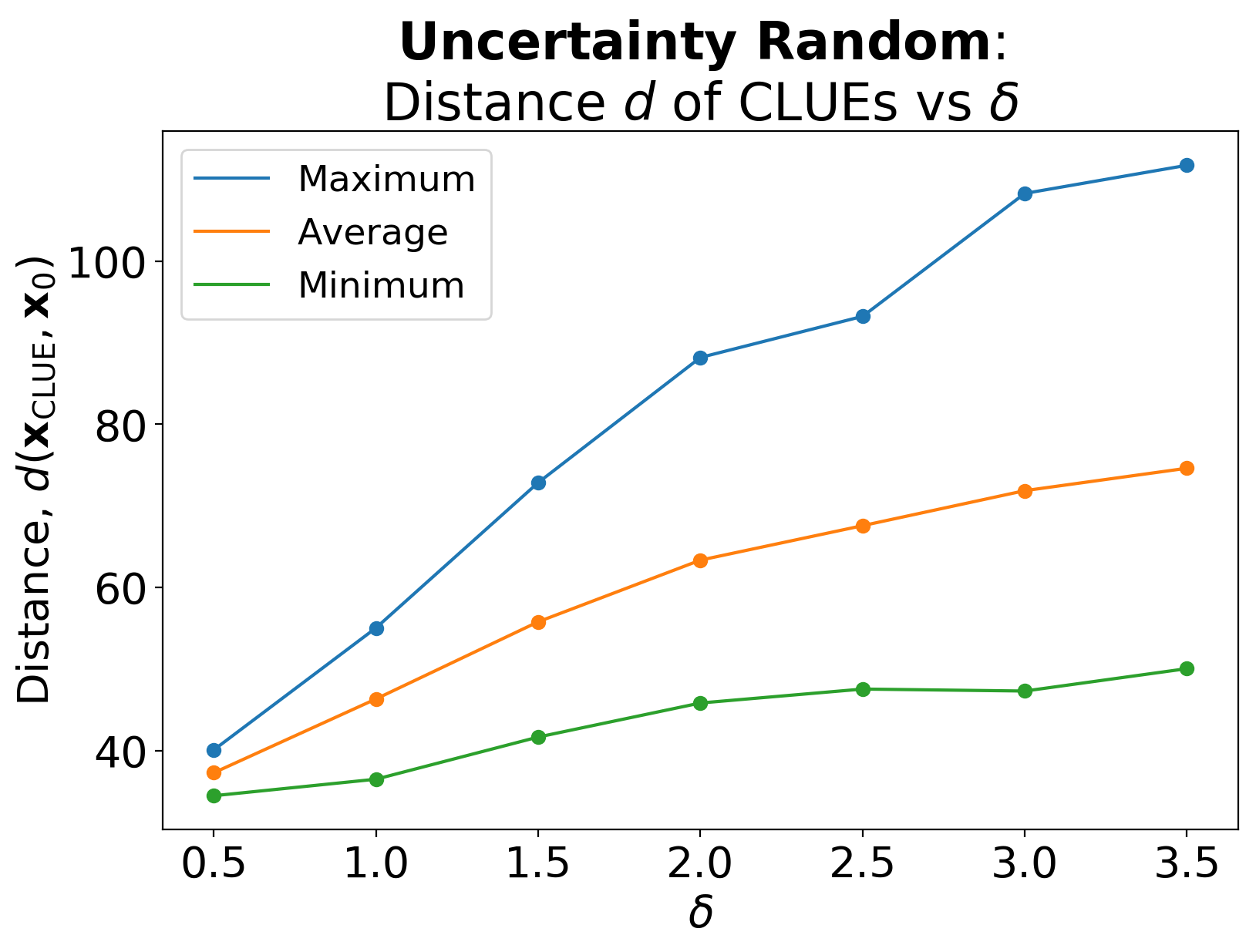}
\end{subfigure}
\caption{\small We reproduce Figure \ref{fig:uncertdistmin} for the \textbf{Uncertainty Random} experiment ($\mathcal{L}_{\mathcal{H}}=\mathcal{H}$ and $\mathcal{S}_1$), plotting the minimum, average and maximum values found in the set of 100 $\delta$-CLUEs averaged over 8 uncertain inputs.}
\label{fig:uncertdistminmax}
\end{figure}

\begin{figure}[ht]
    \centering
    \includegraphics[width=0.45\textwidth]{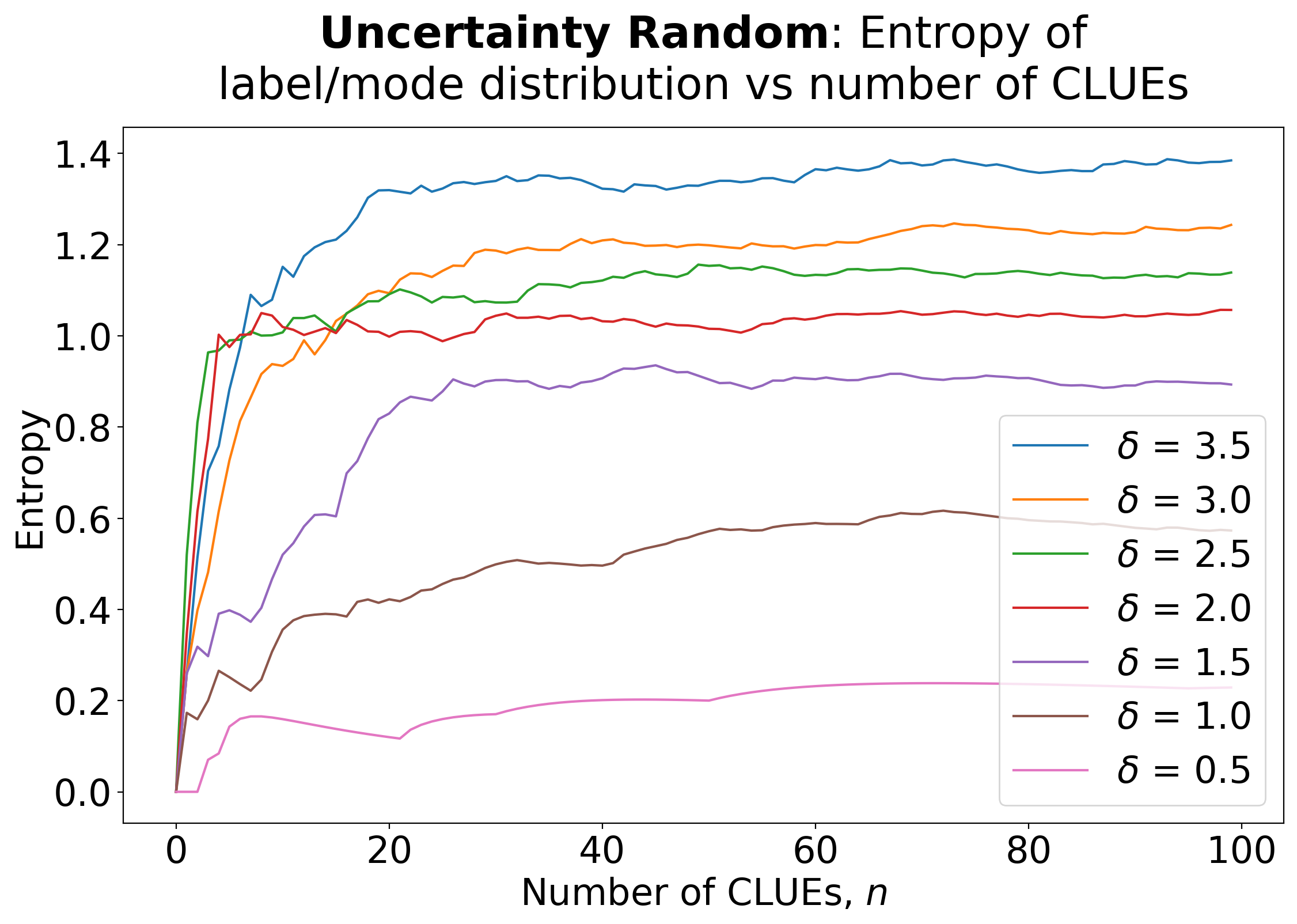}
    \caption{\small A more refined plot of Figure \ref{fig:entropydiversity}, left, to answer the question: ``How many times must we run $\delta$-CLUE in order to saturate the entropy of the label distribution of the $\delta$-CLUEs found?''.}
    \label{fig:labelentropy}
\end{figure}

For an uncertain input $\mathbf{x}_0$, we generate 100 $\delta$-CLUEs and compute the minimum, average and maximum uncertainties/distances from this set, before averaging this over 8 different uncertain inputs. Repeating this over several $\delta$ values produces Figures \ref{fig:uncertdistavg} through \ref{fig:labelentropy}.

Special consideration should be taken in selecting the best method to assess a set of 100 $\delta$-CLUEs: the minimum/average uncertainty/distance $\delta$-CLUEs could be selected, or some form of submodular selection algorithm could be deployed on the set. Figure \ref{fig:uncertdistminmax} shows the variance in performance of $\delta$-CLUEs; the worst $\delta$-CLUEs converge to high uncertainties and high distances that are too undesirable (the selection of $\delta$-CLUEs is then a non-trivial problem to solve, and in our analysis we simply select the best cost $\delta$-CLUE for each CLUE, where cost is a combination of uncertainty and distance).

\begin{figure}[ht]
    \centering
    \includegraphics[scale=0.25]{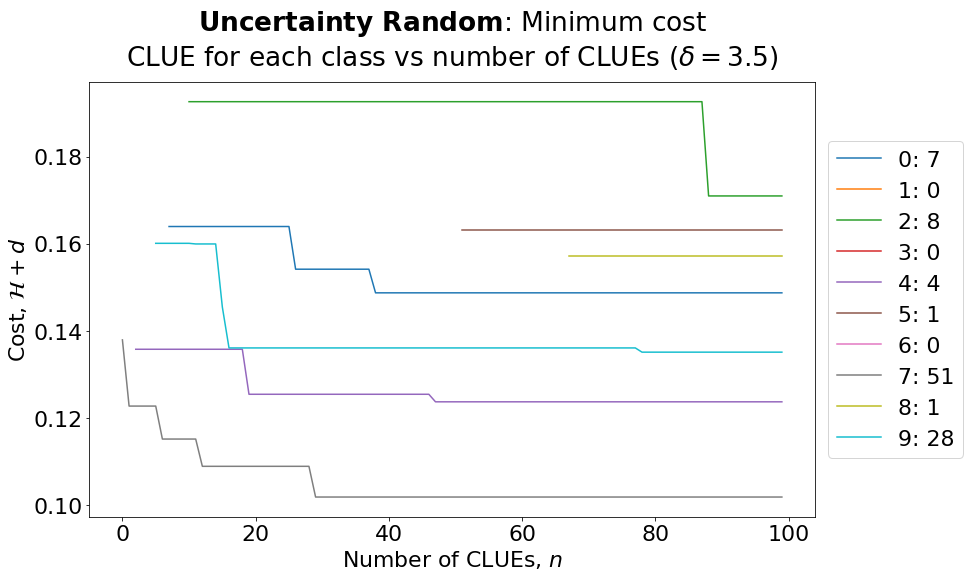}
    \caption{\small For a single uncertain input $\mathbf{x}_0$, we generate $n$ $\delta$-CLUEs and observe how the minimum cost (a combination of uncertainty and distance) of $\delta$-CLUEs for each class converges. Legend shows class labels 0 to 9, and the final number of each discovered by $\delta$-CLUE (summing to 100).}
    \label{fig:costprogression}
\end{figure}

\noindent In Figure \ref{fig:costprogression}, the late convergence of class 2 (green) and the lack of 1s, 3s and 6s suggests that $n>100$ is required, although under computational constraints $n=100$ yields good quality CLUEs for the prominent classes (7 and 9).

\begin{figure}[ht]
    \centering
    \includegraphics[scale=0.18]{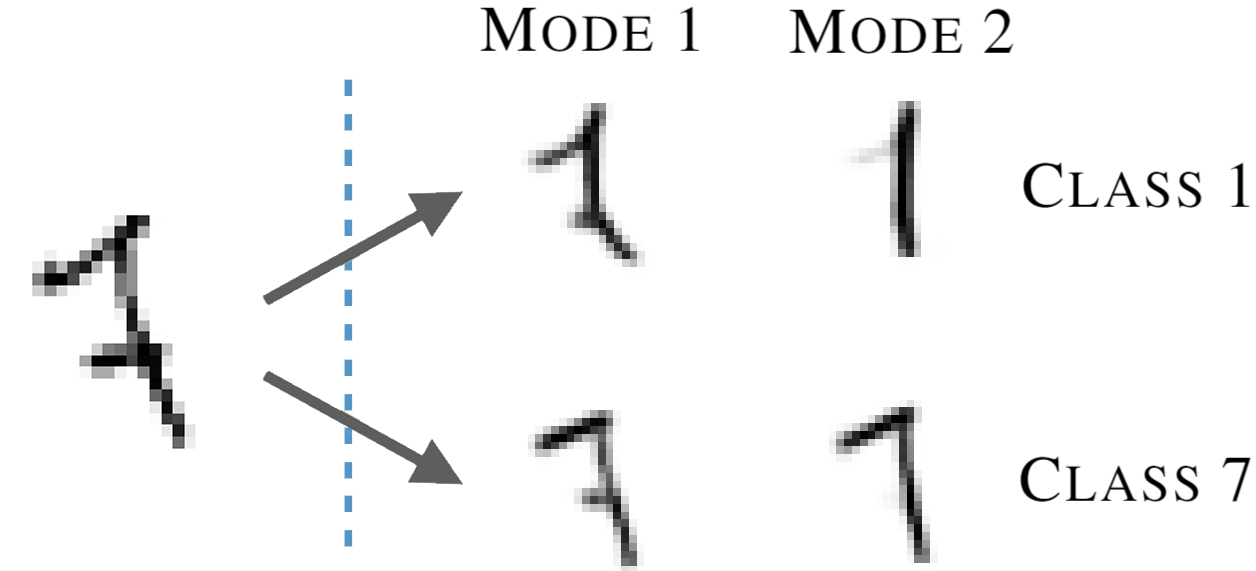}
    \caption{\small MNIST: 10 class labels exist (0 to 9), whereas an undefined number of modes within each class also exist. These modes are counted manually in this paper.}
    \label{fig:modes}
\end{figure}

Figure \ref{fig:modes} demonstrates how convergence of the $\delta$-CLUE set is a function, not only of the class labels found, but also of the different mode changes that result within each class (alternative forms of each label). In the main text (Figure \ref{fig:entropydiversity}), we count manually the mode changes within each class; in future, clustering algorithms such as Gaussian Mixture Models could be deployed to automatically assess these. The concept of modes is important when a low number of classes exists, such as in binary classification tasks, where we may require multiple ways of answering the question: ``what possible mode change could an end user make to modify their classification from a no to a yes?''.

\subsection{Computing a Label Distribution from $\delta$-CLUEs}
\label{appendix:labeldistribution}

This final appendix addresses the task of computing a label distribution from a set of $\delta$-CLUEs, as suggested by takeaway 3 of the main text. We use $\delta=3.5$ and analyse one uncertain input $\mathbf{x}_0$ under the experiment \textbf{Distance Random} where $\mathcal{L}_{\mathcal{H}+d}=\mathcal{H}+d$ and $\mathcal{S}_1$ are used.

\begin{figure}[ht]
\centering
\includegraphics[width=0.45\textwidth]{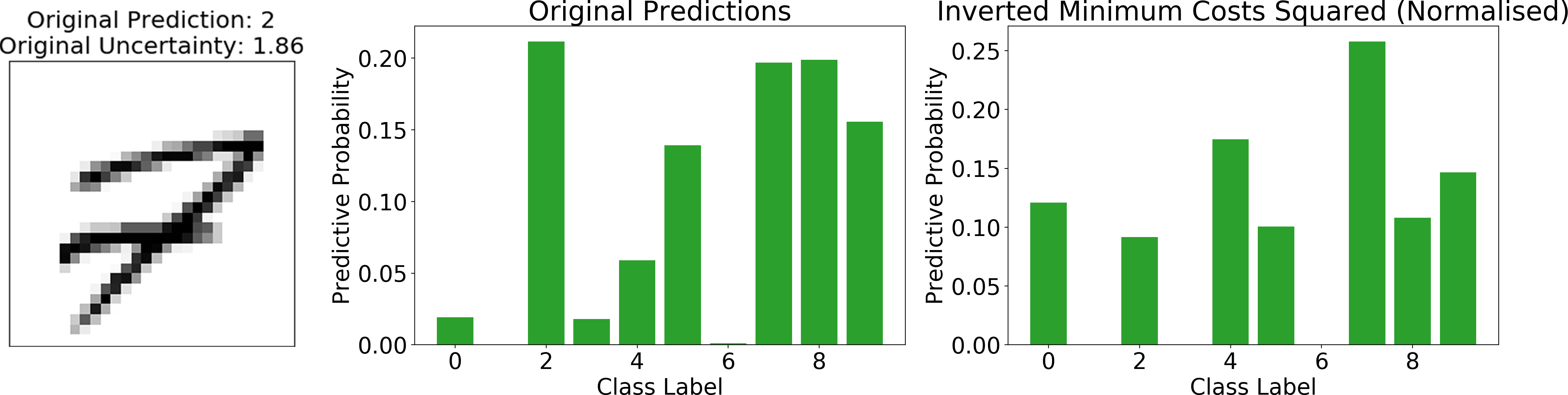}
\caption{\small Left: An original uncertain input that is incorrectly classified. Centre: The original predictions from the BNN. Right: The new \textbf{label distribution} based off of the $\delta$-CLUEs found.}
\label{fig:labeldistribution}
\end{figure}

\begin{figure}[ht]
\centering
\includegraphics[width=0.45\textwidth]{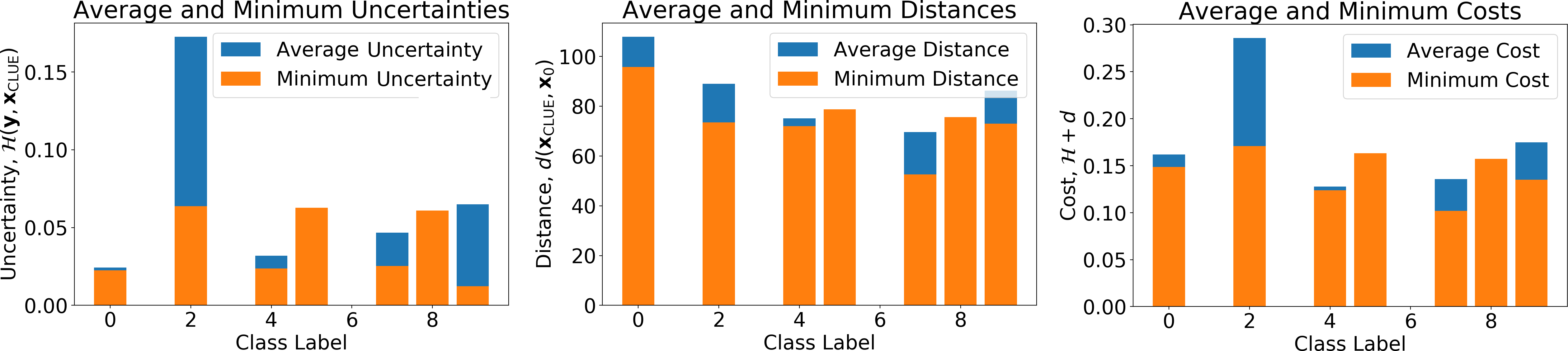}
\caption{\small Left: Average and minimum uncertainties $\mathcal{H}$ for each class in the $\delta$-CLUE set. Centre: Average and minimum distances $d$. Right: Average and minimum costs, where the weight $\lambda_x$ is multiplied by the distance function and added to the uncertainty.}
\label{fig:label}
\end{figure}

For (Figure \ref{fig:labeldistribution}, right), we take the minimum costs from (Figure \ref{fig:label}, right) and take the inverse square.

\subsubsection{Effect of VAE latent dimension, $\rvm$}

This section reproduces the \textbf{Uncertainty Random} experiments, where $\mathcal{L}=\mathcal{H}$, with the random initialisation scheme, $\mathcal{S}_1$, while we vary the latent dimension of the VAE used, $m$.

\begin{figure}[ht]
\centering
\includegraphics[width=0.45\textwidth]{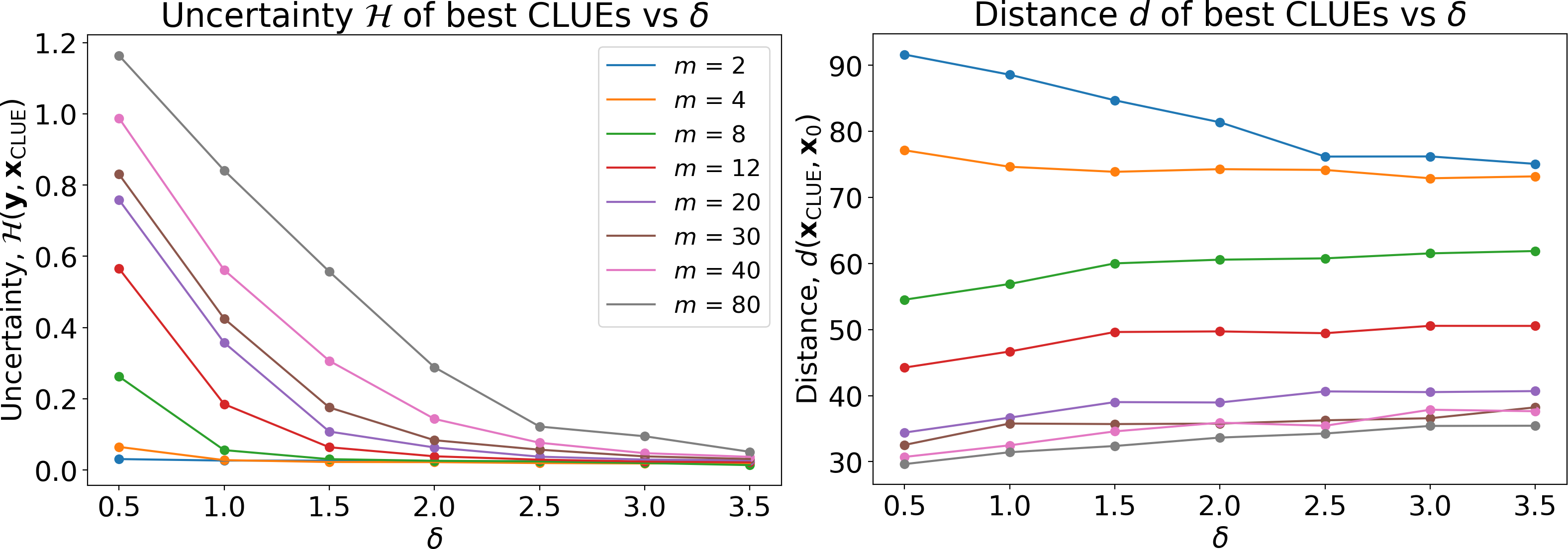}
\caption{\small Reproduction of Figure \ref{fig:uncertdistmin} for a range of VAE latent dimensions $m$. Reconstruction error of high dimensional VAEs is small (uncertainty remains high and distance remains low).}
\label{fig:VAEuncertdist}
\end{figure}

\begin{figure}[ht]
\centering
\includegraphics[width=0.45\textwidth]{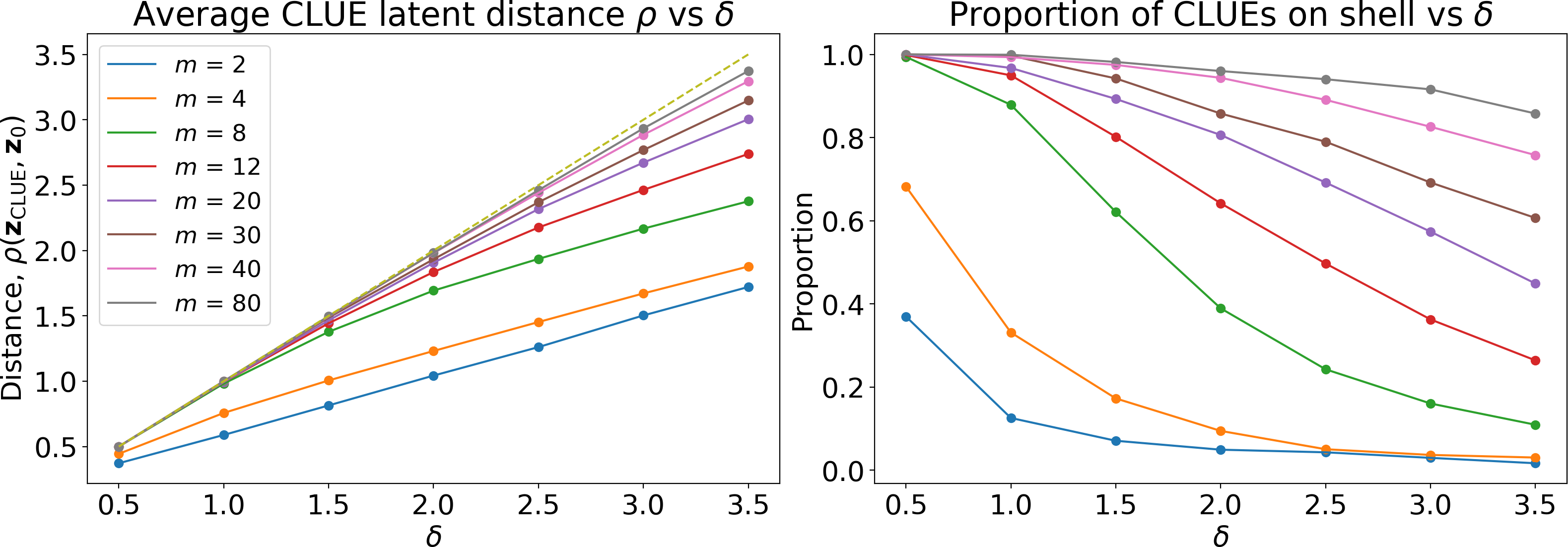}
\caption{\small We highlight the change in the optimal $\delta$ value with $m$ (as latent dimension increases, most minima occur at larger $\delta$, leading to a higher proportion of CLUEs on the shell).}
\label{fig:VAEdelta}
\end{figure}

\begin{figure}[ht]
\centering
\includegraphics[width=0.45\textwidth]{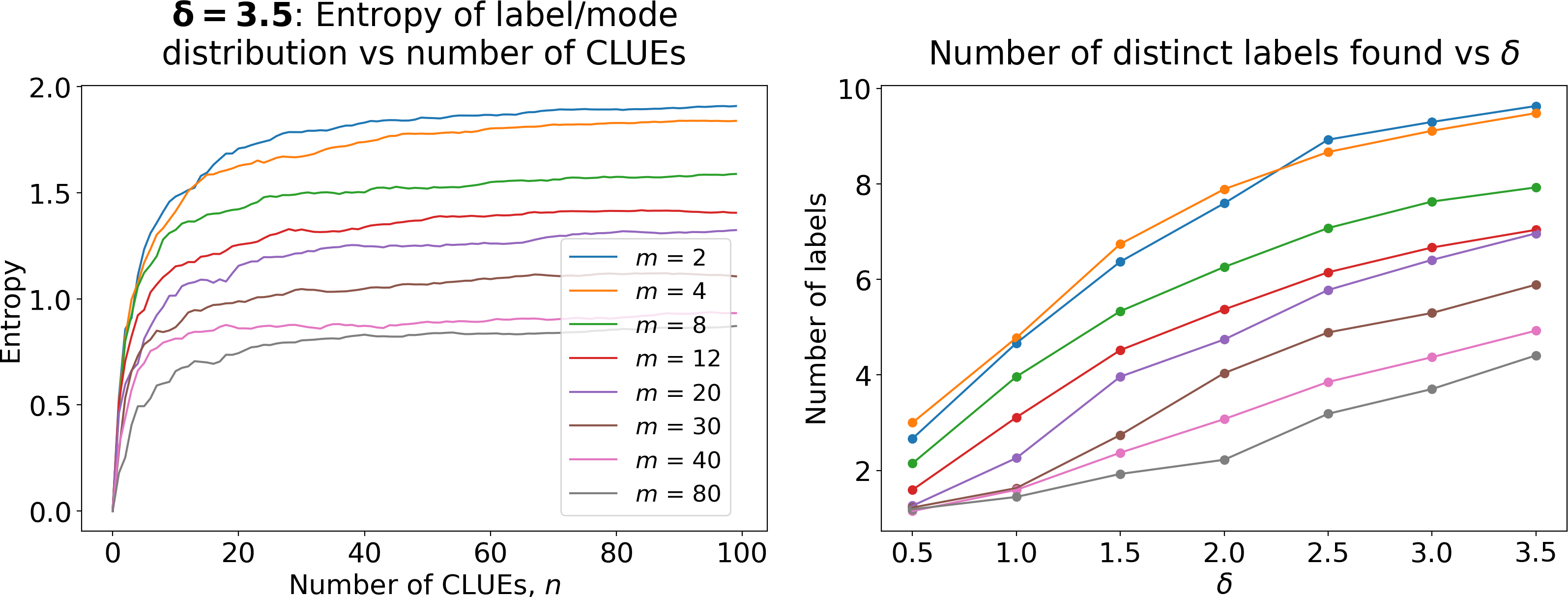}
\caption{\small We again highlight that larger latent dimensions require larger $\delta$ balls to capture necessary minima.}
\label{fig:VAEdiv}
\end{figure}

The first observation that uncertainty, $\mathcal{H}$ increases as $m$ increases is illustrated in Figure~\ref{fig:VAEuncertdist}. Since the reconstruction error increases as we decrease $m$, CLUEs become both blurrier and less uncertain, as the VAE fails to reconstruct a high quality image and resorts to more generic, smoothed shapes. Secondly, we see in Figure~\ref{fig:VAEdelta} that the distribution of CLUEs within the $\delta$ ball is pushed towards the surface of the ball as $m$ is increased. The final observation is that, for a given $\delta$, lower dimensional VAEs achieve greater diversity (Figure~\ref{fig:VAEdiv}) as a result of their failure to reconstruct images (this ties into the second observation).

\begin{figure}[ht]
    \centering
    \includegraphics[width=0.45\textwidth]{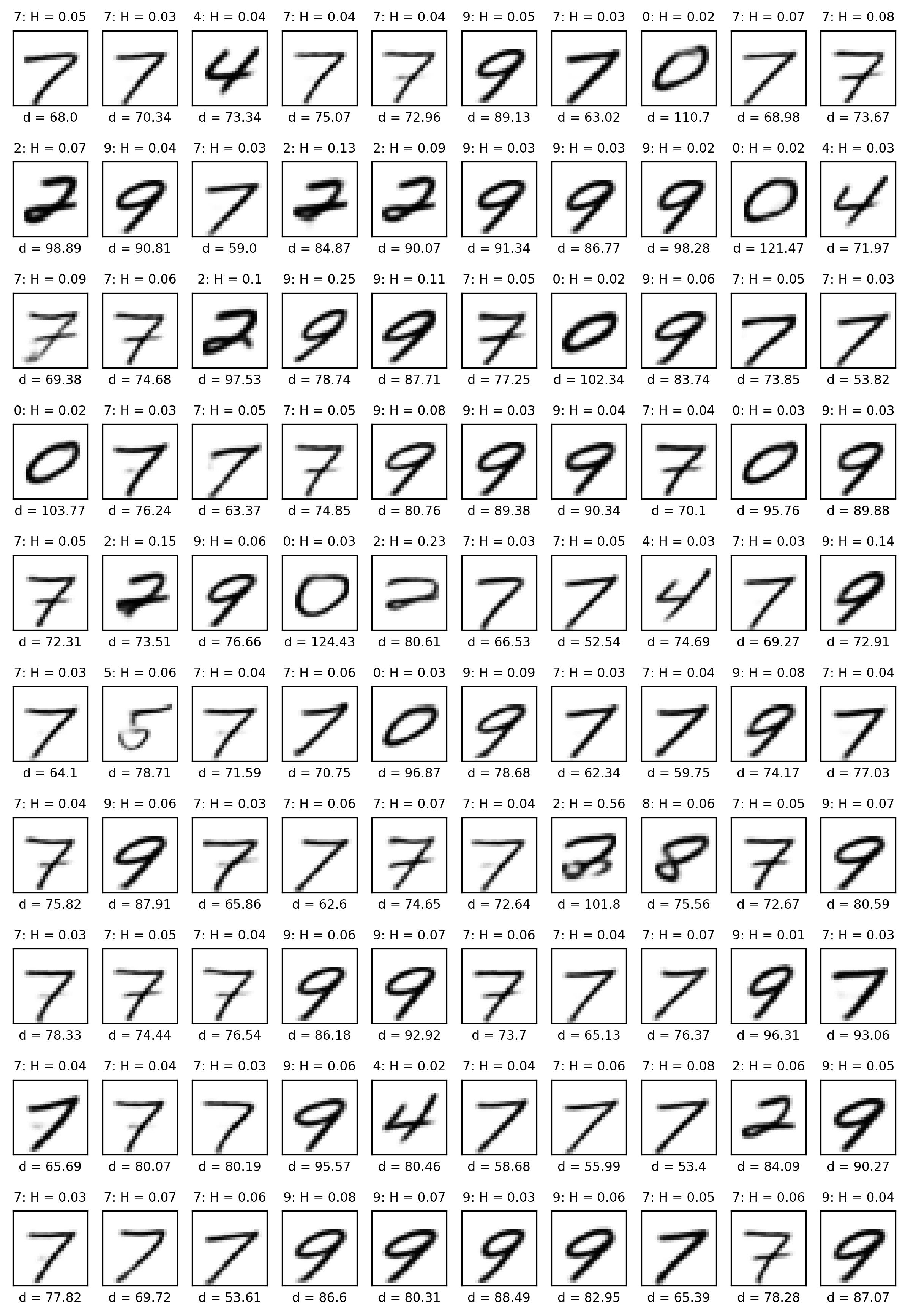}
    \caption{\small The 100 $\delta$-CLUEs yielded in this experiment (\textbf{Distance Random} with $\delta=3.5$). Above digits: Label prediction and uncertainty. Below: Distance from original in input space. Low uncertainty CLUEs may be found at the expense of a greater distance from the original input.}
    \label{fig:deltaCLUEs}
\end{figure}

\begin{figure}[ht]
    \centering
    \includegraphics[width=0.45\textwidth]{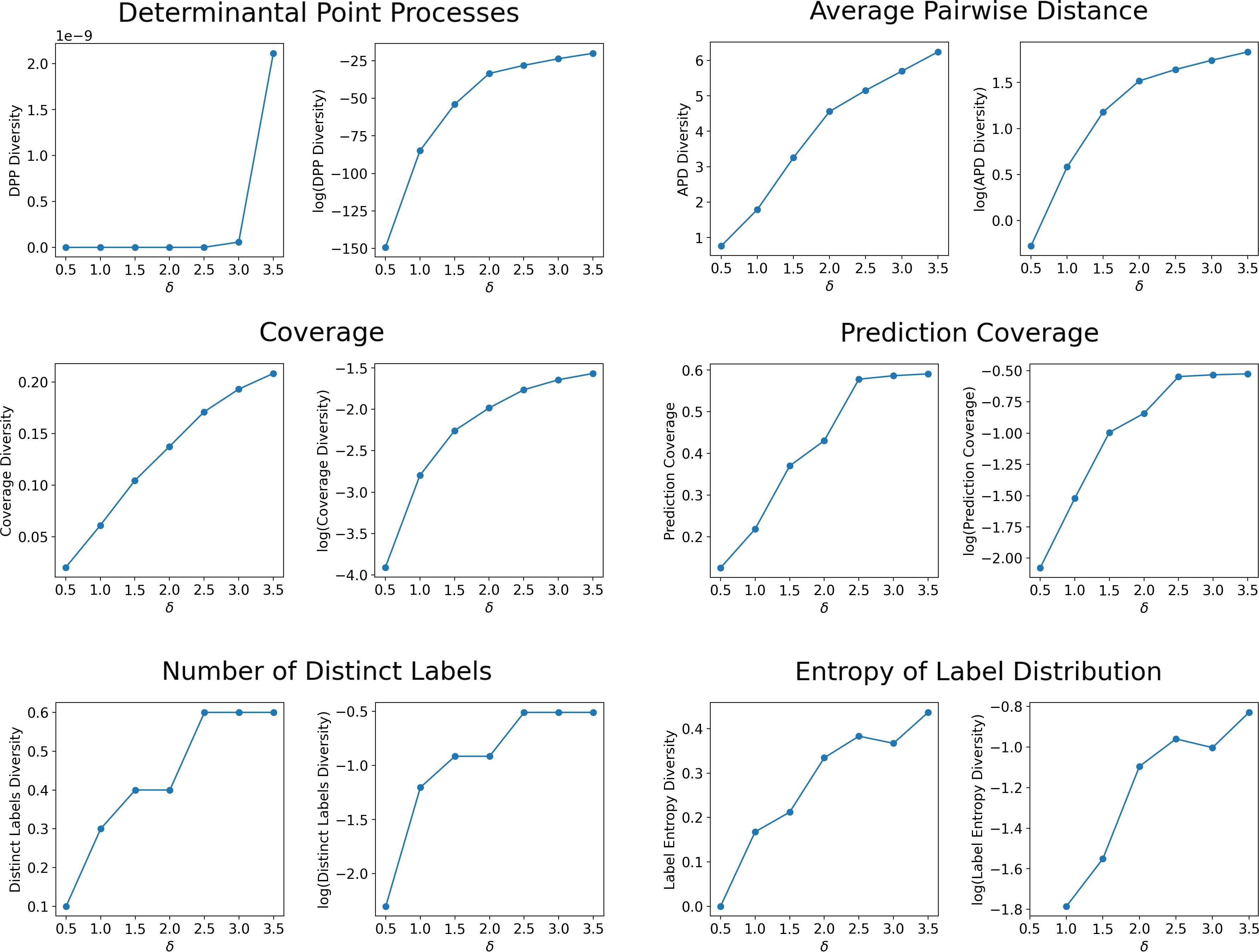}
    \caption{\small For a particular $\delta$ value and uncertain input $x_0$, we compute 100 $\delta$-CLUEs (as in Figure \ref{fig:deltaCLUEs}), repeating this over a range of $\delta$ values. For each of these sets, we apply the various diversity metrics proposed in Table \ref{tab:diversity} of the main text, plotting also the logarithms of the metrics (the latter only appearing to be meaningful with Determinantal Point Processes).}
    \label{fig:diversity_metrics}
\end{figure}

\section{C\quad Diversity Metrics for Counterfactual Explanations}
\label{appendix:diversitymetrics}

This appendix details the properties of each diversity metric in conjunction with sets of counterfactuals from different $\delta$ values. We also provide further clarity on the coverage diversity metric.

Observe in Figure~\ref{fig:diversity_metrics} that as $\delta$ increases, diversity increases virtually monotonically. The metrics operating in $\rvy$-space (prediction coverage, number of distinct labels and entropy of the label distribution) increase less smoothly with $\delta$. Furthermore, the volatility of the DPP measure is highlighted; sets of counterfactuals that led to marginal increases for all other metrics caused a sharp increase at $\delta\approx3.0$ for DPPs.

\begin{figure}[ht]
    \centering
    \includegraphics[width=0.45\textwidth]{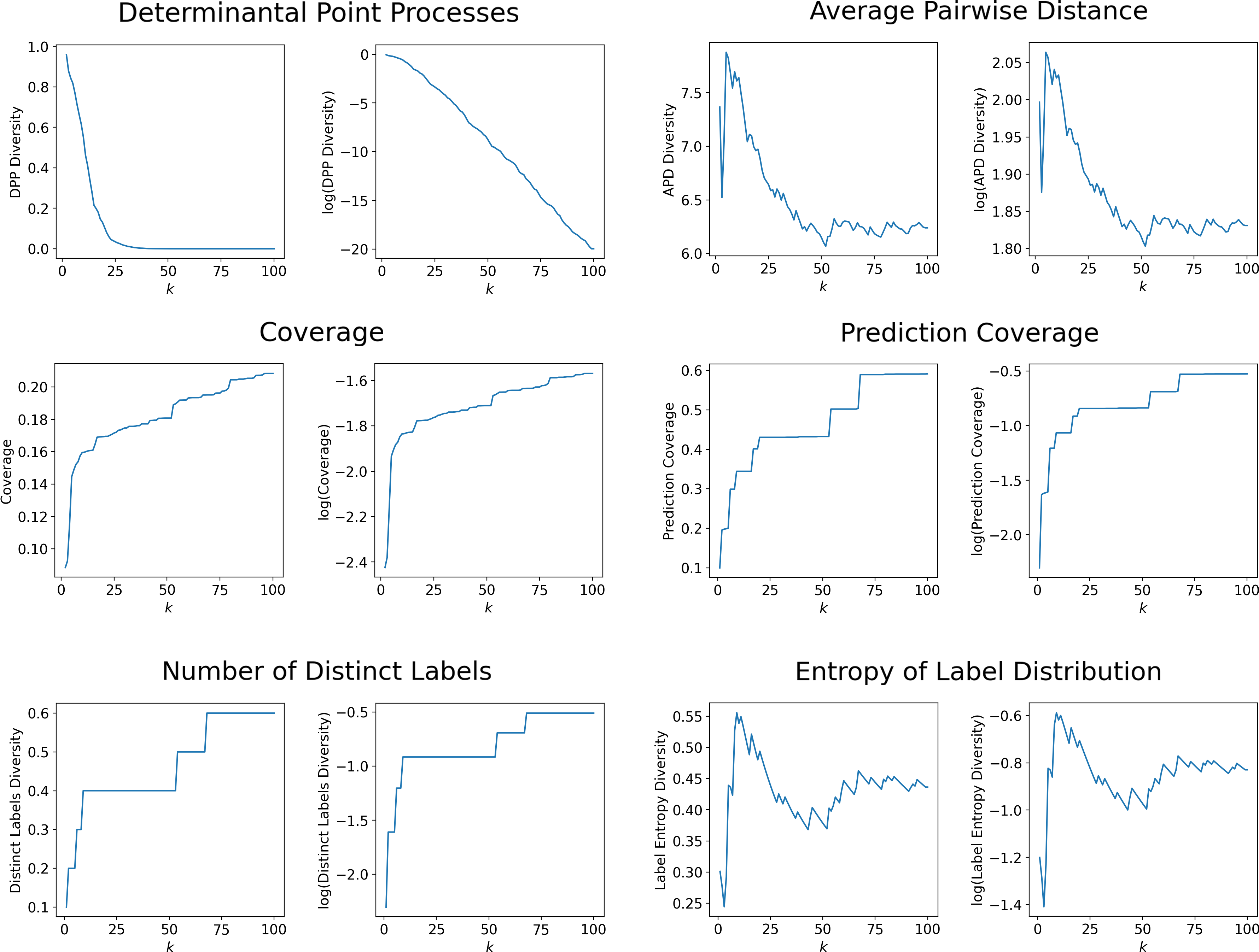}
    \caption{\small For a particular uncertain input $x_0$ and $\delta$ value, we compute 100 $\delta$-CLUEs, and proceed to plot the diversity of the first $k$ CLUEs in the range $1\leq k\leq100$.}
    \label{fig:diversity_metricsk}
\end{figure}

Figure~\ref{fig:diversity_metricsk} demonstrates further interesting properties of the metrics. Intuitively, we might expect the diversity in a set of $k$ counterfactuals to increase with $k$, but this is not the case for many of the metrics. Coverage, prediction coverage and the number of distinct labels all behave in this way, as is expected i.e. the number of distinct labels cannot reduce as more items are introduced. However, DPPs, APD and entropy of the label distribution exhibit decreases as $k$ increases. We expect the latter metric to increase and plateau, as in Figure~\ref{fig:diversityMNISTdeltaCLUE}, with some random noise around this convergence. However, if the diversity of a set does not rise fast enough with $k$ (which it does not in our experiments), DPPs and APDs fundamentally become smaller, as they normalise in some way with respect to $k$. The effect of $k$ on the diversity metrics has implications regarding the tuning of the hyperparameter $\lambda_D$, where the optimal value is now likely to change significantly with $k$. However, even for fixed $k$, this raises concerns for the sequential $\nabla$-CLUE optimisation, to be discussed in Appendix D.

\subsection{Coverage as a Diversity Metric}
\label{appendix:coverage}

\begin{figure}[ht]
    \centering
    \includegraphics[width=0.45\textwidth]{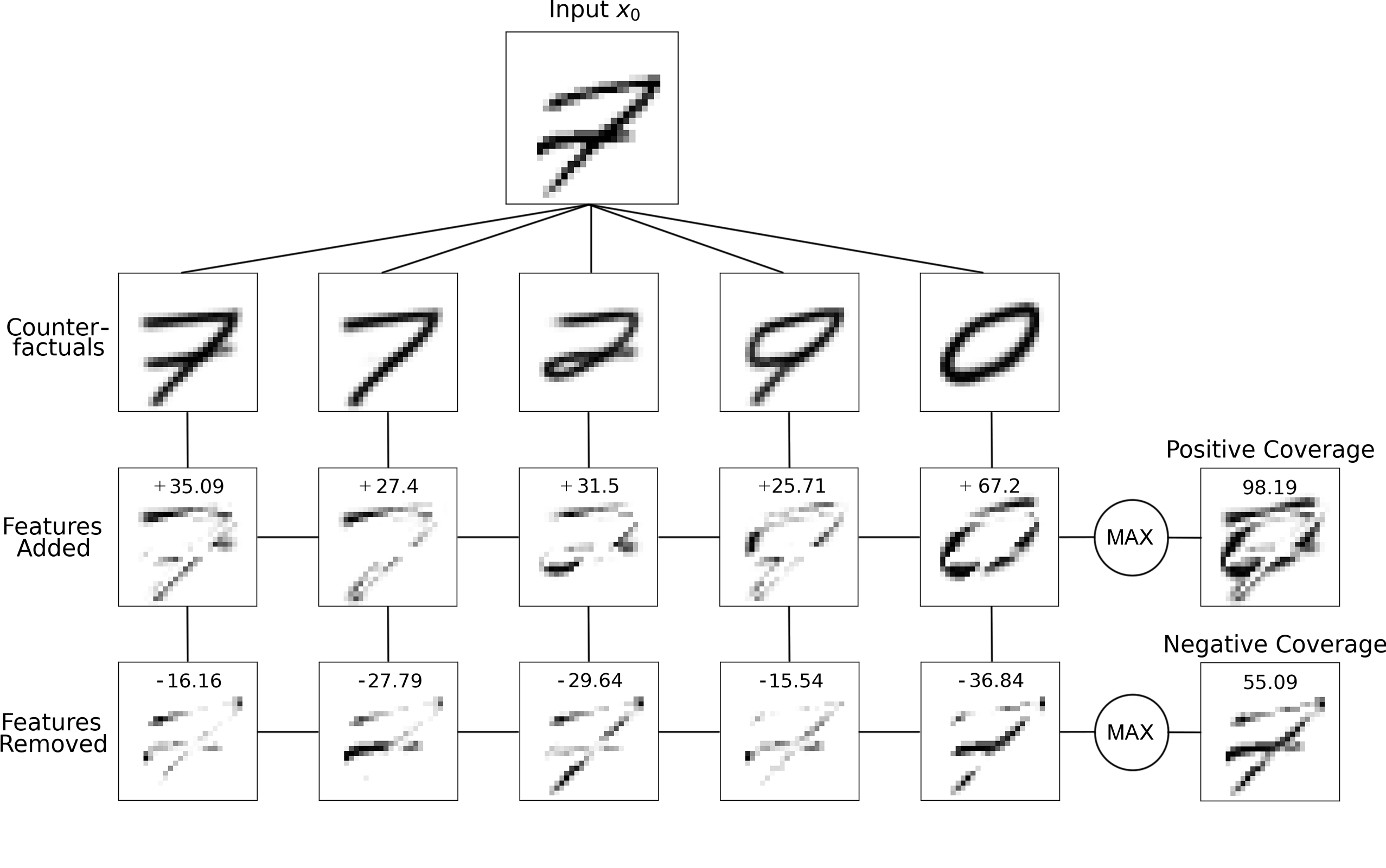}
    \caption{\small To compute positive and negative coverage, we take the positive and negative differences between counterfactuals and the original input, and further combine these by selecting the maximum change observed in a given feature (pixels in this case). We see that the 5 counterfactual explanations shown demonstrate changes that almost completely remove the original input, whilst adding features across a range of other areas. Total coverage is the sum of the positive and negative coverages.}
    \label{fig:coverage}
\end{figure}

Figure~\ref{fig:coverage} visualises this metric within MNIST to aid understanding. We now determine the bound on $D$ under the coverage metric. Take $S_+=\sum_{i=1}^{d'}\max(x_i)$ and $S_-=\sum_{i=1}^{d'}\min(x_i)$ to represent the sum over all features of the maximum and minimum values each feature can take, and that $|\rvx_0|=\sum_{i=1}^{d'}(\rvx_0)_i$ (the sum over all features of the uncertain input $\rvx_0$), where $d'$ is the dimensionality of the feature space. The minimum coverage of a counterfactual ($D=0$) clearly occurs when the counterfactual is simply the original input. The maximum coverage can be calculated as:

\[D_{\text{max}}=\frac{1}{d'}\left((S_+-|\rvx_0|)-(|\rvx_0|-S_-)\right)=\frac{S_+-S_-}{d'}\]
\[\text{(independent of $\rvx_0$).}\]

\noindent In the MNIST experiments performed, we have $d'=28\times28=784$, with the maximum and minimum values of each pixel to be 1 and 0 respectively, thus giving $S_+=784$ and $S_-=0$. This does indeed result in $D_{\text{max}}=1$. If $S_+$ and $S_-$ are known, we can guarantee this normalisation by dividing the coverage by $D_{\text{max}}$. In other applications, where the features can scale infinitely, normalising $D$ is not possible.

In theory, 2 counterfactuals are sufficient to achieve the maximum coverage (one counterfactual of all features at their maximum values, and one counterfactual of all features at their minimum values e.g. one fully black and one fully white image in MNIST). While coverage must can never decrease as $k$ increases, the exact nature of this relationship is dependent on the dataset and the counterfactual generation method (e.g. Figure~\ref{fig:diversity_metricsk}). This is analytically indeterminate and thus cannot be regularised.

\subsection{Future Work}

Future work might include a human subject experiment to determine the metric most aligned with human ideas of diversity; or better still, what each of the metrics represent themselves with regards to human intuition. The set of diversity metrics proposed in this paper are not exhaustive either, and further investigation of other metrics, perhaps with inspiration drawn from said human subject experiments, could provide meaningful insights.

\section{D\quad $\nabla$-CLUE}
\label{appendix:divCLUE}

\begin{figure}[ht]
    \centering
    \includegraphics[width=0.45\textwidth]{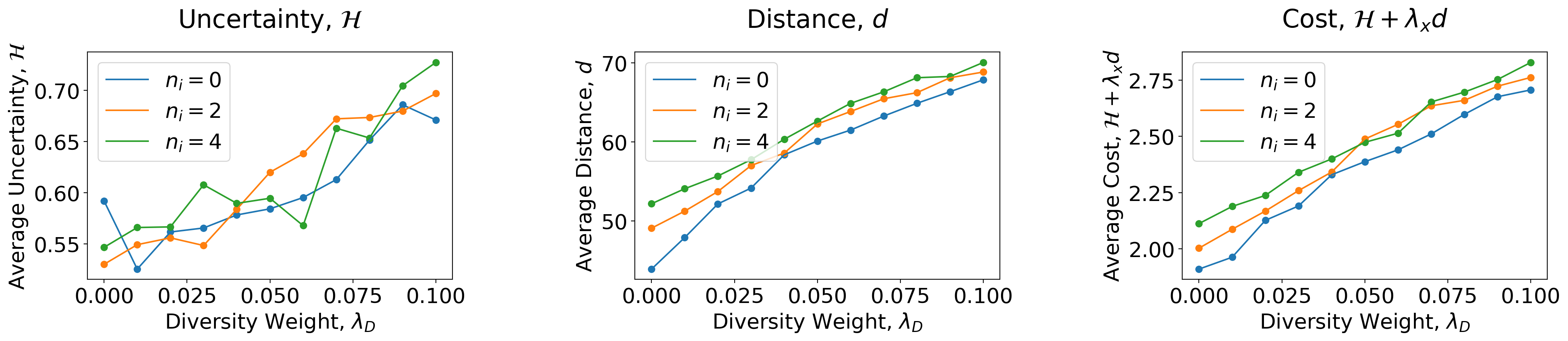}
    \caption{\small Effect of $\lambda_D$ and $n_i$ on performance. Unfortunately, performance degrades as we strive for greater diversity in the optimisation, although the gain in diversity can outweigh this.}
    \label{fig:divsimperformance}
\end{figure}

We observed that maximising the diversity of the \textit{initialisations} within the $\delta$ ball before performing the gradient descent helps to improve diversity; we thus perform an initial gradient descent to maximise the diversity of the initialisations, where $n_i$ is the number of gradient descent steps performed. However, as $n_i$ becomes large and diversity converges, we experience starting initialisations moving as far away from each other as possible in the $\delta$ ball, which can degrade performance with respect to uncertainty. As in the main text, we perform experiments at a fixed $\delta$ value and optimise for DPP diversity in $\rvz$-space. Results are displayed in Figures~\ref{fig:divsimdppapd} through \ref{fig:divsimperformance} (where $k=10$).

\begin{figure}[ht]
    \centering
    \includegraphics[width=0.45\textwidth]{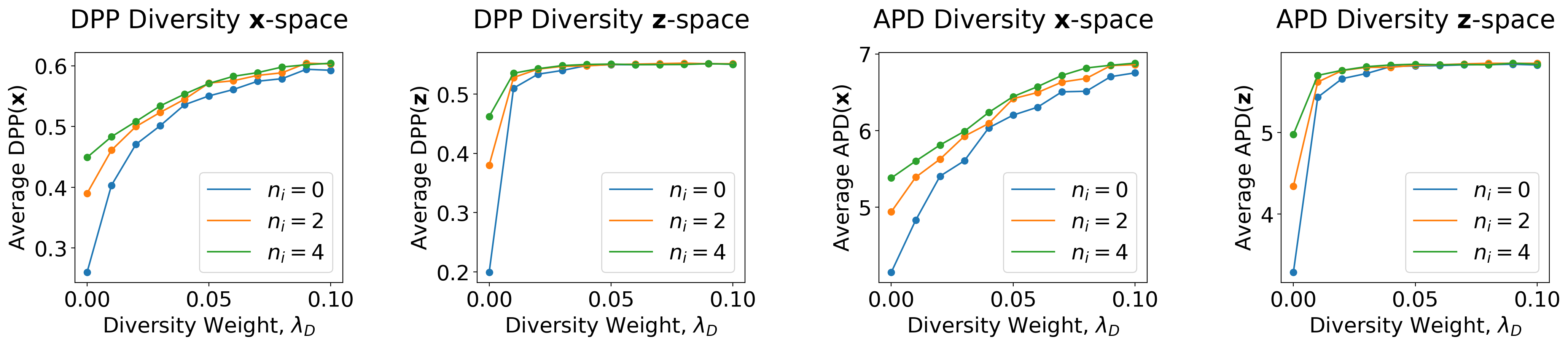}
    \caption{\small Effect of $\lambda_D$ and $n_i$ on diversity. We observe that either parameter can provide a route to achieving more diverse sets of counterfactuals.}
    \label{fig:divsimdppapd}
\end{figure}

\begin{figure}[ht]
    \centering
    \includegraphics[width=0.45\textwidth]{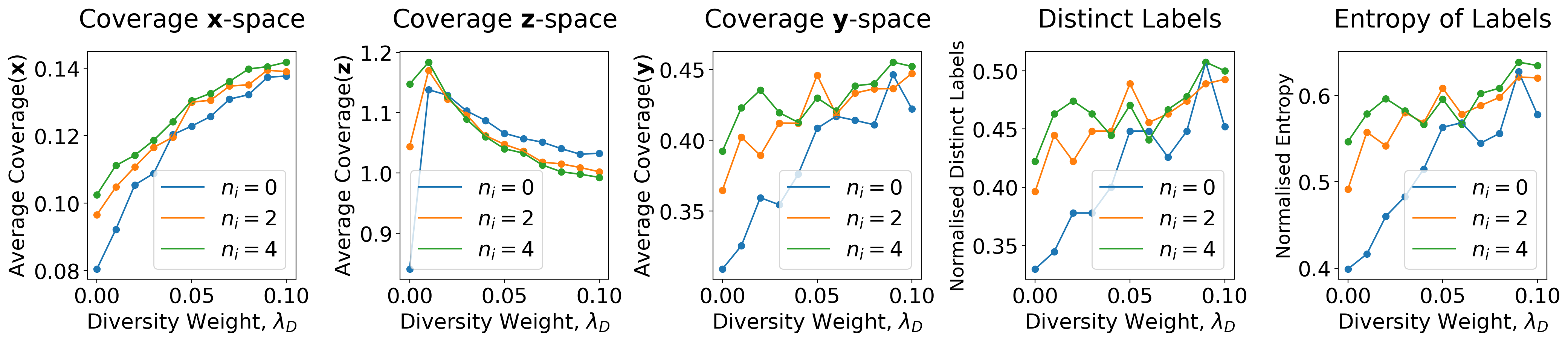}
    \caption{\small Continuation of Figure~\ref{fig:divsimdppapd} for the remaining diversity metrics.}
    \label{fig:divsimcovy}
\end{figure}

\begin{figure}[ht]
    \centering
    \includegraphics[width=0.45\textwidth]{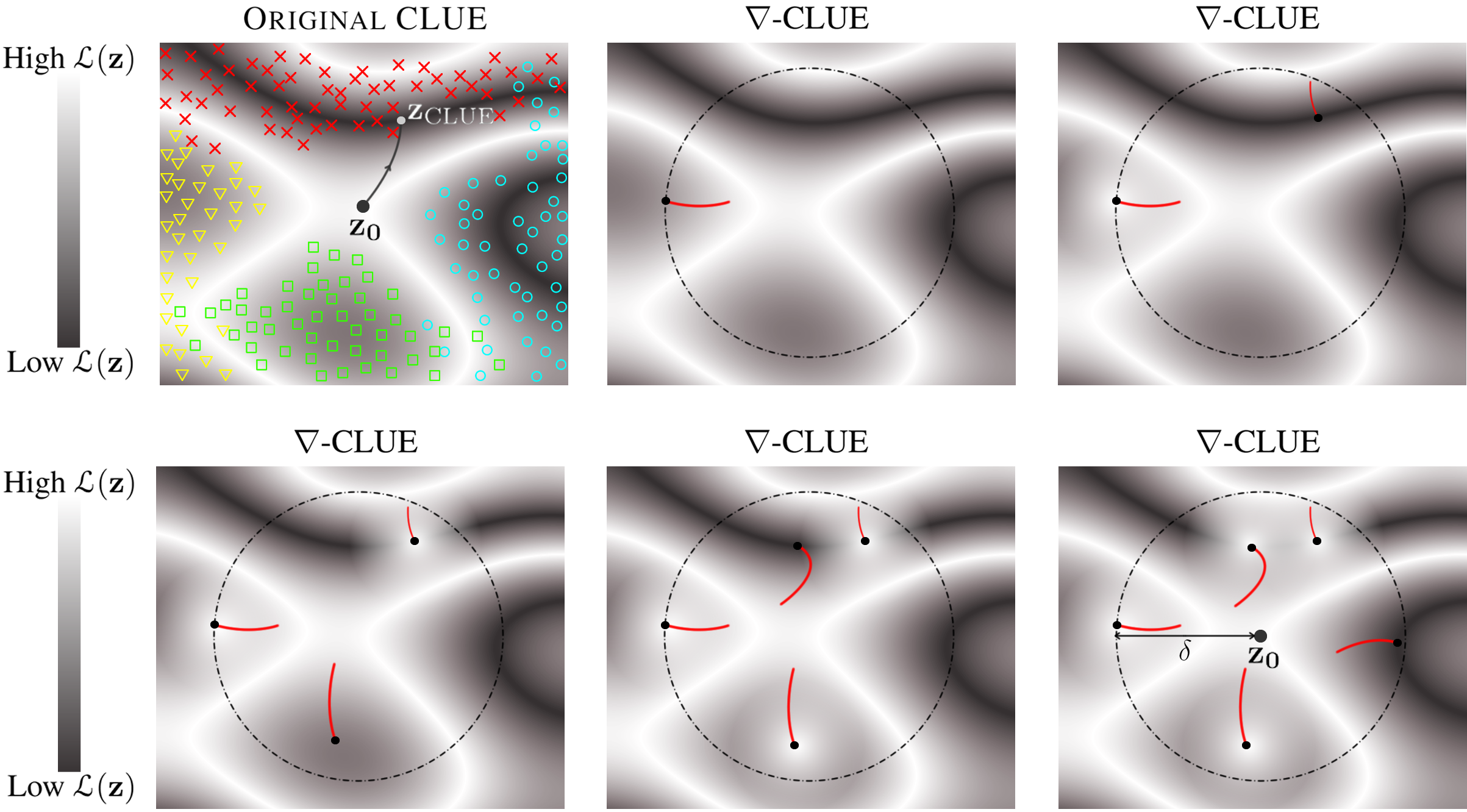}
    \caption{\small Conceptual colour map of objective function $\mathcal{L}(z)$ with $\mathbf{z}_0$ located in high cost region. Upper Left: Gradient descent to region of low cost (original CLUE). Training points in colour. Left to Right, Top to Bottom: Gradient descent constrained to $\delta$-ball using sequential $\nabla$-CLUE. The effect of the updated diversity term at each step is to push the current solution away from previous ones.}
    \label{fig:DivSequential}
\end{figure}

\subsection{Sequential $\nabla$-CLUE optimisation}

Figure \ref{fig:DivSequential} corresponds to Figure \ref{fig:DeltaFinal} of the main text for a sequential $\nabla$-CLUE scheme. Note the presence of regions of cost added with every new CLUE found (analogous to hills in the objective function) due to the diversity component in the objective function. The lower-center image demonstrates how this might affect the gradient descent of future CLUEs when in close proximity of older ones i.e. the optimisation path curves towards a new minima. Possible pitfalls of this method include the manner with which $D$ scales with $k$ (it is undesirable to have to re-tune the hyperparameter $\lambda_D$ at each iteration). For instance, we might wish to remove the normalisation term $\binom{k}{2}$ of APD diversity when performing sequential $\nabla$-CLUE.

\subsection{Sequential $\nabla$-CLUE with Penalty Terms}

This section trials the use of a penalty term instead of a diversity term in the objective of sequential $\nabla$-CLUE. The following experiments are conducted with $\delta=4$, $D=0$ and an added penalty term of $\frac{\lambda_D}{d(\rvz, \rvz_\mathrm{CLUE})}$ for each new counterfactual $\rvz_\mathrm{CLUE}$. This is run on the 8 most uncertain MNIST digits, generating $k=100$ CLUEs for each. We calculate diversity/performance of each set of 100 and average over all 8. Note that at $\lambda_D=0$, this is equivalent to running $\delta$-CLUE, and the loss function is not weighted by distance (i.e. $\mathcal{L}=\mathcal{H}$).

\begin{figure}[ht]
    \centering
    \includegraphics[width=0.45\textwidth]{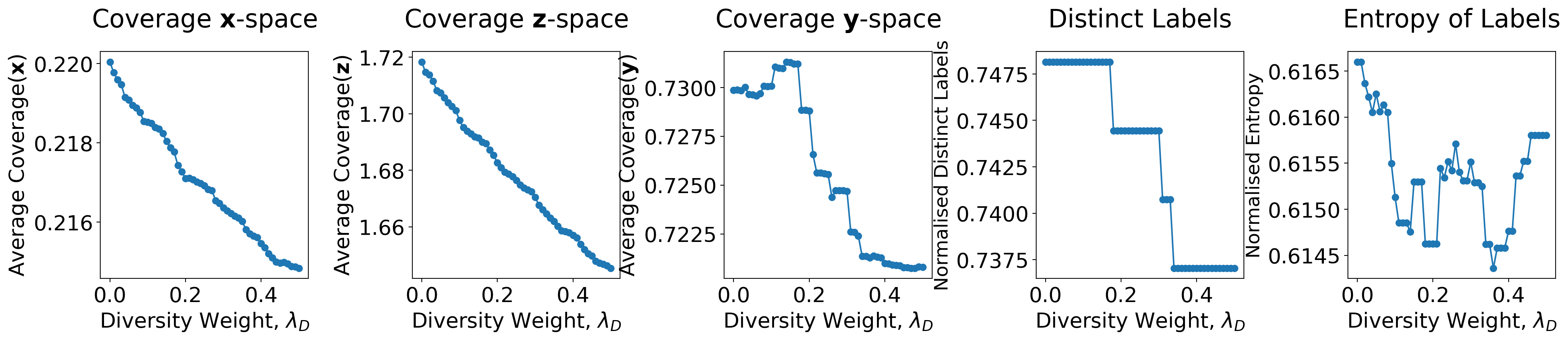}
    \caption{\small DPP and APD Diversity in $\rvx$ and $\rvz$-space as a function of $\lambda_D$. We see that similar patterns are exhibited in both $\rvx$ and $\rvz$-space.}
    \label{fig:diversityMNIST11}
\end{figure}

\begin{figure}[ht]
    \centering
    \includegraphics[width=0.45\textwidth]{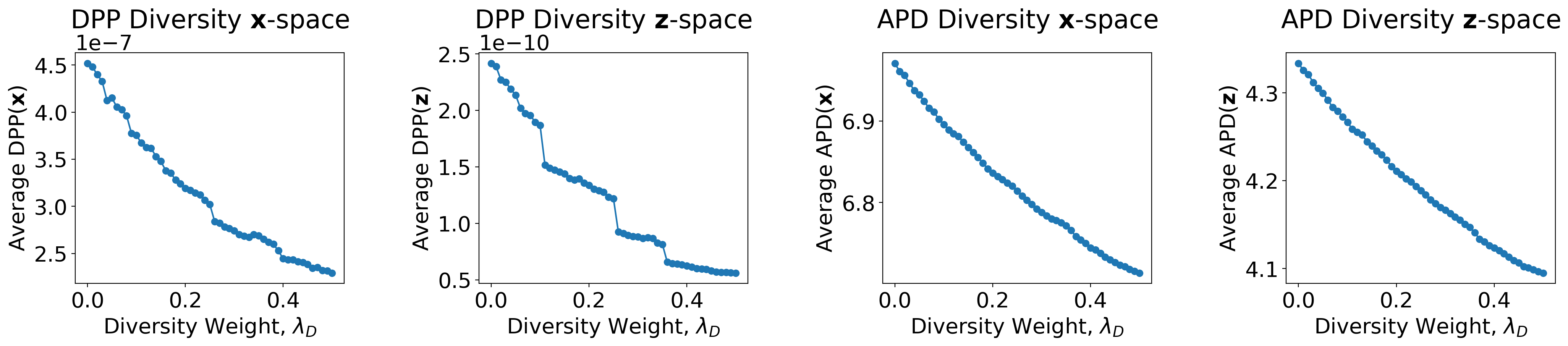}
    \caption{\small Coverage in $\rvx$, $\rvz$ and $\rvy$-space, Distinct Labels and Entropy of Labels as a function of $\lambda_D$.}
    \label{fig:diversityMNIST22}
\end{figure}

\begin{figure}[ht]
    \centering
    \includegraphics[width=0.45\textwidth]{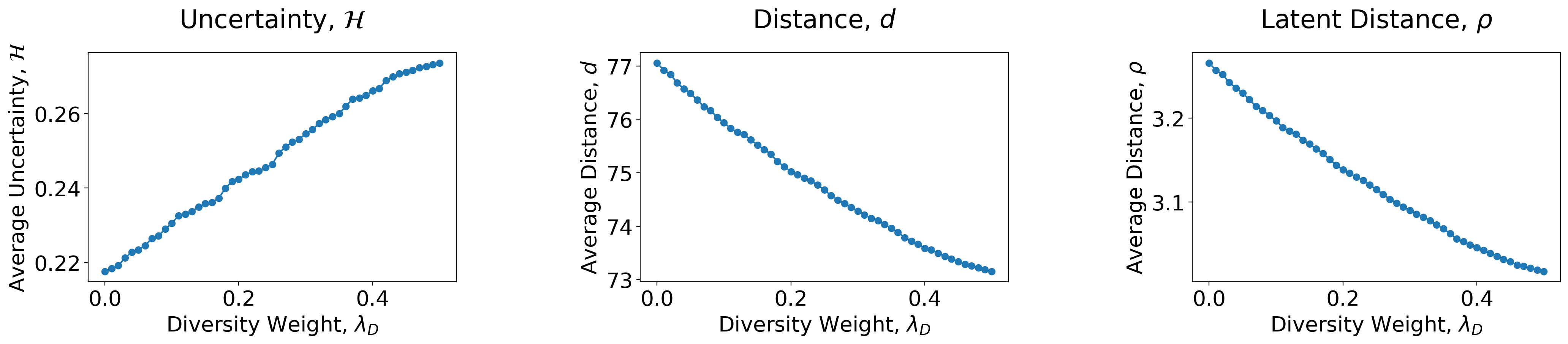}
    \caption{\small Uncertainty, $\mathcal{H}$ and distance in $\rvx$ and $\rvz$-space ($d$ and $\rho$ respectively) as a function of $\lambda_D$. Adding stronger diversity terms appears to push solutions towards the centre of the $\delta$ ball as suggested by the graph on the right. Our belief that on average input space distance decreases roughly monotonically as latent space distance decreases is confirmed.}
    \label{fig:performanceMNIST1}
\end{figure}

Figure \ref{fig:performanceMNIST1} details the effect of $\lambda_D$ on performance. This analysis suggests that $\delta$-CLUE performs better on all fronts for $k=100$ CLUEs (though $\nabla$-CLUE performance may change for a smaller number of CLUEs i.e. $k=10$). Figures \ref{fig:diversityMNIST11} and \ref{fig:diversityMNIST22} show that diversity actually decreases as a function of $\lambda_D$, implying that simple penalty terms, as described above, are insufficient for achieving diversity in the metrics put forward.

\subsection{Future Work}

We devote this section to performing a full ablative analysis of all diversity metrics, since only DPP diversity in $\rvz$-space was trialled in the main paper. We would also experiment more fully the strategy of finding diverse initialisations as opposed to optimising for diversity; the latter method, used in this paper, has been shown to compromise the performance of the CLUEs found, and thus finding the best starting initialisations and performing $\delta$-CLUE (where $\lambda_D=0$) might yield equally diverse sets that perform better.

\section{E\quad GLAM-CLUE: GLobal AMortised CLUE}
\label{appendix:GLAM-CLUE}

\begin{figure*}[ht]
    \centering
    \includegraphics[width=0.95\textwidth]{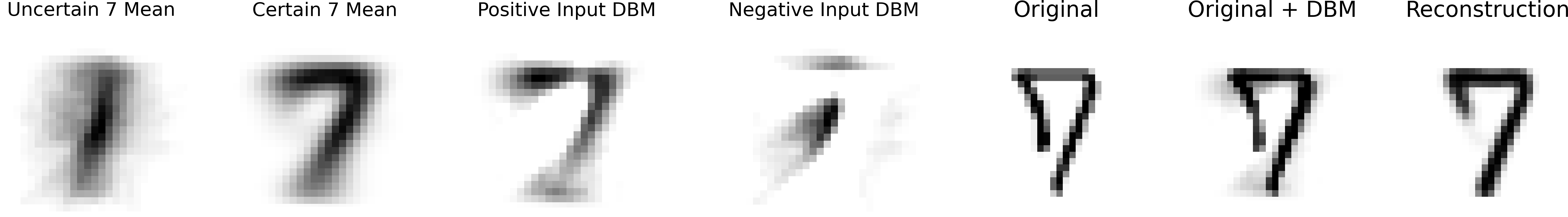}
    \caption{\small Visualisation of the input DBM baseline. The mean of all uncertain 7s in the MNIST training data is taken, followed by the mean of all certain 7s is shown in the 1st and 2nd plots. The 3rd and 4th plots show the positive and negative changes made when moving from uncertainty to certainty. The 5th to 7th plots illustrate how a final, certain counterfactual explanation is produced using this baseline (by adding the difference between means in input space and reconstructing the result).}
    \label{fig:inputDBMfull}
\end{figure*}

Although drawing inspiration from Transitive Global Translations (TGTs), as proposed by \citet{plumb2020explaining}, our method performs a different operation; instead of learning translations in input space that result in high quality mappings in a lower dimensional latent space, we find that results are best when learning translations in latent space, as described in the main text. This is seen also in the fact that the latent space DBM baseline outperforms input DBM; the difference between means translation is a special case of the GLAM-CLUE translation that we propose, and is the value we use as an initialisation during gradient descent. We also provide a visualisation for the input space DBM baseline in Figure~\ref{fig:inputDBMfull}. In the case of image data, the resulting image when DBM was added to the original input had to be clipped to match the scale of the data (in our case, between 0 and 1).

\subsection{Future Work}

\begin{figure}[ht]
    \centering
    \includegraphics[width=0.5\textwidth]{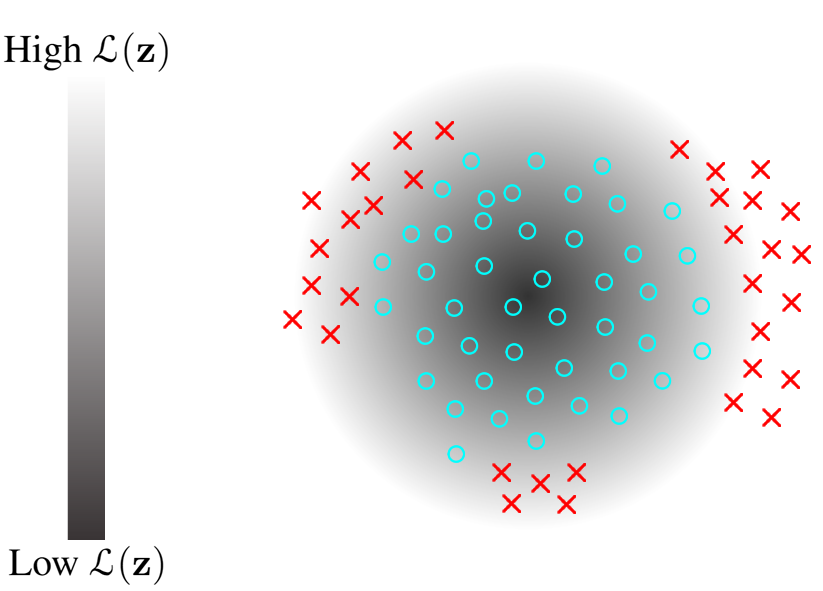}
    \caption{\small 2D visualisation of a shortcoming of GLAM-CLUE. The group of uncertain points (red) is not easily mapped onto the group of certain points (blue) by a single translation, unless further division of the uncertain group (into 3 clusters for instance) is performed, or a more complex mapper is learnt.}
    \label{fig:LossGLAM-CLUE}
\end{figure}

While a GLAM-CLUE translation shows very good performance in the experiments demonstrated in the main text, it is not clear that performance would be maintained in all situations. We question the performance in cases where a group of uncertain points are not easily separated from a group of certain points by a simple translation (as in Figure~\ref{fig:LossGLAM-CLUE}).

To this end, there are two further avenues to explore: the use of more complex mapping functions, or the potential to split the uncertain groups into groups that translations perform well on (clustering the uncertain points in Figure~\ref{fig:LossGLAM-CLUE} into 3 groups). This latter approach would maintain GLAM-CLUE's utility in computational efficiency, as we demonstrate that learning simple translations is extremely fast.

We have the additional issue of selecting an appropriate $\lambda_\theta$ parameter in the algorithm to best tune the trade-off between uncertainty and distance. \citet{Dosovitskiy2020You} propose a method that replaces multiple models trained on one loss function each by a single model trained on a distribution of losses. A similar approach could be taken by using a distribution over individual terms of our objective and varying the hyperparameter weight at the \textbf{inference step}. This could yield a powerful technique for minimising uncertainty and distance but allowing the trade-off between the two to be selected \textbf{post-training}.

As far as more complex datasets are concerned, preliminary trials on the black and white Synbols dataset showed that the DBM baselines produced almost incoherent results. Our understanding is that, in input space, taking the mean of a particular class that contains an equal distribution of points with white backgrounds and points with black backgrounds will result in a cancellation between the two, such that the mean vector is close to zero. The same analogy in latent space might be that black points within a particular class may not be clustered in a similar region to those of white points for the same class. As such, further clustering, as alluded to above and in Figure~\ref{fig:LossGLAM-CLUE} is probably necessary.

\end{document}